\documentclass[conference]{IEEEtran}

\usepackage{etoc}

\usepackage[numbers]{natbib}
\usepackage{multicol}
\usepackage[bookmarks=true]{hyperref}
\usepackage{comment}
\usepackage{amsmath}
\usepackage{amssymb}
\usepackage{graphicx}                                     
\usepackage[export]{adjustbox}
\usepackage[dvipsnames]{xcolor}                           
\usepackage{url}                                         

\usepackage{float}
\usepackage{times}
\usepackage{multirow}
\usepackage{multicol}
\usepackage{booktabs}
\usepackage{subcaption}
\usepackage{placeins}
\usepackage[normalem]{ulem} 

\usepackage[ruled,vlined,linesnumbered]{algorithm2e}
\SetCommentSty{scriptsize}            
\SetKwComment{tcp}{\(\triangleright\)\ }{}  
\SetAlFnt{\small}        
\SetAlCapNameFnt{\small} 
\SetAlgoNlRelativeSize{-1}      

\newcommand{\imgrow}[3]{%
  \begin{subfigure}[t]{0.32\textwidth}
    \centering
    \includegraphics[width=\linewidth,height=0.13\textheight,keepaspectratio]{#1}
  \end{subfigure}\hfill
  \begin{subfigure}[t]{0.32\textwidth}
    \centering
    \includegraphics[width=\linewidth,height=0.13\textheight,keepaspectratio]{#2}
  \end{subfigure}\hfill
  \begin{subfigure}[t]{0.32\textwidth}
    \centering
    \includegraphics[width=\linewidth,height=0.13\textheight,keepaspectratio]{#3}
  \end{subfigure}
}

\newcommand{\rowsubcap}[2]{%
  \vspace{0.4em}
  \caption*{\small\textbf{#1:} #2}
  \vspace{0.7em}
}

\IEEEoverridecommandlockouts

\begin{document}


\title{
Efficient Multi-Robot Motion Planning with Precomputed Translation-Invariant Edge Bundles
}

\author{
Himanshu Gupta${^{1}}$, Paul Motter${^{2}}$, Aritra Chakrabarty${^{2}}$, Rishabh Sodani${^{3}}$, Srikrishna Bangalore Raghu${^{2}}$, \\
Alessandro Roncone${^{2}}$, Bradley Hayes${^{2}}$ and Zachary Sunberg${^{1}}$
\thanks{*This work was supported by NSF CAREER Award \#2340958.}
\thanks{Affiliations: $^{1}$Smead Aerospace Engineering Sciences, University of Colorado Boulder; $^{2}$Department of Computer Science, University of Colorado Boulder; $^{3}$Cherry Creek High School, Greenwood Village, CO, USA.
}
}




%

\maketitle

\begin{abstract}

Solving multi-robot motion planning (MRMP) requires generating collision-free kinodynamically feasible trajectories for multiple interacting robots.
We introduce Kinodynamic Translation-Invariant Edge Bundles or KiTE-Extend, a planner-agnostic action selection mechanism for sampling-based kinodynamic motion planning.
KiTE-Extend uses a library of trajectory segments computed offline to guide action selection during online planning, improving the ability of existing planners to identify feasible motion segments without altering state propagation, collision checking, or cost evaluation, and without changing their theoretical guarantees.
While KiTE-Extend can modestly improve single-agent planners, its benefits are most clear in the multi-agent setting, where it is able to explore more effectively and significantly improve planning through the dense spatiotemporal constraints introduced by robot-robot interaction.
Through experiments on multiple kinodynamic systems and environments, we show that KiTE-Extend reduces planning time and improves scalability across the three most common MRMP paradigms: centralized, prioritized, and conflict-based.
\textcolor{blue}{\href{https://kite-extend-mrmp.github.io/}{(Project Webpage)}}

\end{abstract}
\section{Introduction}

Coordinating multiple robots in shared environments requires computing trajectories that avoid both static obstacles and inter-robot collisions. 
We consider the multi-robot motion planning (MRMP) problem. 
Given multiple robots governed by kinodynamic constraints, the objective is to compute a set of time-parameterized trajectories to their respective goal regions that are dynamically feasible for each robot and collision-free for all robots over time.
Multi-agent path finding (MAPF), the discrete analogue of MRMP has been extensively studied in the literature \cite{stern2019multi}, where agents move along edges of a finite graph with known traversal times.
However, solution methods for MAPF do not extend straightforwardly to MRMP, as MRMP operates in continuous state spaces, must account for robot geometry, and cannot assume that transitions between discrete states are dynamically realizable~\cite{kottingerConflictBasedSearchMultiRobot2022,moldagalieva2024db}.

Existing MRMP methods can be broadly categorized into three paradigms based on how multi-robot coordination is handled. 
Centralized planners search directly in the joint state space, offering a principled formulation but suffering from exponential growth in computational complexity with the number of robots~\cite{wagner2015subdimensional,shome2020drrt}.
Prioritized planners plan sequentially according to a fixed or adaptive priority ordering, treating previously planned robots as dynamic obstacles; while often effective in practice, their performance is sensitive to priority choices, and completeness typically requires exploring a large priority tree \cite{van2005prioritized,yakovlev2019,zhang2025alternating}.
Conflict-based search (CBS)-style methods strike a balance by performing a coupled search over conflicts and constraints at a high level, while invoking a single-robot planner at the low level to compute trajectories that satisfy accumulated spatiotemporal constraints \cite{kottingerConflictBasedSearchMultiRobot2022,moldagalieva2024db}.

\begin{figure}
    \centering    \includegraphics[width=0.8\columnwidth]{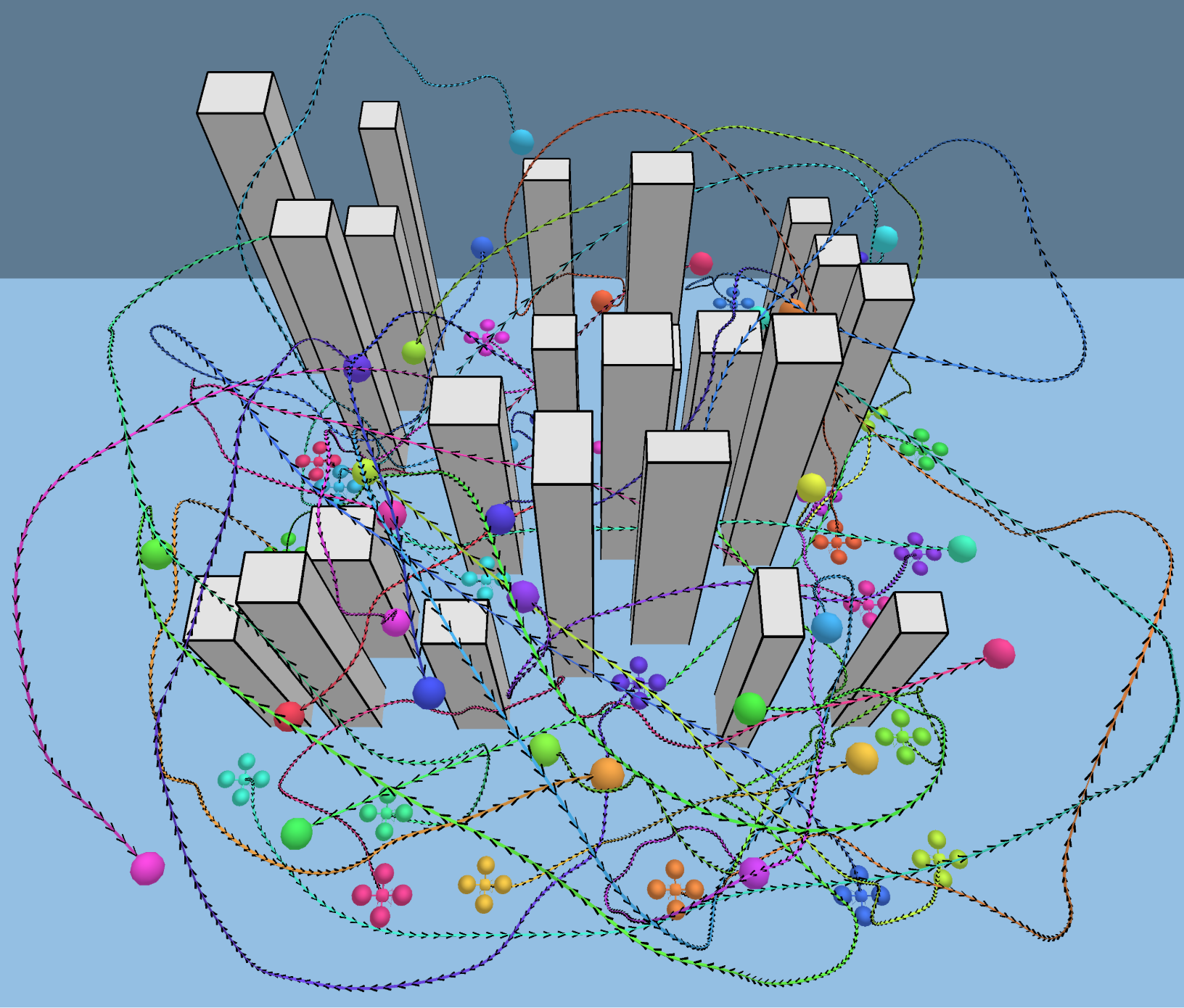}
    \caption{
    \small
    Example multi-robot solution for 30 quadcopters with double-integrator dynamics in a cluttered environment. KCBS is used for high-level coordination, with KiTE-Extend guided RRT as the low-level kinodynamic planner. Colored curves denote agent trajectories, with matching-colored markers indicating start and goal states; gray prisms represent static obstacles.
    }
    \label{fig:environment}
\end{figure}

Despite substantial differences in coordination strategies, all MRMP paradigms incur the cost of generating dynamically feasible trajectories under coupled inter-robot constraints, which in practice is dominated by repeated low-level node expansion and feasibility checking under evolving spatiotemporal constraints.
To mitigate this cost, recent works incorporate offline computation within search-based frameworks, leveraging translation invariance to reuse precomputed motion segments across the workspace~\cite{moldagalieva2024db,moldagalieva2025dbecbs}.
In these approaches, motion representations are first constructed to facilitate combinatorial coordination, and dynamic feasibility is subsequently enforced through repair, tracking, or optimization stages.
While effective in some settings, this separation introduces additional computational overhead, as the cost of the optimization stage scales poorly with both the number of agents and the trajectory horizon.

A key property of sampling-based planning (SBP) for kinodynamic systems is that dynamical feasibility is preserved by construction through forward simulation, avoiding the need for a separate repair step.
This property motivates a different role for offline computation in MRMP: rather than producing motion representations that must later be repaired, offline computation can be used to directly accelerate feasible SBP expansions during online planning, while still exploiting translation invariance to enable reuse.

This paper introduces \textbf{Ki}nodynamic \textbf{T}ranslation-Invariant \textbf{E}dge-Bundles or KiTE-Extend, a planner-agnostic action selection scheme for sampling-based kinodynamic motion planning.
KiTE-Extend leverages offline computation to bias online node expansion using translation-invariant kinodynamic edge bundles, improving the efficiency of reaching useful successor states while leaving the underlying planner unchanged.
Unlike motion-primitive or lattice-based planners that define a discrete search space, KiTE-Extend operates within a continuous sampling-based framework, where edge bundles are used only to guide expansion rather than restrict it.
Candidate motions are re-propagated from the current state and validated under current constraints.
As a result, KiTE-Extend preserves continuous exploration and maintains the theoretical properties of the base sampling-based planner.
KiTE-Extend operates purely at the level of action selection and is compatible with a broad class of sampling-based kinodynamic planners and can be integrated into centralized, prioritized, and CBS-style MRMP frameworks alike.

The main contributions of this work are:

\begin{itemize}
    \item We introduce KiTE-Extend, a planner-agnostic action selection scheme that exploits translation-invariant offline computation to accelerate kinodynamic node expansion while preserving the theoretical guarantees of sampling-based planning.
    \item We demonstrate that incorporating KiTE-Extend into sampling-based planners improves success rates, reduces computation time, and yields higher-quality solutions when used within centralized, prioritized, and CBS-style MRMP frameworks across cluttered environments and multiple kinodynamic systems.
\end{itemize}

We emphasize that our focus is not on designing new high-level coordination strategies, but on improving the efficiency of SBP through offline computation for MRMP.
Our results show that incorporating offline reuse within sampling-based planning consistently improves performance, and can outperform existing approaches that use offline computation within discontinuity-bounded, optimization-based pipelines such as db-CBS~\cite{moldagalieva2024db}.
While more sophisticated coordination mechanisms can further improve performance, they are orthogonal to our contribution; the proposed approach can be combined with such methods to yield additional gains.


\section{RELATED WORK}

Single-agent kinodynamic motion planning has been extensively studied using optimization-based methods \cite{schulman2014motion,malyuta2022convex}, lattice-based search \cite{pivtoraiko2011kinodynamic,pivtoraiko2012differentially}, and sampling-based planning (SBP) \cite{lavalle2001randomized,li2016asymptotically}. 
Among these, SBP has been widely adopted due to its generality and strong theoretical guarantees, including probabilistic completeness and asymptotic optimality. 
In practice, SBP's performance is often dominated by node selection and expansion mechanisms, particularly for kinodynamic systems where feasibility constraints are tight, foreshadowing the difficulties encountered in multi-robot settings.

\emph{Kinodynamic expansion:} In SBP, extending the search tree toward a sampled state for kinodynamic systems requires solving a steering problem involving system dynamics. 
For many robotic systems, exact solutions to such two-point boundary value problems (TPBVPs) are unavailable, and computing feasible connections requires nonlinear trajectory optimization, which is computationally expensive. 
To mitigate this cost, prior work has explored approximate steering via local linearization \cite{perez2012lqr} or analytic steering functions for specific system classes \cite{webb2013kinodynamic}, limiting applicability to systems with favorable structure.
Other approaches avoid solving TPBVPs by forward propagating randomly sampled controls for random durations \cite{hsu2002randomized,li2016asymptotically}, or by selecting among multiple propagated trajectories based on proximity to the sampled state \cite{kalisiak2006rrt,littlefield2017informed}.
However, random or weakly guided control propagation is uninformative for highly nonlinear or underactuated systems, leading to poor exploration efficiency \cite{schramm2022learning,atreya2022state}.
When such expansion inefficiencies are compounded by repeated replanning, as in MRMP, they quickly dominate overall computation.

\emph{Learned expansions:}
Recent work improves node expansion by replacing random control sampling with learned steering functions or action proposals \cite{atreya2022state,schramm2022learning,hassidof2025train}.
These approaches are primarily evaluated in single-agent settings with static environments.
In practice, learned expansions must still undergo revalidation and subtree repair \cite{atreya2022state} or be rejected in favor of fallback random sampling \cite{schramm2022learning,hassidof2025train}, diminishing their informativeness and resulting in wasted computation when replanning is triggered repeatedly.

\emph{Offline reuse:}
Other complementary works explore offline precomputation to improve kinodynamic expansion by constructing dynamically feasible trajectory segments.
Edge bundles, large collections of kinodynamically feasible trajectory segments sampled across the state space, have been used to guide search~\cite{shome2021asymptotically,pasricha2024virtues}.
However, achieving adequate coverage in higher-dimensional or larger environments requires large libraries, and candidate segments must be repeatedly filtered under changing spatiotemporal constraints, introducing significant online overhead.
Lattice-based planners~\cite{pivtoraiko2011kinodynamic} and motion-primitive approaches~\cite{sakcak2019sampling} exploit translation invariance to precompute reusable trajectory segments over discretized state spaces, reducing online control synthesis at the cost of restricting the search to a finite set of states and transitions.
Recent discontinuity-bounded planners, iDb-A$*$~\cite{ortiz2024idb_astar} and iDb-RRT~\cite{ortiz2024idb_rrt}, further refine this idea, but rely on post hoc trajectory optimization to repair infeasible concatenations.
Such repair-based approaches introduce significant computational overhead, as the cost of the underlying optimization scales poorly with the number of constraints and the trajectory horizon, limiting scalability in interaction-rich MRMP settings.

\emph{MRMP:}
Coordinating dynamically constrained agents while avoiding collisions makes MRMP substantially more complex than single-agent planning.
A variety of MRMP approaches have been proposed to improve scalability, including continuous space–time reasoning~\cite{tang2020multi,sim2025st}, decomposition-based coordination~\cite{qin2025k,mcbeth2023scalable}, relaxed continuity~\cite{moldagalieva2025db}, and heuristic prioritization~\cite{zhang2025alternating}, often trading completeness or exact kinodynamic feasibility for scalability.
Conflict-Based Search (CBS) has therefore been widely adopted due to its favorable balance between scalability and principled coordination~\cite{kottingerConflictBasedSearchMultiRobot2022,moldagalieva2024db}.
K-CBS \cite{kottingerConflictBasedSearchMultiRobot2022} integrates sampling-based kinodynamic planners into CBS while preserving probabilistic completeness guarantees.
Recent works such as db-CBS~\cite{moldagalieva2024db} and db-ECBS~\cite{moldagalieva2025dbecbs} incorporate offline computations into MRMP.
These works use db-A* as a low-level planner to rapidly generate approximately feasible trajectories, followed by joint trajectory optimization to enforce continuity and inter-agent collision constraints.
While effective for small teams, this approach requires solving large joint optimization problems whose cost grows with both the number of agents and the trajectory horizon, limiting scalability in densely coupled settings.
Another method, db-LaCAM~\cite{moldagalieva2025db}, improves scalability through more efficient coordination and by removing the repair step, but does not guarantee feasible trajectories due to relaxed continuity.
In contrast, SBP constructs dynamically feasible trajectories through forward simulation and does not require a repair step.
Consequently, effectively leveraging offline computation to improve the efficiency of SBP in MRMP remains an important and largely open direction.

The preceding discussion highlights a gap between offline kinodynamic reuse and its effective deployment under repeated, constraint-driven expansion in MRMP.
KiTE-Extend directly addresses this gap by reframing node expansion as a retrieval problem over a large translation-invariant space of kinodynamically feasible trajectory segments.
While both motion primitives and translation-invariant edge bundles represent reusable trajectory objects, the key distinction lies in how they are used within the planner.
Search-based planners use motion primitives to define the edges of a fixed graph, restricting planning to a finite set of states and transitions.
In contrast, KiTE-Extend operates within a continuous sampling-based framework, using edge bundles only to guide node expansion rather than define the search space.
Candidate motions are re-propagated from the current state and validated under current spatiotemporal constraints, enabling efficient filtering without repeated optimization, learned inference, or feasibility repair.
As a result, KiTE-Extend improves the efficiency of SBP in MRMP while preserving its key properties.


\section{PROBLEM FORMULATION}\label{section_problem_statement}

\paragraph{System Model and Collision Constraints}
We consider a team of $N$ agents indexed by $i\in\{1,\dots,N\}$. 
Agent $i$ has state $x_i(t)\in\mathcal{X}_i\subset\mathbb{R}^{d_i}$  and control input $u_i(t)\in\mathcal{U}_i\subset\mathbb{R}^{m_i}$.
Its dynamics are governed by the differential equation
\begin{equation}
    \dot{x}_i(t) = f_i\!\left(x_i(t),u_i(t)\right).
    \label{eq:dynamics_continuous}
\end{equation}
A \emph{continuous-time} trajectory for agent $i$ is the 3-tuple $\pi_i \triangleq \bigl(x_i(\cdot),u_i(\cdot),T_i\bigr)$ defined over the time horizon $t\in[0,T_i]$, such that $x_i(0)=x_i^{s}$ and $x_i(T_i)\in\mathcal{G}_i$, where $\mathcal{G}_i\subset\mathcal{X}_i$ is a goal region.

Agents move in a workspace $\mathcal{W}\subset\mathbb{R}^{d_w}$ containing static obstacles $\mathcal{O}\subset\mathcal{W}$, where $S_i(x_i)$ is the set of points in workspace occupied by agent $i$ when in state $x_i$.
A trajectory is \emph{collision-free with respect to static obstacles} if it never intersects obstacles:
\begin{equation}
    \forall t\in[0,T_i],\quad S_i(x_i(t)) \cap \mathcal{O} = \emptyset.
    \label{eq:static_collision_free}
\end{equation}
Two agents $i\neq j$ are \emph{pairwise feasible} if they never collide:
\begin{equation}
    \forall t\in[0,T],\quad S_i(x_i(t)) \cap S_j(x_j(t)) = \emptyset,
    \label{eq:pairwise_collision_free}
\end{equation}
where $T \triangleq \max\{T_i,T_j\}$ and we adopt the standard convention that an agent remains at its terminal state after arrival 
(i.e., $x_i(t)=x_i(T_i)$ for $t\ge T_i$).

\paragraph{Translation Invariance}
Many mobile-robot models exhibit symmetry with respect to global translations of the workspace. We assume the following translation-invariance property.
We can decompose each agent state as $x_i = [x_{i,\text{pos}},x_{i,\text{rem}}]$, where $x_{i,\text{pos}}\in\mathbb{R}^{d_w}$ denotes translational position in the workspace and $x_{i,\text{rem}}$ collects the remaining components (e.g., heading, velocities, or internal states).
For any translation $\beta\in\mathbb{R}^{d_w}$, define the translated state
\[
x_i \oplus \beta \;\triangleq\; [\,x_{i,\text{pos}}+\beta,\;x_{i,\text{rem}}\,].
\]
We assume agent $i$ has \emph{translation-invariant dynamics}: for any translation $\beta \in \mathbb{R}^{d_w}$, any admissible control $u_i(\cdot)$ and any solution $x_i(\cdot)$ of \eqref{eq:dynamics_continuous}, the translated trajectory $x_i(\cdot)\oplus \beta$ is also a solution under the same control.
Equivalently, for all translations $\beta \in \mathbb{R}^{d_w}$, all $x_i$ and all $u_i$,
\begin{equation}
f_i(x_i\oplus \beta,\,u_i) \;=\; f_i(x_i,\,u_i).
\label{eq:translation_invariant_dynamics}
\end{equation}

Translation invariance holds for the dynamics of each agent. Obstacles $\mathcal{O}$ and other agents are defined in a fixed world frame, so translating an agent trajectory can change its collisions: a shifted trajectory $x_i(\cdot)\oplus\beta$ can intersect $\mathcal{O}$ or $S_j(x_j(t))$ even when $x_i(\cdot)$ does not.

\paragraph{Objective}
We seek a set of trajectories $\Pi \triangleq \{\pi_i\}_{i=1}^N$ such that each $\pi_i=(x_i(\cdot),u_i(\cdot),T_i)$ starts at $x_i^s$, reaches the goal region $\mathcal{G}_i$, satisfies the dynamics \eqref{eq:dynamics_continuous},
and the team satisfies the collision constraints \eqref{eq:static_collision_free}--\eqref{eq:pairwise_collision_free}.

We study the feasibility problem:
\begin{equation}\label{eq:mrmp_feasibility}
\begin{aligned}
\textbf{find}\quad
& \Pi = \{\pi_i\}_{i=1}^N,\ \ \pi_i=(x_i(\cdot),u_i(\cdot),T_i) \\
\textbf{s.t.}\quad
& \dot{x}_i(t) = f_i(x_i(t),u_i(t)), && \forall\, t\in[0,T_i],\ \forall i \\
& x_i(0)=x_i^s,\ \ x_i(T_i)\in\mathcal{G}_i, && \forall i \\
& S_i(x_i(t))\cap \mathcal{O}=\emptyset, && \forall\, t\in[0,T_i],\ \forall i \\
& S_i(x_i(t))\cap S_j(x_j(t))=\emptyset, && \forall\, t\in[0,T] \\
& &&\forall\, 1\le i<j\le N
\end{aligned}
\end{equation}
where $T \triangleq \max\{T_i,T_j\}$

Once a feasible solution $\Pi$ is found, an additive per-agent cost is computed as follows:
\begin{equation}
    C_i(\pi_i) \triangleq \int_{0}^{T_i} \ell_i(x_i(t),u_i(t))\,dt,
\end{equation}
where $\ell_i(\cdot)\ge 0$ is a user-defined running cost. Team cost is
\begin{equation}
    C(\Pi) \triangleq \sum_{i=1}^N C_i(\pi_i),
\end{equation}


\section{METHODOLOGY}
\label{sec:methodology}

We introduce KiTE-Extend, a planner-agnostic mechanism for accelerating kinodynamic sampling-based motion planning by improving the efficiency of local tree expansions. 
KiTE-Extend does not modify the system dynamics, collision model, or optimality objective of the underlying planner. 
Instead, it augments local node expansion using offline precomputed motion proposals that are validated online under the same feasibility checks as the baseline planner.

\subsection{Sampling-Based Motion Planning}
\label{subsec:sbmp_baseline}

Kinodynamic sampling-based motion planners typically follow the structure described in Algorithm~\ref{alg:sbp-basic}.
They incrementally grow a search tree by repeatedly sampling a target state, selecting a nearby tree node, and attempting short-horizon extensions from that node using randomly sampled control–duration pairs.
While generating candidate control inputs is typically inexpensive, many randomly propagated rollouts either result in collisions, violate state or spatiotemporal constraints, or make poor progress toward the sampled target, especially in interaction-rich MRMP settings.
Consequently, planners often spend substantial computation evaluating large numbers of low-quality candidate expansions.
In this template, KiTE-Extend modifies the local node extension routine $\texttt{Rand-Extend}(\cdot)$.
Instead of naive random control sampling, it proposes a small, ranked set of candidate control–duration pairs drawn from an offline library, while leaving node selection, propagation, and validity checks unchanged.

\subsection{Kinodynamic Translation-Invariant Edge Bundles (KiTE)}
\label{subsec:kite_bundles}

KiTE is built around a library of short, dynamically feasible motion segments, referred to as \emph{edge bundles}, that are generated offline and reused during planning.
Each edge corresponds to rolling out the system dynamics under a constant control input for a finite duration, producing a short trajectory segment and its terminal state.
For translation-invariant systems, these motion segments can be reused across workspace locations while preserving their local dynamical behavior.
This enables the construction of reusable motion libraries that are independent of the global environment geometry.

\begin{algorithm}[t]
    \caption{Sampling Based Planner}
    \label{alg:sbp-basic}
    \DontPrintSemicolon
    \SetVlineSkip{2pt}

    \SetKwInput{KwData}{Input}
    \SetKwInput{KwResult}{Output}

    \SetKwFunction{Sample}{Sample}
    \SetKwFunction{Select}{SelectNode}
    \SetKwFunction{Valid}{IsValid}
    \SetKwFunction{Add}{Add}
    \SetKwFunction{Extract}{ExtractPath}

    \SetKwProg{Proc}{Procedure}{}{}
    \SetKwFunction{R}{Rand}
    \SetKwFunction{Extend}{Extend}
    \SetKwFunction{SampleAction}{SampleAction}
    \SetKwFunction{SampleDuration}{SampleDuration}
    \SetKwFunction{Propagate}{Propagate}
    \SetKwFunction{Dist}{Dist}

    \KwData{start $x_s$, goal region $\mathcal{G}$, free space $\mathcal{X}_{\mathrm{free}}$, budget $N_{\max}$, extension trials $K \ge 1$}
    \KwResult{Trajectory $\pi$ or Infeasible}

    Initialize tree $T$ with root $x_s$\;

    \For{$k=1..N_{\max}$}{
        $x_{\mathrm{rand}} \leftarrow \Sample(\mathcal{X}_{\mathrm{free}})$\;
        $v \leftarrow \Select(T, x_{\mathrm{rand}})$\;
        $(x_{\mathrm{new}}, \tau, ok) \leftarrow \R$-$\Extend(v, x_{\mathrm{rand}}, K)$\; 
        \If{$ok$}{
            \Add($T, x_{\mathrm{new}}, \tau, v$)\;
            \If{$x_{\mathrm{new}} \in \mathcal{G}$}{
                \Return{$\Extract(T, x_{\mathrm{new}})$}\;
            }
        }
    }
    \Return Infeasible\;

    \BlankLine
    \hrule
    \BlankLine

    \Proc{\R-\Extend{$v, x_{\mathrm{rand}}, K$}}{
        $x_v \leftarrow$ state of $v$;\
        $ok \leftarrow \texttt{false}$;\
        $d_{\min} \leftarrow +\infty$\;

        \For{$j=1..K$}{
            $u \leftarrow \SampleAction()$\;
            $\Delta t \leftarrow \SampleDuration()$\;
            $(x', \tau') \leftarrow \Propagate(x_v, u, \Delta t)$\;

            \If{\Valid($\tau'$) \textbf{and} $\Dist(x', x_{\mathrm{rand}}) < d_{\min}$}{
                $d_{\min} \leftarrow \Dist(x', x_{\mathrm{rand}})$\;
                $x_{\mathrm{new}} \leftarrow x'$;\
                $\tau \leftarrow \tau'$;\
                $ok \leftarrow \texttt{true}$\;
            }
        }

        \Return $(x_{\mathrm{new}}, \tau, ok)$\;
    }

\end{algorithm}

\paragraph{Precomputing the edge bundle}
Algorithm~\ref{alg:kite_bundle_gen} is used to construct the edge bundle $\mathcal{B}$ offline.
Each edge is generated from a canonical start state $x_0$ whose positional components are fixed at the origin of the workspace, while the remaining state components are sampled within admissible limits.
A control input $u$ and a duration $T$ are sampled, where $T$ is an integer multiple of $\Delta t$ and $T \leq T_{\max}$.
The system dynamics are propagated under constant control $u$ for duration $T$ to obtain a trajectory segment $\tau$ and terminal state $x_f$.
Only dynamically feasible segments are retained.
Each edge is stored as $(\kappa,u,T,x_f)$, where $\kappa$ denotes the projection of the start state onto its translation-invariant components.
An optional index over stored keys can be constructed to support efficient retrieval during planning.

\begin{algorithm}[t]
\caption{\textsc{KiTE}: Precomputing translation-invariant edges}
\label{alg:kite_bundle_gen}
\DontPrintSemicolon

\KwIn{Time Discretization $\Delta t$;
Maximum duration $T_{\max}$;
Target bundle size $E$
}
\KwOut{Edge bundle $\mathcal{B}$; optional index $\mathcal{I}$}

\tcp{Start states are sampled at the workspace origin
}
$\mathcal{B}\gets\emptyset$;\
$e \gets 0$\;

\While{$e < E$}{
    $x_0 \gets \texttt{SampleStartAtOrigin}()$\;
    $u \gets \texttt{SampleAction}()$\;
    $T \gets \Delta t \cdot \texttt{UniformInt}\!\left(1,\left\lfloor T_{\max}/\Delta t \right\rfloor\right)$\;
    $\tau \gets \texttt{Propagate}(x_0,u,T,\Delta t)$\;
    \If{$\texttt{IsDynFeasible}(\tau,u)$}{
        $x_f \gets \tau[\mathrm{end}]$\;
        $\kappa \gets \texttt{Key}(x_0)$\;
        $\mathcal{B} \gets \mathcal{B}\cup\{(\kappa,u,T,x_f)\}$\;
        $e \gets e+1$\;
    }
}

\tcp{Optional: build a retrieval structure over stored keys (e.g., KD-tree).}
$\mathcal{I} \gets \texttt{BuildIndex}(\{\kappa \mid (\kappa,u,T,x_f)\in\mathcal{B}\})$\;
\Return $\mathcal{B},\mathcal{I}$\;
\end{algorithm}

\paragraph{Retrieval, ranking, and fallback during node expansion}
Upon the first expansion of a node $x_{\mathrm{near}}$, KiTE-Extend retrieves an edge set by querying the edge bundle.
Specifically, edges are selected whose stored keys $\kappa$ lie within a $\delta$-radius neighborhood of $\kappa(x_{\mathrm{near}})$, yielding candidate edges whose starting states have translation-invariant components similar to those of $x_{\mathrm{near}}$.
The retrieved edge set is stored locally for each node and reused across future expansions from that node.
At each expansion step, the available candidates are ranked using a heuristic that favors progress toward the sampled target $x_{\mathrm{rand}}$, e.g., based on the distance between the stored terminal state $x_f$ and $x_{\mathrm{rand}}$.
Edges are then attempted in ranked order using a stride-$p$ traversal of the ranked list.
Each attempt consists of re-propagation under $(u,T)$ followed by validity checking (see Fig.~\ref{fig:edge_bundle_selection}).
Whenever an edge is attempted, it is removed from the node's candidate set and is not reconsidered for future expansions from the same node.

\paragraph{Preservation of theoretical properties}
KiTE-Extend preserves the exploration structure of the underlying sampling-based planner.
State sampling, node selection, and nearest-neighbor queries are unchanged, and only the local extension routine is modified to prioritize candidate motions drawn from the edge bundle.
Candidate edges retrieved for a node are finite and monotonically decrease as they are attempted and removed from the node's candidate set.
If no retrieved candidate yields a valid extension, or if no retrieved candidates remain for the node, KiTE-Extend falls back to the baseline random extension step.
Additionally, with probability $\epsilon$, KiTE-Extend performs a random extension regardless of bundle availability.
As a result, every node retains a non-zero probability of sampling from the full admissible control space, preserving the exploration behavior and theoretical properties of the underlying planner.


\begin{algorithm}[t]
\caption{\texttt{KiTE}-\texttt{Extend}}
\label{alg:kite_extend}
\DontPrintSemicolon

\KwIn{$x_{\mathrm{near}}$; sampled target $x_{\mathrm{rand}}$; edge bundle $\mathcal{B}$; index $\mathcal{I}$; radius $\delta$; skip factor $p$; random fallback probability $\epsilon$}
\KwOut{Accepted extension $(x_{\mathrm{new}},\tau)$ or \texttt{FAIL}}

\tcp{$\epsilon$-greedy random expansion.}
\If{$\texttt{Uniform}(0,1) < \epsilon$}{
    \Return $\R$-$\Extend(x_{\mathrm{near}},x_{\mathrm{rand}},K{=}1)$\;
}

\tcp{Initialize the node-local available edge set on first expansion of $x_{\mathrm{near}}$.}
\If{$\mathcal{C}(x_{\mathrm{near}})$ is undefined}{
    $\kappa_{\mathrm{near}} \gets \texttt{Key}(x_{\mathrm{near}})$\;
    $\mathcal{C}(x_{\mathrm{near}}) \gets \{e \in \mathcal{B} \mid \|\kappa(e)-\kappa_{\mathrm{near}}\|\le \delta\}$ \tcp*[r]{using $\mathcal{I}$}
}

\tcp{If no retrieved edges are available, use random expansion.}
\If{$\mathcal{C}(x_{\mathrm{near}})=\emptyset$}{
    \Return $\R$-$\Extend(x_{\mathrm{near}},x_{\mathrm{rand}},K{=}1)$\;
}

\tcp{Rank by predicted progress toward $x_{\mathrm{rand}}$ using stored endpoints.}
$\mathcal{C}_{\mathrm{rank}} \gets \texttt{SortBy}(\mathcal{C}(x_{\mathrm{near}}), \; d(x_f(e),x_{\mathrm{rand}}))$\;

\For{$j \gets 1$ \KwTo $\left\lceil |\mathcal{C}_{\mathrm{rank}}|/p \right\rceil$}{
    $e \gets \mathcal{C}_{\mathrm{rank}}[\,1 + (j-1)p\,]$\;
    $\mathcal{C}(x_{\mathrm{near}}) \gets \mathcal{C}(x_{\mathrm{near}}) \setminus \{e\}$ \tcp*[r]{Mark $e$ unavailable for future expansions from $x_{\mathrm{near}}$.}
    $(u,T) \gets \texttt{EdgeParams}(\mathcal{B}, e)$\;
    $(x_{\mathrm{new}},\tau) \gets \texttt{Propagate}(x_{\mathrm{near}},u,T,\Delta t)$\;
    \If{$\texttt{IsValid}(\tau)$}{
        \Return $(x_{\mathrm{new}},\tau)$\;
    }
}

\tcp{Fallback if no stride-$p$ candidate from $\mathcal{C}_{\mathrm{rank}}$ succeeds.}
\Return $\R$-$\Extend(x_{\mathrm{near}},x_{\mathrm{rand}},K{=}1)$\;
\end{algorithm}

\subsection{Single-Agent Instantiation}
\label{subsec:kite_single_agent}

KiTE-Extend plugs into any kinodynamic sampling-based planner at the point where a short-horizon control--duration pair is proposed for local expansion.
Given an expansion state $x_{\mathrm{near}}$ and a guidance signal (e.g., a sampled target $x_{\mathrm{rand}}$), KiTE-Extend (Algorithm~\ref{alg:kite_extend}) retrieves and ranks candidate motions from the offline bundle, and returns the first valid propagated segment.
The guidance signal is planner-dependent and is used only to rank candidate edges before propagation.
In RRT-style planners, the sampled target $x_{\mathrm{rand}}$ naturally serves this role by ranking edges based on the proximity of their predicted endpoints.
In cost-aware planners, the same mechanism can rank candidates using the planner’s existing local cost or progress metric evaluated at the predicted endpoint.

\subsection{Multi-Robot Instantiations}
\label{subsec:multi_robot_instantiations}

We integrate the same low-level KiTE-Extend primitive into three representative multi-robot motion planning paradigms.
In each case, only the mechanism by which agents are coupled differs; the underlying KiTE-based candidate generation remains unchanged.

\begin{figure}
    \centering
    \begin{subfigure}[t]{\linewidth}
        \centering
        \includegraphics[width=\linewidth]{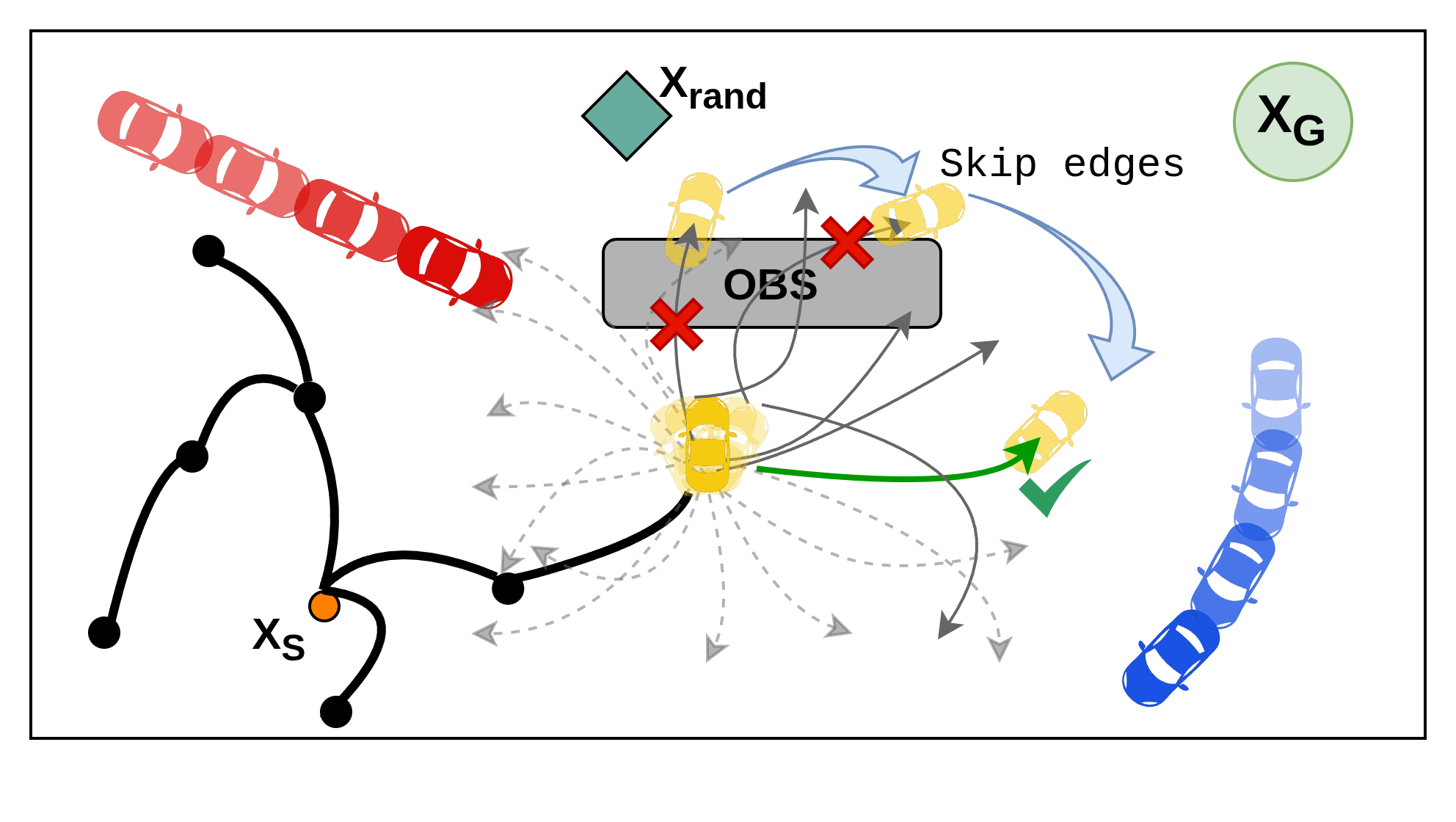}
        \label{fig:kite_extend_method}
    \end{subfigure}
    
    \vspace{-2.0\baselineskip}
    
    \begin{subfigure}[t]{\linewidth}
        \centering
        \includegraphics[width=\linewidth]{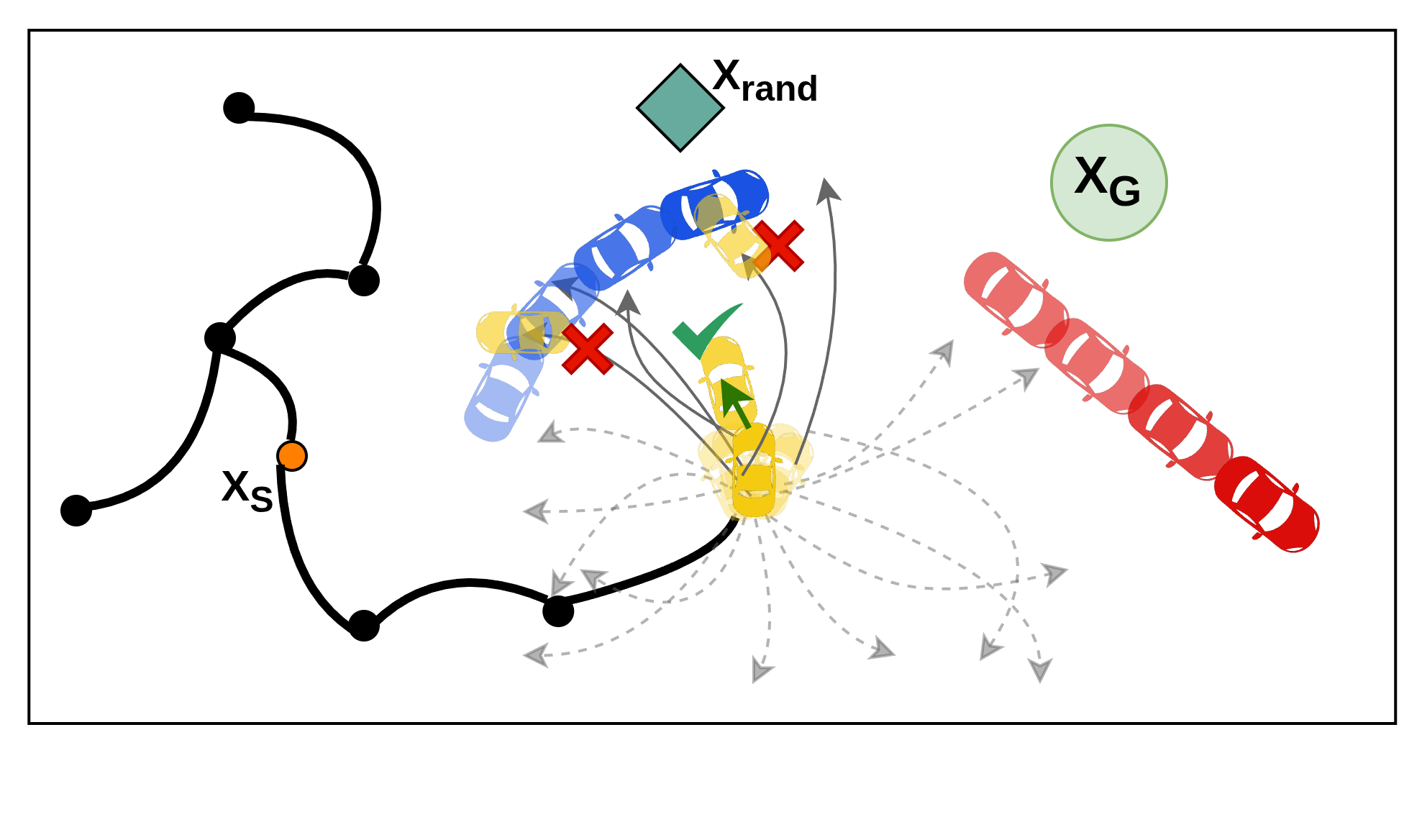}
        \label{fig:extend_short_arrow}
    \end{subfigure}

    \vspace{-2.0\baselineskip}    
\caption{\small
\textbf{Ranked edge-bundle expansion in KiTE-Extend.}
\textit{Top:} From a given tree node, KiTE ranks precomputed, dynamically feasible trajectory segments by the distance of their endpoints to the sampled state $x_{\mathrm{rand}}$ and attempts them in order, skipping infeasible edges caused by obstacles.
\textit{Bottom:} Under agent collision constraints, the same mechanism naturally favors safe alternatives, including near-zero-velocity trajectories that avoid conflicts.
Yellow vehicles denote the controlled agent, red and blue vehicles denote dynamic obstacles, and faint yellow states indicate candidate edge-bundle start configurations sampled within a $\delta$-ball around the current node.
}

    \label{fig:edge_bundle_selection}
\end{figure}

\begin{figure*}[t!]
    \centering
    \begin{subfigure}[t]{0.18\textwidth}
        \centering
        \includegraphics[width=\linewidth]{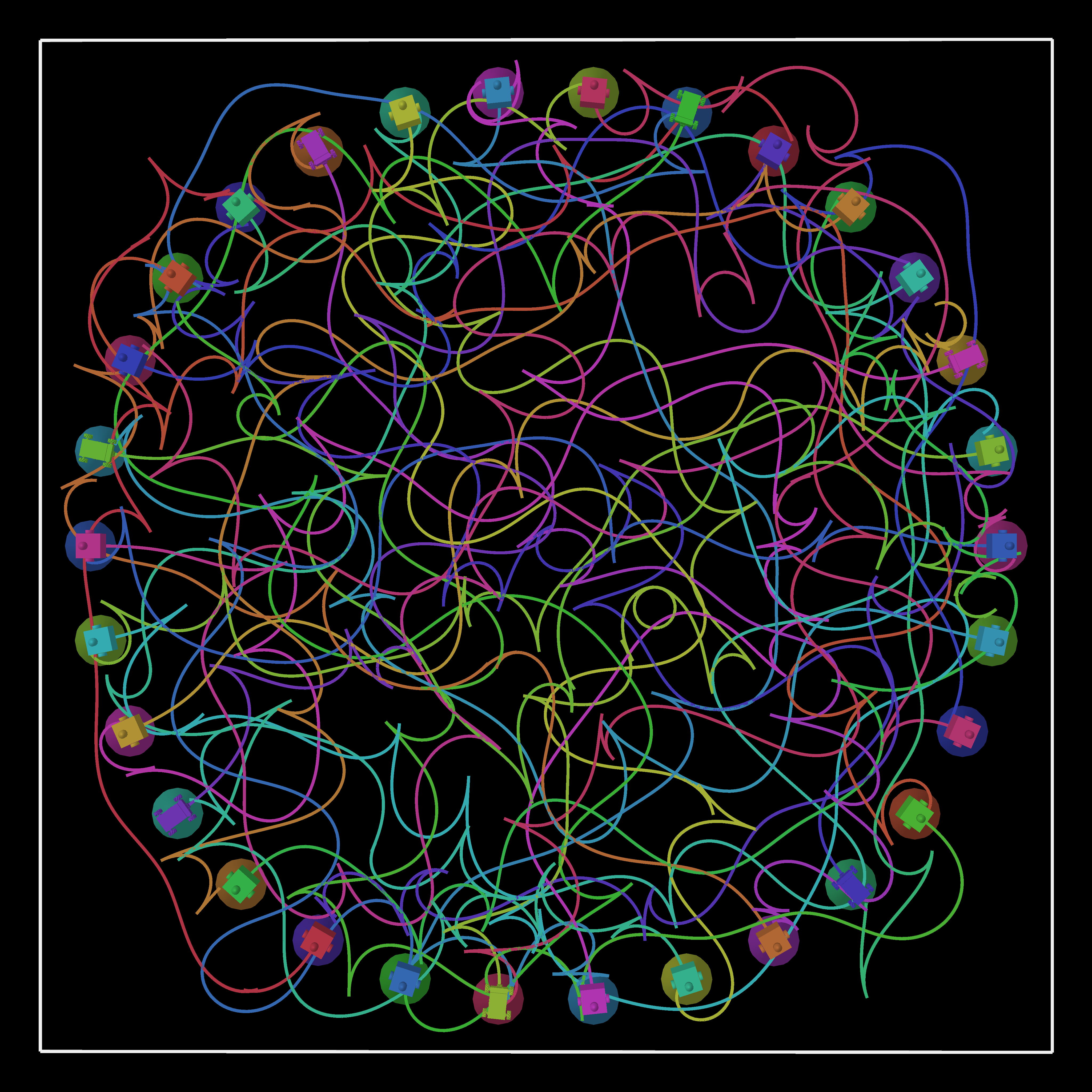}
        \caption{\small Swap\\(20$\times$20)}
        \label{fig_domain1}
    \end{subfigure}%
    \hfill
    \begin{subfigure}[t]{0.18\textwidth}
        \centering
        \includegraphics[width=\linewidth]{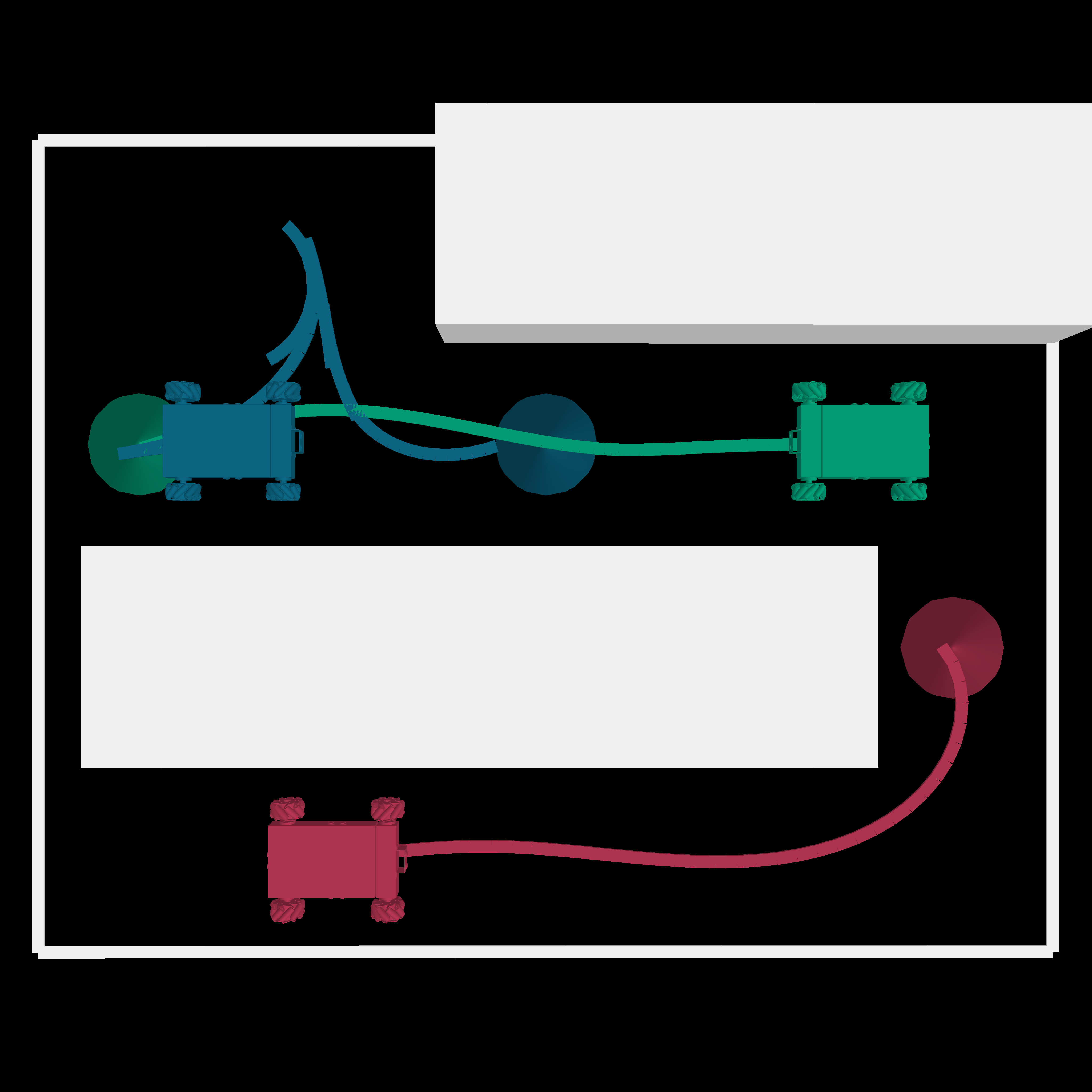}
        \caption{\small Narrow\\(5$\times$4)}
        \label{fig_domain2}
    \end{subfigure}%
    \hfill
    \begin{subfigure}[t]{0.18\textwidth}
        \centering
        \includegraphics[width=\linewidth]{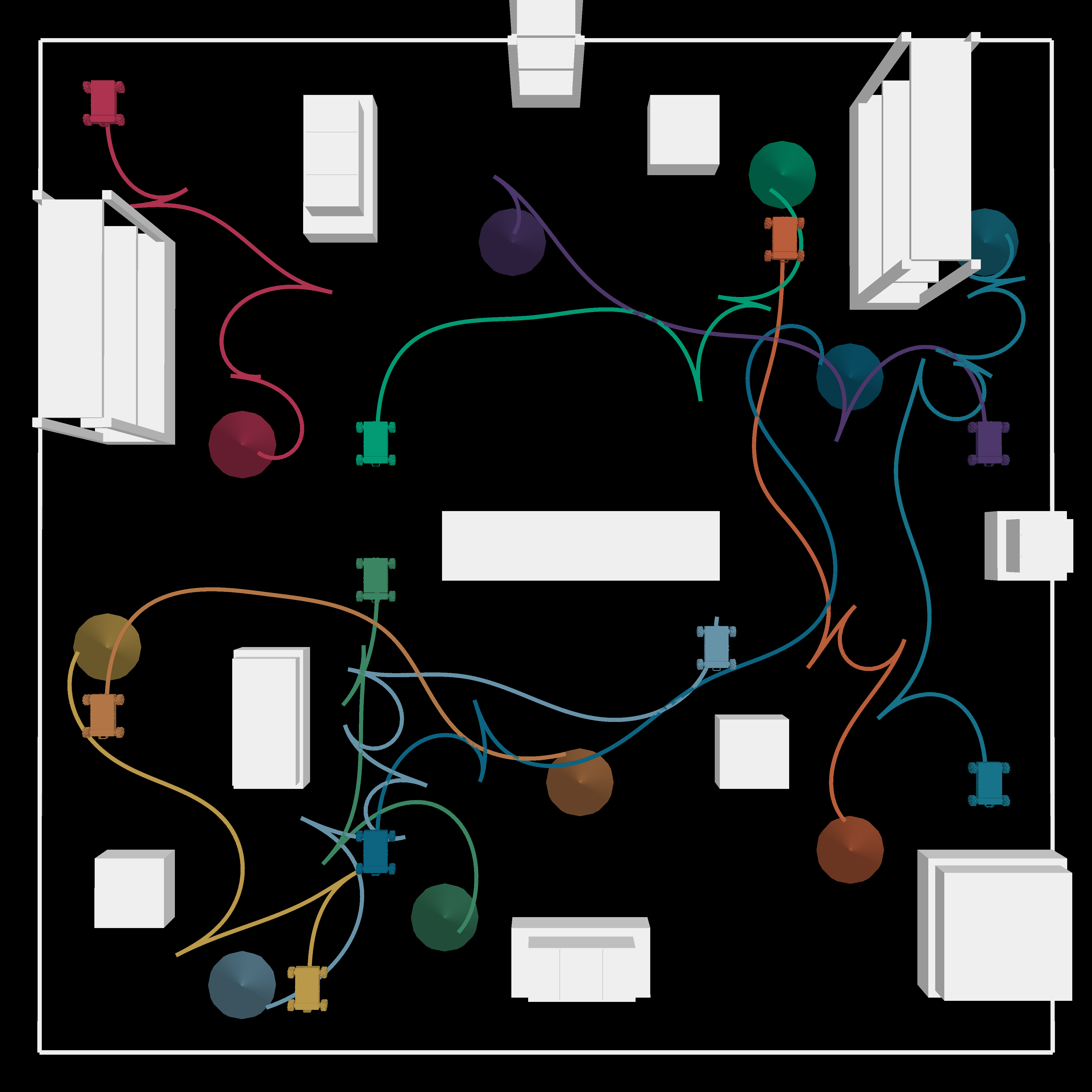}
        \caption{\small Small Cluttered (15$\times$15)}
        \label{fig_domain3}
    \end{subfigure}%
    \hfill
    \begin{subfigure}[t]{0.18\textwidth}
        \centering
        \includegraphics[width=\linewidth]{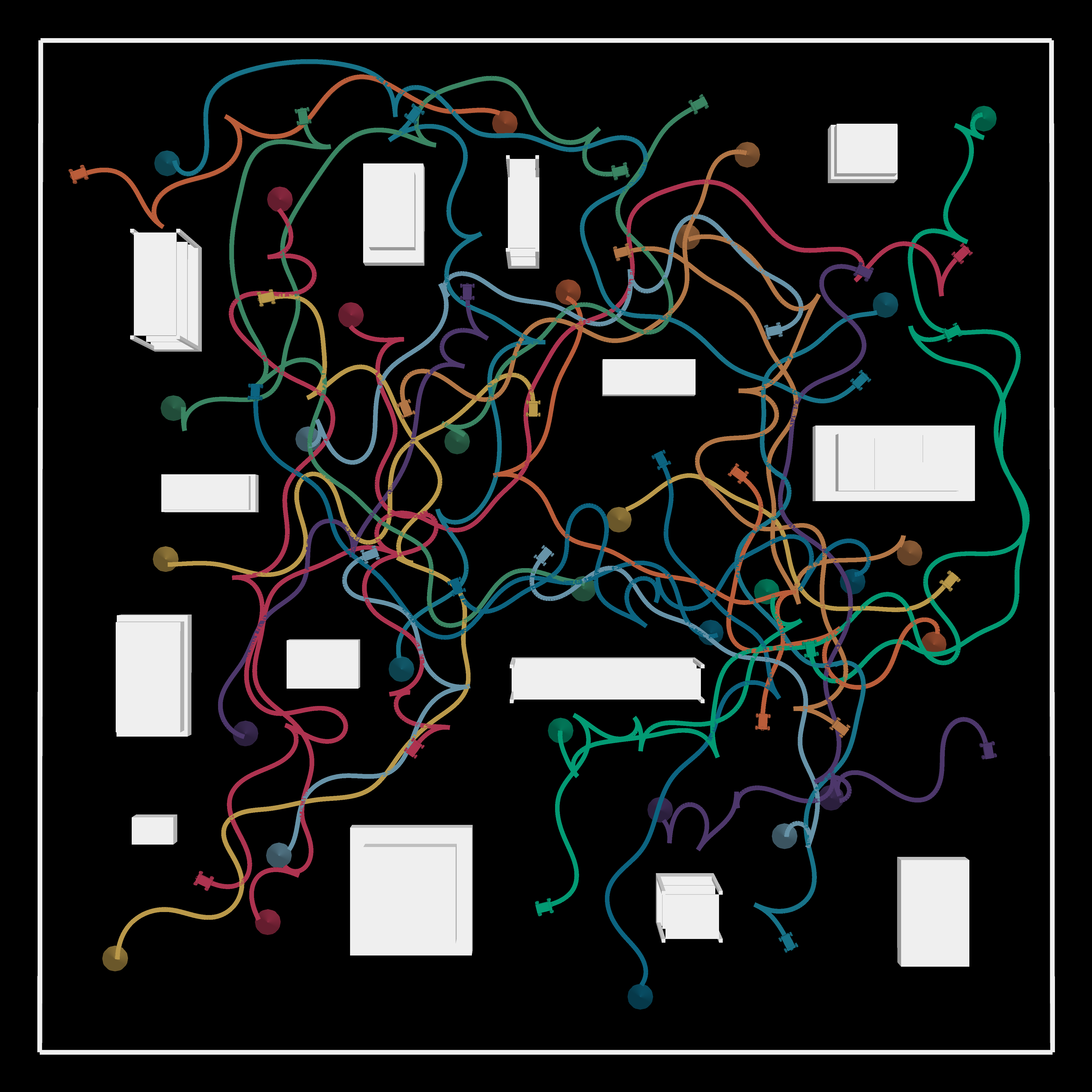}
        \caption{\small Large Cluttered (40$\times$40)}
        \label{fig_domain4}
    \end{subfigure}%
    \hfill
    \begin{subfigure}[t]{0.18\textwidth}
        \centering
        \includegraphics[width=\linewidth]{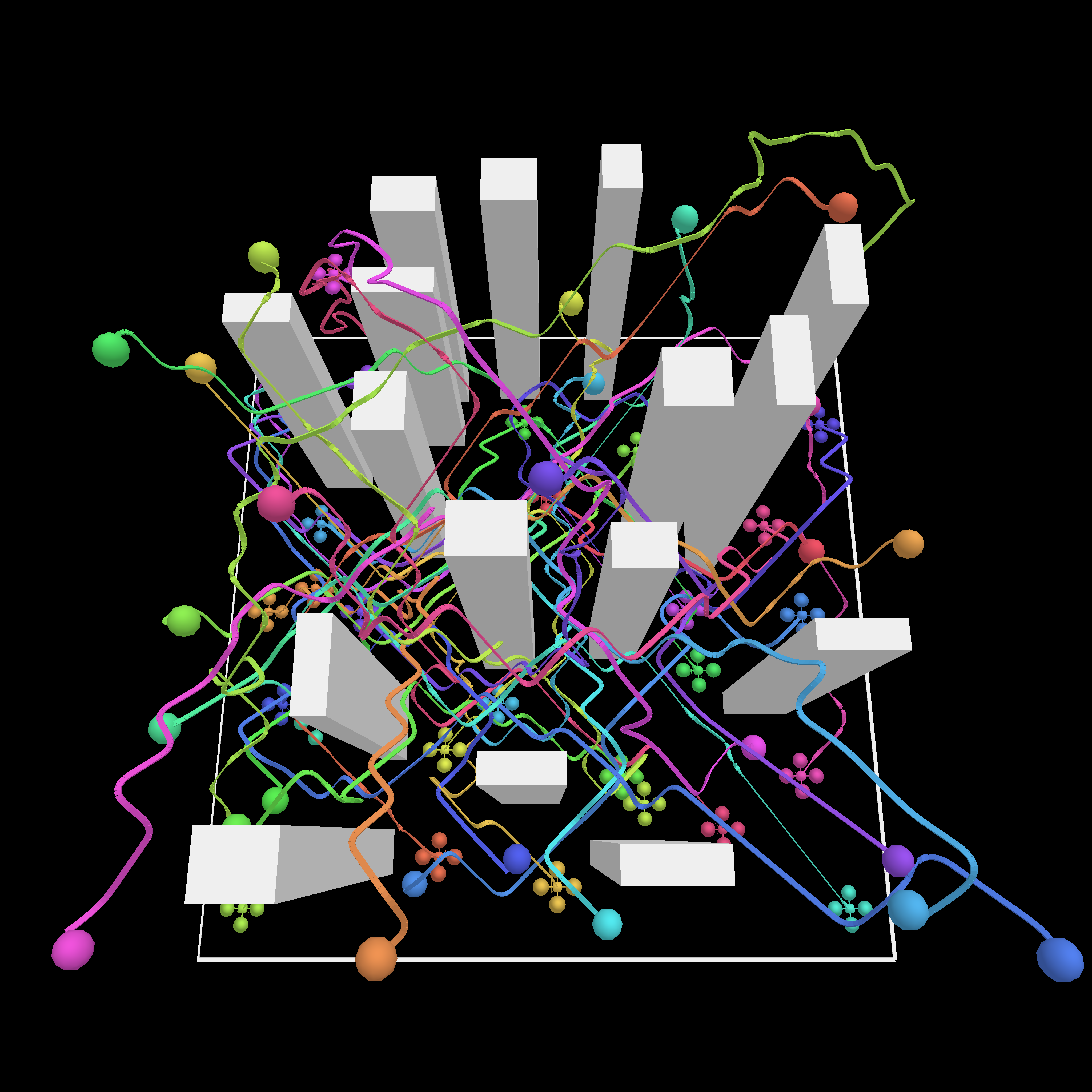}
        \caption{\small Large Cluttered 3D (14$\times$14$\times$10)}
        \label{fig_domain5}
    \end{subfigure}

    \caption{
    \small
    Benchmark environments. The goal region of each vehicle is a circle/sphere of the same color.
    }
    \label{fig_simulation_enviroments}
\end{figure*}

\subsubsection{Centralized Joint-State Planning}
\label{subsec:kite_centralized}

In centralized planning, the search tree is constructed in the joint state space $\mathcal{X}_1 \times \cdots \times \mathcal{X}_N$.
During a joint expansion, KiTE-Extend is invoked independently for each agent from its current component state, yielding candidate controls and durations $(u_i,T_i)$.
Since the proposed durations may differ across agents, we synchronize the joint rollout by propagating all agents for
$T_{\min} \triangleq \min_i T_i$, truncating longer segments accordingly and evaluating interactions at the shared discrete time instants.
The joint extension is accepted only if all truncated agent segments are individually valid and no collisions occur between any pair of agents at any discrete time $t \in [0, T_{\min}]$.




\subsubsection{Prioritized Planning}
\label{subsec:kite_prioritized}
Here, agents are planned sequentially under a fixed priority ordering.
When planning for agent $i$, the committed trajectories of agents $1,\ldots,i-1$ are treated as time-indexed dynamic obstacles.
During propagation, candidate segments proposed by KiTE-Extend with duration $T$ are accepted only if they avoid both static obstacles and higher-priority agents at all discrete times $t \in [0,T]$.

\subsubsection{Conflict-Based Search}
\label{subsec:kite_cbs}

In CBS-style planners~\cite{kottingerConflictBasedSearchMultiRobot2022}, coordination is enforced by adding spatiotemporal constraints and replanning only the affected agent.
We use KiTE-Extend as the low-level expansion primitive when replanning a single agent under a given constraint set.
Candidate segments are retrieved and ranked as in the unconstrained case, but are rejected if any state along the propagated segment violates an active constraint at the corresponding discrete time.
Since CBS repeatedly replans at the low level, improving local node expansion efficiency has an outsized impact in this setting.

\emph{Optional pruning and reuse:}
Replanning from scratch at every branch of the conflict tree is simple but can be wasteful, as large portions of a previously explored low-level tree may remain valid after adding a new constraint.
As an engineering optimization, we optionally reuse portions of a previously explored low-level tree after adding a new constraint.
Given a conflict interval $[t_s,t_f]$, we discard the subtree downstream of the first edge whose time span overlaps $t_s$.
All remaining edges are revalidated against the new constraint, and expansion resumes from the remaining tree. 
This optimization reduces repeated low-level planning effort during high-level branching without altering feasibility semantics.


\section{Experiments} \label{section_experiments}

\subsection{Kinodynamic Systems and Environments}

We consider three different models of kinodynamic systems and match our robot parameters to those used in \cite{kottingerConflictBasedSearchMultiRobot2022,moldagalieva2024db}.

\textbf{Unicycle (UC, 2D):}
Circular robot with state $(x,y,\theta)$, controls $(v,\omega)$ and dynamics
$\dot{x}=v\cos\theta$, $\dot{y}=v\sin\theta$, $\dot{\theta}=\omega$.
Control bounds are $v \in [-0.5, 0.5]\ \mathrm{m/s}$ and
$\omega \in [-0.5, 0.5]\ \mathrm{rad/s}$ and radius is $r=0.4\ m$.

\textbf{Second-Order Car (SOC, 2D):}
Rectangular robot with state $(x,y,\theta,v,\phi)$, controls $(a,\omega)$ and dynamics
$\dot{x}=v\cos\theta$, $\dot{y}=v\sin\theta$,
$\dot{\theta}=\frac{v}{L}\tan\phi$, $\dot{v}=a$, $\dot{\phi}=\omega$.
Bounds are $v \in [-1, 1]\ \mathrm{m/s}$, $\phi \in [-\pi/3, \pi/3]\ \mathrm{rad}$,  $a \in [-2, 2]\ \mathrm{m/s^2}$ and
$\omega \in [-0.5, 0.5]\ \mathrm{rad/s}$ with length $l=0.7\ m$ and width $w= 0.4\ m$.

\textbf{Double Integrator (DI, 3D):}
Spherical robot with state $\mathbf{x}=[\mathbf{p}^\top,\mathbf{v}^\top]^\top$ where $\mathbf{p}=(p_x,p_y,p_z)$, $\mathbf{v}=(v_x,v_y,v_z)$, controls
$\mathbf{u}=(a_x,a_y,a_z)$ and dynamics
$\dot{\mathbf{p}}=\mathbf{v}$, $\dot{\mathbf{v}}=\mathbf{u}$.
Velocity and accelerations are bounded as $v_{x,y,z} \in [-0.5,\,0.5]\ \mathrm{m/s}$, $a_{x,y,z} \in [-2.0,\,2.0]\ \mathrm{m/s^2}$ and radius is $r=0.1\ m$.

We evaluate the planning performance of the systems operating in 2D space across the first four environments shown in Fig.~\ref{fig_simulation_enviroments}.
Each environment emphasizes distinct challenges in MRMP.
Swap forces agents to exchange destinations in a confined area, and Narrow Corridor requires explicit ordering and waiting behavior. 
Both small and large cluttered environments require avoiding inter-agent collisions and collisions with arbitrarily placed static obstacles.
For the DI model, we evaluate planning performance in 3D versions of the Swap and Large Cluttered environments.

\begin{table*}[!ht]
\caption{\small Representative results across systems and environments at two team sizes per setting. Full results (510 conditions) are reported in Appendix.
}
\begin{center}

\resizebox{1\linewidth}{!}{%
\begin{tabular}{l|c|ccc|ccc|ccc|ccc}
\toprule
\textbf{Domain} & \textbf{Variant}
& \multicolumn{3}{c|}{\textbf{cRRT}}
& \multicolumn{3}{c|}{\textbf{pRRT}}
& \multicolumn{3}{c|}{\textbf{KCBS}}
& \multicolumn{3}{c}{\textbf{dbCBS}} \\

& 
& SR (\%) & CT (s) & PT (s)
& SR (\%) & CT (s) & PT (s)
& SR (\%) & CT (s) & PT (s)
& SR (\%) & CT (s) & PT (s) \\[1mm]
\hline
\hline

{Small\;Cluttered} & Base & \textbf{100} & 13.2 $\pm$ 1.2 & 428 $\pm$ 8.0 & \textbf{100} & 1.34 $\pm$ 0.11 & 132 $\pm$ 3.1 & \textbf{100} & 2.17 $\pm$ 0.16 & 116 $\pm$ 2.0 & 0 & - & - \\
{SOC (N=4)} & KiTE & \textbf{100} & \textbf{0.639 $\pm$ 0.049} & \textbf{223 $\pm$ 5.5} & \textbf{100} & \textbf{0.258 $\pm$ 0.019} & \textbf{102 $\pm$ 1.8} & \textbf{100} & \textbf{0.479 $\pm$ 0.042} & \textbf{87.6 $\pm$ 1.4} & - & - & - \\
\hline

{Small\;Cluttered} & Base & 1 & 221 $\pm$ 0.0 & 1668 $\pm$ 0.0 & 95 & 7.27 $\pm$ 1.2 & 278 $\pm$ 5.3 & \textbf{100} & 7.13 $\pm$ 0.39 & 188 $\pm$ 2.4 & 0 & - & - \\
{SOC (N=8)} & KiTE & \textbf{93} & \textbf{13.0 $\pm$ 2.3} & \textbf{643 $\pm$ 16} & \textbf{97} & \textbf{5.68 $\pm$ 2.1} & \textbf{219 $\pm$ 3.5} & \textbf{100} & \textbf{1.72 $\pm$ 0.12} & \textbf{146 $\pm$ 1.6} & - & - & - \\
\hline

{Small\;Cluttered} & Base & 0 & - & - & 73 & 50.4 $\pm$ 5.5 & 632 $\pm$ 11 & 73 & 141 $\pm$ 9.2 & 290 $\pm$ 2.6 & 0 & - & - \\
{SOC (N=15)} & KiTE & \textbf{33} & \textbf{244 $\pm$ 8.1} & \textbf{2211 $\pm$ 71} & \textbf{79} & \textbf{26.0 $\pm$ 4.5} & \textbf{504 $\pm$ 6.7} & \textbf{99} & \textbf{43.2 $\pm$ 4.1} & \textbf{250 $\pm$ 1.9} & - & - & - \\
\hline

{Large\;Cluttered} & Base & 95 & 82.4 $\pm$ 6.1 & 2476 $\pm$ 23 & 87 & 4.67 $\pm$ 1.2 & 644 $\pm$ 8.0 & \textbf{100} & 3.89 $\pm$ 0.22 & 588 $\pm$ 6.0 & 100 & 96.5 $\pm$ 0.26 & 266 $\pm$ 0.0 \\
{UC (N=5)} & KiTE & \textbf{100} & \textbf{2.97 $\pm$ 0.18} & \textbf{853 $\pm$ 13} & \textbf{89} & \textbf{2.84 $\pm$ 1.9} & \textbf{379 $\pm$ 4.1} & \textbf{100} & \textbf{1.23 $\pm$ 0.097} & \textbf{340 $\pm$ 3.1} & - & - & - \\
\hline

{Large\;Cluttered} & Base & 0 & - & - & 6 & 30.3 $\pm$ 4.1 & 2557 $\pm$ 132 & 90 & 106 $\pm$ 6.9 & 1753 $\pm$ 10 & 0 & - & - \\
{UC (N=20)} & KiTE & \textbf{43} & \textbf{248 $\pm$ 4.3} & \textbf{7353 $\pm$ 146} & \textbf{13} & \textbf{28.3 $\pm$ 9.1} & \textbf{1477 $\pm$ 30} & \textbf{100} & \textbf{38.7 $\pm$ 3.8} & \textbf{1070 $\pm$ 4.6} & - & - & - \\
\hline

{Large\;Cluttered} & Base & 0 & - & - & 4 & 98.3 $\pm$ 26 & 4284 $\pm$ 64 & 4 & 237 $\pm$ 28 & 2702 $\pm$ 44 & 0 & - & - \\
{UC (N=30)} & KiTE & 0 & - & - & \textbf{6} & \textbf{17.4 $\pm$ 2.9} & \textbf{2390 $\pm$ 81} & \textbf{65} & \textbf{140 $\pm$ 7.5} & \textbf{1637 $\pm$ 3.6} & - & - & - \\
\hline

{Narrow\;Corridor} & Base & \textbf{100} & 2.40 $\pm$ 0.17 & 192 $\pm$ 4.4 & 26 & 51.9 $\pm$ 12 & 110 $\pm$ 5.6 & 27 & 81.4 $\pm$ 15 & 96.9 $\pm$ 5.1 & 100 & 5.08 $\pm$ 0.033 & 28.6 $\pm$ 0.048 \\
{UC (N=3)} & KiTE & \textbf{100} & \textbf{2.33 $\pm$ 0.17} & \textbf{140 $\pm$ 4.8} & \textbf{34} & \textbf{34.0 $\pm$ 8.6} & \textbf{81.2 $\pm$ 3.3} & \textbf{53} & \textbf{36.1 $\pm$ 7.1} & \textbf{74.6 $\pm$ 3.2} & - & - & - \\
\hline

{Swap 3D} & Base & 93 & 27.9 $\pm$ 4.1 & 313 $\pm$ 4.1 & \textbf{100} & 0.0549 $\pm$ 0.0012 & 69.4 $\pm$ 0.52 & \textbf{100} & 0.0632 $\pm$ 0.0012 & 70.7 $\pm$ 0.50 & 60 & 290 $\pm$ 0.69 & 64.6 $\pm$ 0.0 \\
{DI (N=5)} & KiTE & \textbf{100} & \textbf{0.115 $\pm$ 0.0031} & \textbf{140 $\pm$ 1.9} & \textbf{100} & \textbf{0.0160 $\pm$ 0.00030} & \textbf{64.1 $\pm$ 0.50} & \textbf{100} & \textbf{0.0272 $\pm$ 0.0084} & \textbf{64.8 $\pm$ 0.45} & - & - & - \\
\hline

{Swap 3D} & Base & 0 & - & - & \textbf{100} & 0.269 $\pm$ 0.0073 & 253 $\pm$ 1.4 & \textbf{100} & 0.275 $\pm$ 0.0047 & 251 $\pm$ 0.97 & 0 & - & - \\
{DI (N=20)} & KiTE & \textbf{100} & \textbf{3.57 $\pm$ 0.060} & \textbf{989 $\pm$ 12} & \textbf{100} & \textbf{0.0635 $\pm$ 0.0017} & \textbf{229 $\pm$ 0.90} & \textbf{100} & \textbf{0.0639 $\pm$ 0.00081} & \textbf{226 $\pm$ 0.89} & - & - & - \\
\hline

{Swap 3D} & Base & 0 & - & - & \textbf{100} & 0.473 $\pm$ 0.012 & 421 $\pm$ 2.1 & \textbf{100} & 0.461 $\pm$ 0.0066 & 409 $\pm$ 1.4 & 0 & - & - \\
{DI (N=30)} & KiTE & \textbf{100} & \textbf{11.7 $\pm$ 0.17} & \textbf{1864 $\pm$ 20} & \textbf{100} & \textbf{0.0989 $\pm$ 0.00082} & \textbf{375 $\pm$ 1.2} & \textbf{100} & \textbf{0.111 $\pm$ 0.0013} & \textbf{373 $\pm$ 1.1} & - & - & - \\
\hline

{Large\;Cluttered} & Base & 0 & - & - & \textbf{100} & 4.55 $\pm$ 0.46 & 1171 $\pm$ 3.9 & \textbf{100} & 5.60 $\pm$ 0.66 & 831 $\pm$ 2.8 & 0 & - & - \\
{DI (N=30)} & KiTE & 0 & - & - & \textbf{100} & \textbf{1.51 $\pm$ 0.25} & \textbf{780 $\pm$ 3.2} & \textbf{100} & \textbf{2.02 $\pm$ 0.39} & \textbf{768 $\pm$ 2.8} & - & - & - \\
\hline

\bottomrule
\end{tabular}
}

\vspace{0.5ex}
{\scriptsize SR: Success Rate \textbar{} CT: Computation Time \textbar{} PT: Total Path Time}
\label{table_final_shortlisted_results}
\end{center}
\end{table*}

\subsection{Baselines}

We compare four MRMP approaches.
Three are sampling-based and differ only in their high-level coordination
strategy, while sharing a common sampling-based planning framework.
The fourth represents a fundamentally different search-and-repair approach.
For each sampling-based MRMP method, we evaluate two variants that differ only
in their node expansion mechanism.
The first uses a standard sampling-based expansion scheme, in which $k$ random
control rollouts are generated, and the rollout terminating closest to the
randomly sampled target state is selected (Alg.~\ref{alg:sbp-basic}).
The second replaces this expansion step with KiTE-Extend (Alg.~\ref{alg:kite_extend}), while keeping all other components unchanged.
Although KiTE-Extend is agnostic to the underlying sampling-based planner, we instantiate it within the RRT algorithm~\cite{lavalle2001randomized} in this work.

\paragraph{Centralized joint-state planning}
\textbf{cRRT} plans in the joint state space using the RRT algorithm.
\textbf{cRRT+KiTE} differs only in the use of KiTE for joint-state node expansion.

\paragraph{Prioritized planning}
\textbf{pRRT} performs sequential planning under a fixed global priority ordering, using RRT for each agent.
\textbf{pRRT+KiTE} replaces the per-agent node expansion step with KiTE.
                          
\paragraph{Conflict-Based Search}
\textbf{K-CBS} is a probabilistically complete conflict-based search planner that uses RRT for low-level replanning under constraints.
\textbf{K-CBS+KiTE} replaces the low-level expansion mechanism with KiTE in RRT while keeping the high-level CBS logic unchanged.

\paragraph{Discontinuity-bounded CBS (dbCBS)}
While it is not a SBP approach, \textbf{dbCBS} is a competing probabilistically complete baseline. It utilizes translation-invariant precomputations to speed up search before solving a joint optimization for ensuring feasibility~\cite{moldagalieva2024db}.
For improved performance, we use db-A$^*_\omega$ instead of db-A$^*$ as the low-level planner within dbCBS, following the updated formulation in~\cite{moldagalieva2025dbecbs}.

\subsection{Experimental Details}

Each experimental setting is defined by a choice of environment, kinodynamic system, and number of robots.
For each setting, we run 100 independent trials with different random seeds, using the same set of seeds across the three sampling-based baselines.
The number of robots varies from 4 to 30, depending on the environment.
All experiments were run on a AMD Ryzen Threadripper PRO 7985WX system (64 cores, 515 GB RAM) using a single-threaded implementation.

Each planner is given a maximum runtime budget of 300 seconds per trial.
A trial is considered successful if the planner produces a set of dynamically feasible, collision-free trajectories for all robots within the allotted time.
We report (i) success rate (SR), (ii) total path computation time (CT), and (iii) solution cost measured as the sum of individual robot trajectory durations (PT).
Reported computation times include only online planning time, and averages for time and cost metrics are computed only over successful trials.

For planners using KiTE-Extend, edge bundles are precomputed once per kinodynamic system and are environment agnostic.
Edge bundle generation takes approximately 5 to 10 seconds per system and is excluded from reported runtimes.
We use translation-invariant edge bundles of size 30k for the Unicycle, 50k for the Second-Order Car, and 100k for the Double Integrator.
An ablation study on the effect of edge bundle size is provided in the Appendix.
In addition to translational invariance, we explored rotational symmetry for UC and SOC, which required significantly smaller edge bundles to achieve comparable performance.
These results are omitted to maintain consistency with the dbCBS comparison setting, which does not exploit such symmetries.
For fairness, we use the same maximum number of propagations per expansion step in both extension routines.
Specifically, KiTE-Extend dynamically sets the skip factor to
$p=\lceil |\mathcal{C}(x_{\mathrm{near}})|/10 \rceil$
at each expansion  with $\epsilon=0.01$, while Rand-Extend uses a rollout count of $K=10$.

We implemented all sampling-based baselines within a common Python codebase, using Numba just-in-time compilation to eliminate interpreter overhead.
For dbCBS, we use the authors’ publicly available implementation \cite{dbcbs-github}.
To ensure compatibility and fair comparison, our implementations match the collision-checking scheme, kinodynamic system parameters, and time discretization used by dbCBS, allowing us to reuse the motion primitives and solver they provide without modification.
Trajectories are generated using fourth-order Runge-Kutta integration with a fixed time step $\Delta t = 0.1$, and collision checking is performed at this temporal resolution across all planners.
For fair comparison with dbCBS, which resolves conflicts without agent merging, we evaluate K-CBS with merging disabled.
Since the released dbCBS implementation supports only Unicycle and Double Integrator systems, comparisons with dbCBS are reported for those systems only.


\section{RESULTS}

\begin{figure*}[!ht]
    \centering
    \includegraphics[width=0.1125\linewidth]{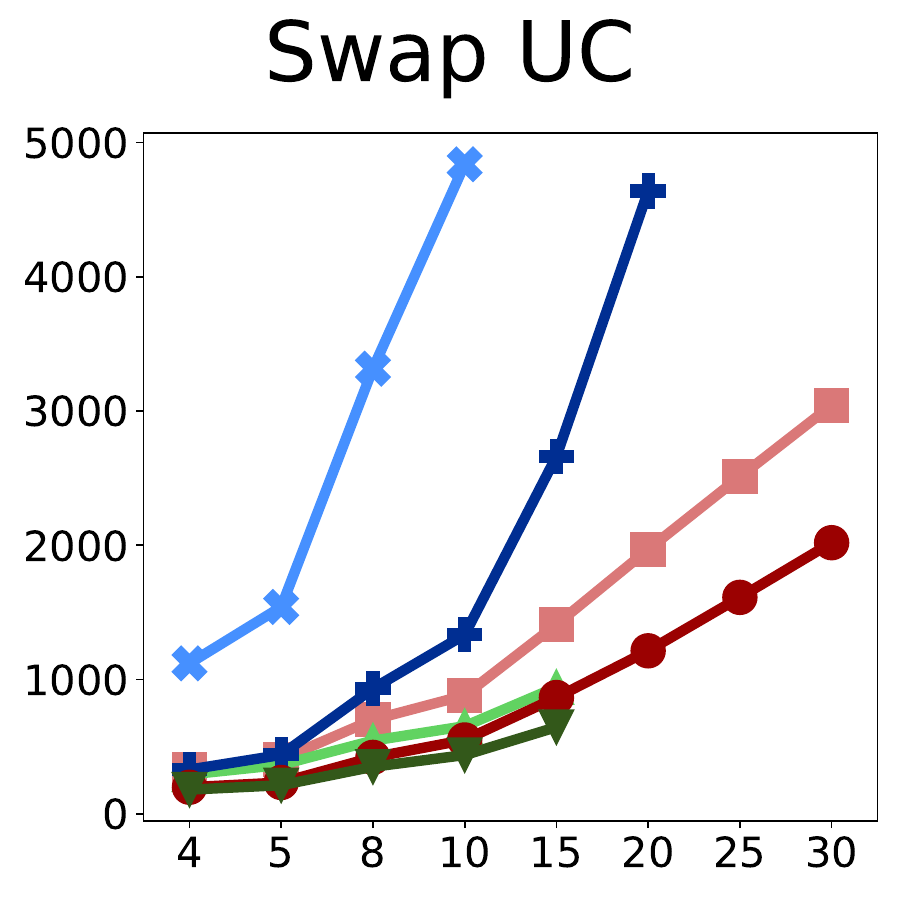}
    \includegraphics[width=0.1125\linewidth]{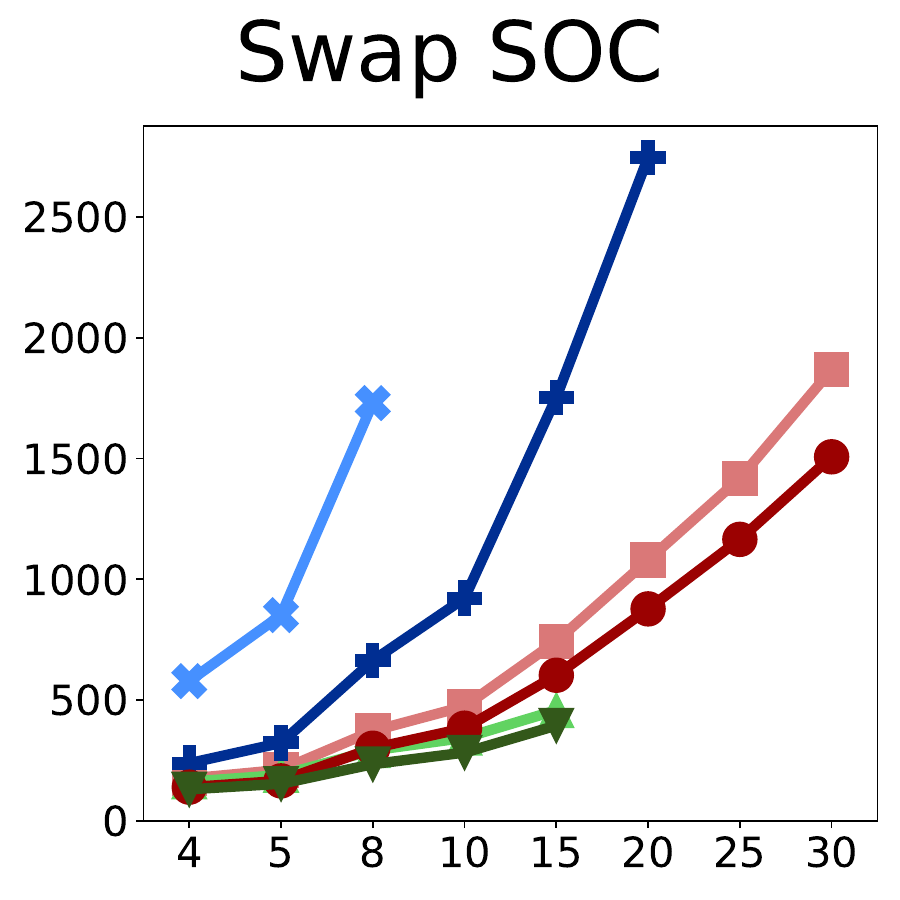}
    \includegraphics[width=0.1125\linewidth]{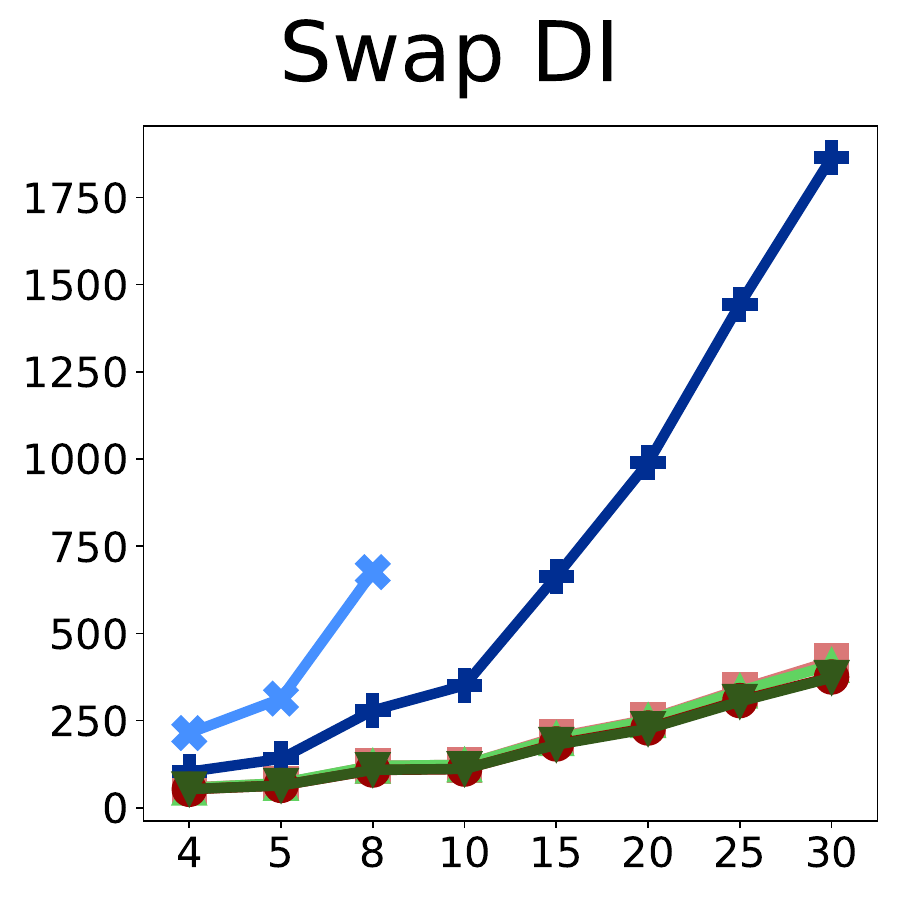}
    \includegraphics[width=0.1125\linewidth]{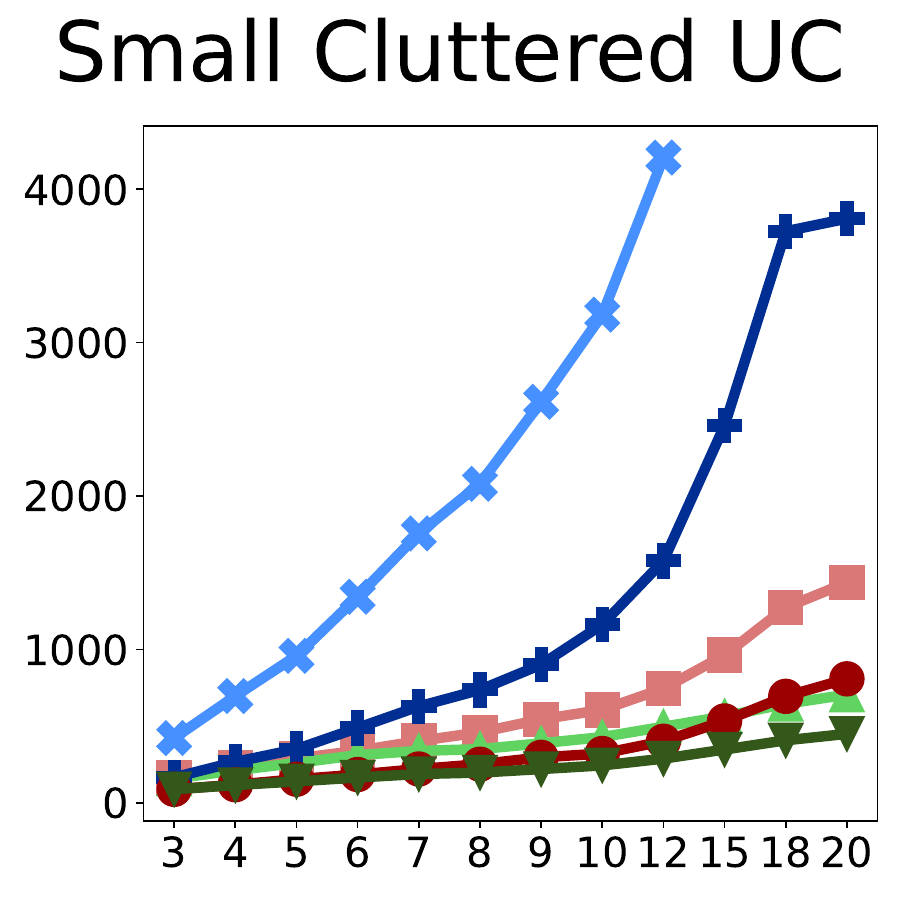}
    \includegraphics[width=0.1125\linewidth]{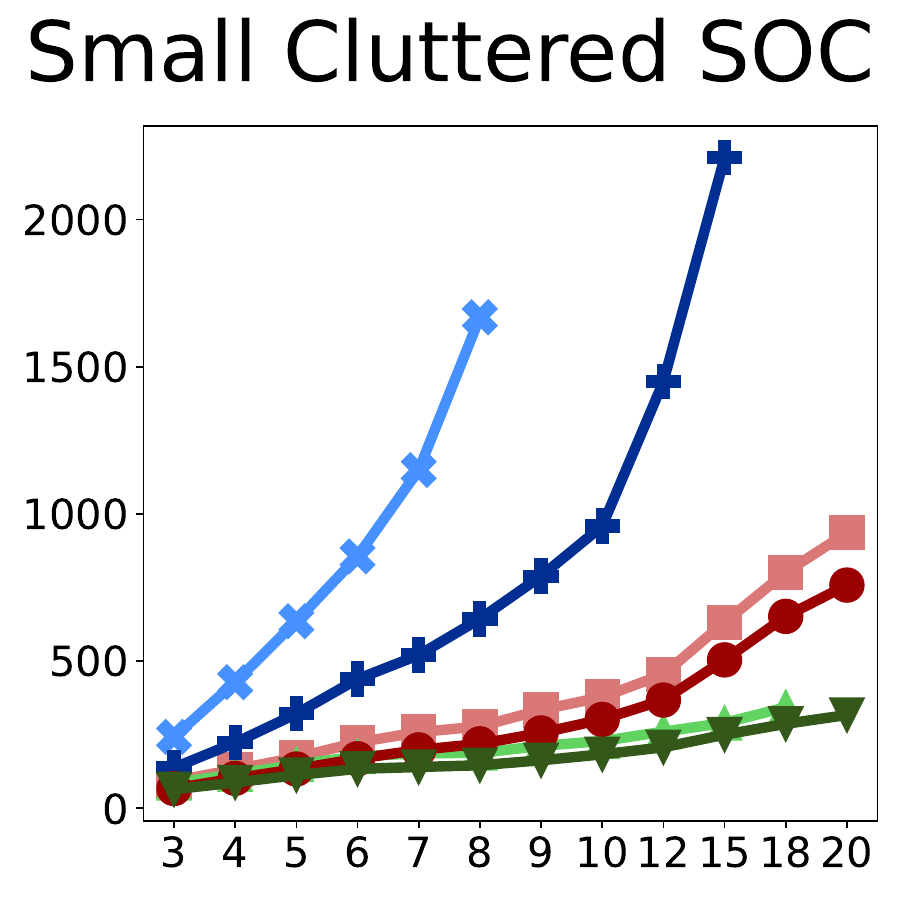}
    \includegraphics[width=0.1125\linewidth]{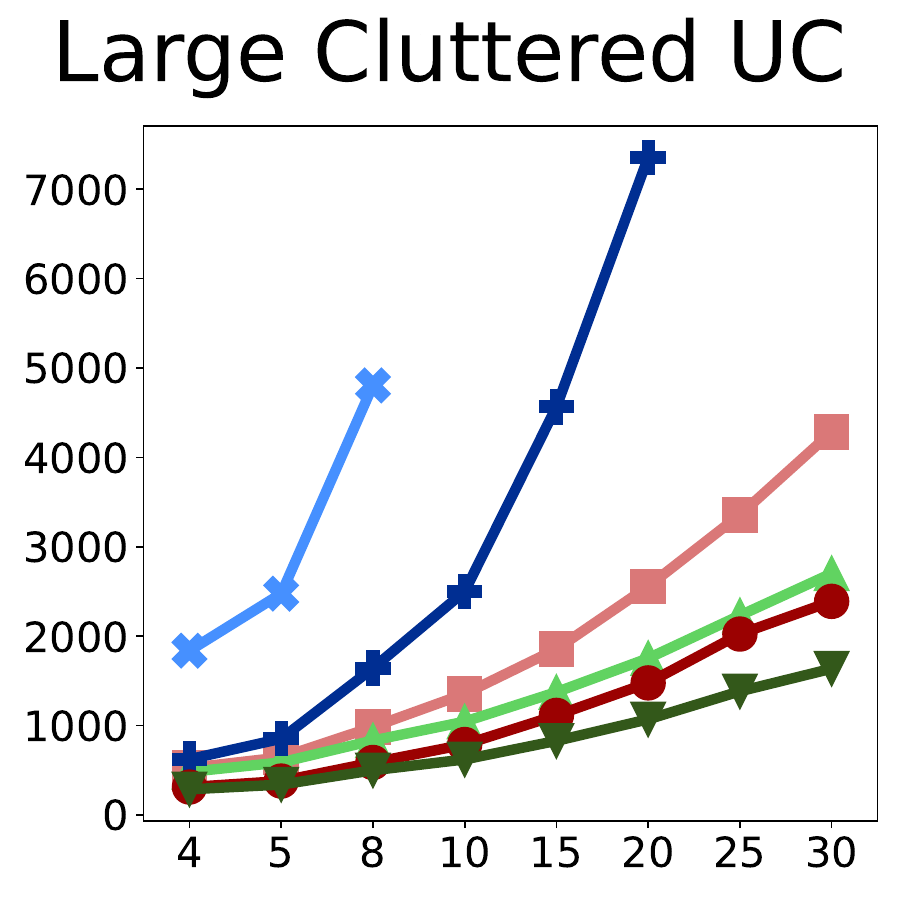}
    \includegraphics[width=0.1125\linewidth]{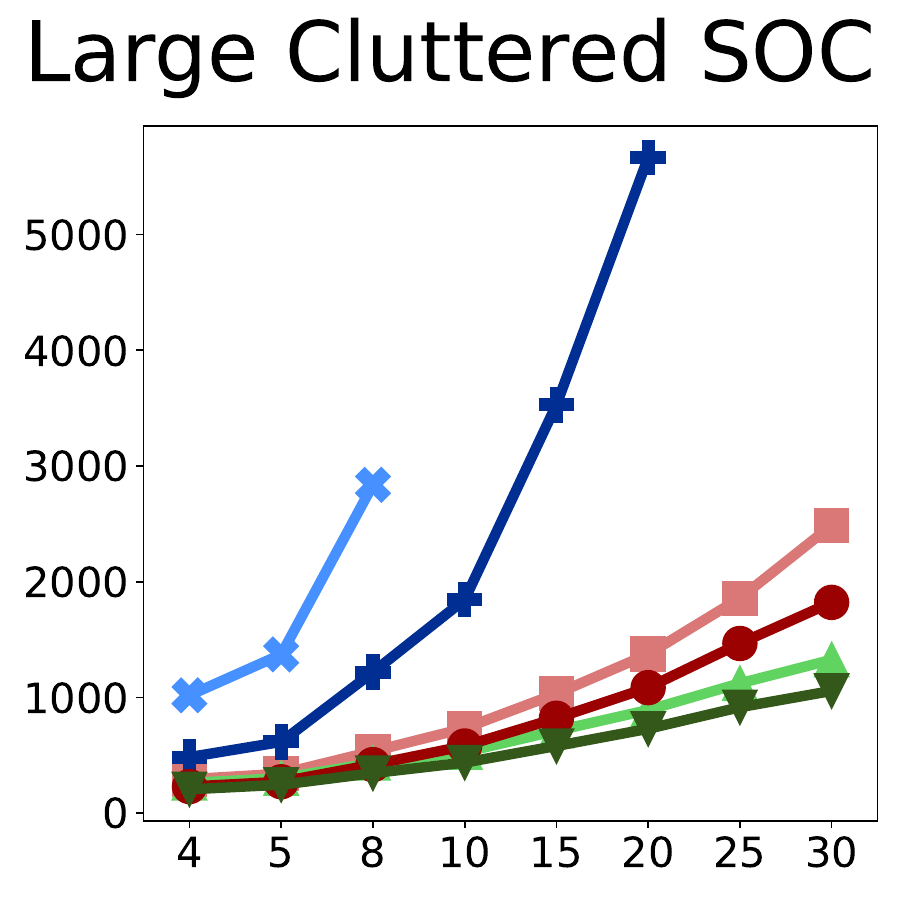}
    \includegraphics[width=0.1125\linewidth]{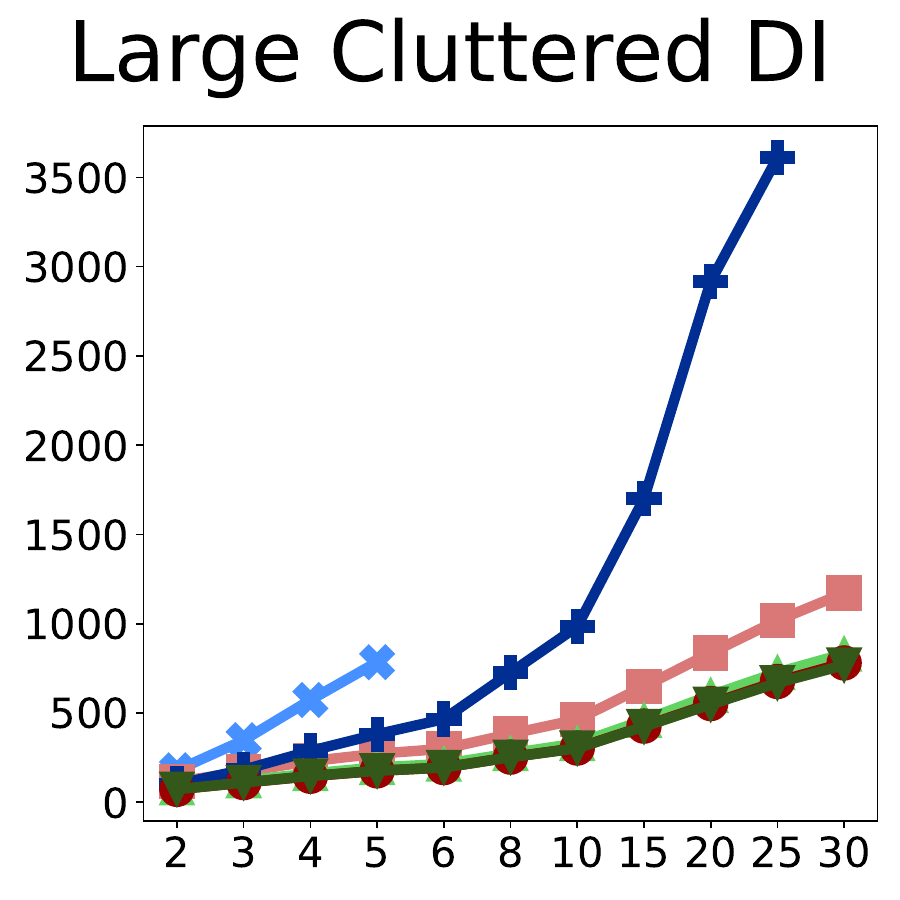}
    \vspace{0.1em}    
    \includegraphics[width=0.5\linewidth]{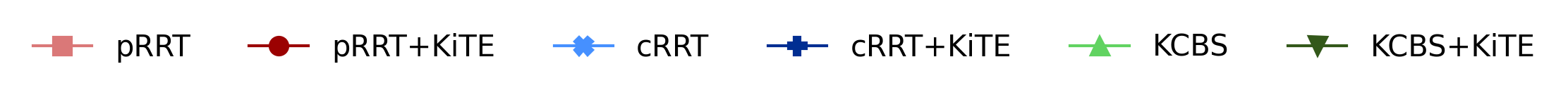}

    \caption{
    \small Average total solution path time (seconds, y-axis) versus number of agents (x-axis) across environments and dynamical models. Darker curves correspond to KiTE-Extend variants, which consistently achieve lower solution times than their baseline counterparts.
    }
    \label{fig:all_path_times_oneline}
\end{figure*}

Table \ref{table_final_shortlisted_results} summarizes representative performance trends across all kinodynamic systems and environments.
Complete results are provided in the Appendix (Tables~\ref{table_unicycle_complete_results}–\ref{table_di_complete_results}) for the Unicycle, Second-Order Car, and Double Integrator systems.
Across all settings, KiTE-Extend improves planning performance, but the nature of the improvement depends on the baseline coordination strategy.
For cRRT, KiTE-Extend primarily improves scalability and feasibility in hard multi-agent regimes; for pRRT and KCBS, KiTE-Extend primarily improves computation time and solution quality.

KiTE-Extend substantially improves the scalability of centralized planning. 
As expected, baseline cRRT degrades rapidly with increasing team size and often fails beyond moderate robot counts. 
In contrast, cRRT+KiTE maintains high success rates in cases where cRRT fails, e.g., swap and small cluttered environments with 15 and 20 robots.
These results indicate that improving the node expansion mechanism significantly extends the practical applicability of joint-state sampling-based planning to larger teams.
In standard cRRT, joint controls are sampled uniformly at random, making coordinated, collision-free progress increasingly unlikely as the number of robots grows.
KiTE-Extend increases the frequency of productive expansions while reducing time spent exploring unpromising branches.
However, as the robot count increases further, cRRT+KiTE's success rate reduces, reflecting the persistent combinatorial complexity of joint-state planning.

KiTE-Extend makes pRRT consistently faster, while largely preserving its strong success rates.
Across the representative settings in Table~\ref{table_final_shortlisted_results}, pRRT+KiTE reduces computation time by a mean of approximately 4.2× while achieving comparable success rates.
For pRRT, success is largely governed by the imposed priority ordering.
KiTE-Extend improves the efficiency of each robot's individual planning call and can recover additional successes in some harder settings, but it does not systematically resolve priority-induced infeasibility.

KiTE-Extend improves KCBS by accelerating low-level replanning while leaving the high-level CBS constraint search unchanged.
Since KCBS repeatedly invokes its low-level planner as constraints are incrementally added, reductions in per-call planning cost accumulate across the high-level search, leading to substantial runtime improvements.
Across the evaluated settings, KCBS+KiTE consistently achieves large computation-time reductions while maintaining, and often improving, success rates as the number of robots increases.
Beyond speed, KiTE-Extend alters the set of motions considered during low-level replanning.
The edge candidates considered by KiTE-Extend frequently include low velocity or pause-like motions that remain feasible when agents must yield to other agent's trajectories, naturally supporting stop–wait–move behaviors that are effective in tightly coupled multi-robot settings (see Fig~\ref{fig:edge_bundle_selection}).
In contrast, KCBS without KiTE-Extend relies on random control sampling, making such coordinated waiting behavior unlikely to arise consistently. 
This effect becomes especially pronounced when low-level search trees are pruned and reused across replans, where KCBS+KiTE readily exploits feasible waiting or recovery motions following constraint updates. Results for experiments with pruning can be found in Table \ref{table_kcbs_pruning_results_soc} in the Appendix.

Across all coordination paradigms, KiTE consistently improves solution quality as measured by total path duration. 
When comparing trials in which both the baseline planner and KiTE-augmented planners succeed, KiTE-Extend produces shorter duration trajectories on average, with reductions of approximately 40–60\% for cRRT and 20–30\% for pRRT and KCBS in the evaluated settings (see Fig~\ref{fig:all_path_times_oneline}).
Although edge bundles are generated through random offline sampling, performance gains arise from retrieving locally compatible trajectory segments and prioritizing those that make the most progress toward the sampled target state during online expansion.
We observed that retrieving locally compatible edge candidates alone already improves performance over naive random sampling, but the combination of retrieval and ranking consistently yields the strongest performance across all MRMP paradigms, highlighting the importance of the ranking mechanism.
This improvement reduces detours and oscillatory motion while still allowing low-velocity or wait-like behavior when required for feasibility.

dbCBS achieves very low path durations in small and moderately sized Unicycle environments when it succeeds, often producing shorter trajectories than sampling-based planners. 
However, this improvement comes at the cost of substantially higher computation time and poor scalability as the number of agents increases or the environment size grows. 
In larger or more cluttered settings, dbCBS frequently requires orders of magnitude more time to find a feasible solution, and in many cases fails to do so within the allotted time budget. 
In contrast, KiTE-guided planners trade some solution optimality for significantly improved scalability, producing feasible solutions much more quickly while maintaining competitive, though not always minimal path durations.

\section{Conclusion} \label{section_conclusion}

This paper presented KiTE-Extend, a planner-agnostic action selection scheme that leverages translation-invariant offline computation to accelerate node expansion in sampling-based kinodynamic motion planning.
By reframing node expansion as a retrieval and ranking problem over precomputed trajectory segments, KiTE-Extend enables efficient reuse of offline computation while preserving feasibility through forward simulation.
This makes KiTE particularly effective for multi-robot motion planning, where planning under spatiotemporal constraints amplifies the cost of inefficient expansion.
Across cluttered environments and multiple kinodynamic systems, augmenting sampling-based planners with KiTE-Extend improves success rates, reduces computation time, and yields higher-quality solutions when used within centralized, prioritized, and CBS-style MRMP frameworks, without sacrificing theoretical guarantees.
More broadly, this work highlights low-level expansion efficiency as a central bottleneck in MRMP and suggests a principled path toward scalability through improved reuse of feasible expansions rather than increased coordination complexity.

\emph{Limitations:} 
As the dimensionality of the translation-invariant state components increases, larger edge bundles may be required to achieve adequate coverage, increasing offline computation and retrieval cost.
This challenge becomes significant for high-dimensional systems with limited symmetry.
Future work will investigate structured bundle generation, additional symmetry exploitation, and problem-specific retrieval strategies to improve scalability in such settings.

\FloatBarrier
\bibliographystyle{plainnat}
\bibliography{himanshu_references,paul_references,mrmp_references}


\clearpage
\appendix

\addcontentsline{toc}{section}{Appendix}


\etocsetnexttocdepth{subsection} 
\localtableofcontents

\FloatBarrier

\section*{Overview}

This appendix provides complete experimental results, ablation studies, and additional analyses that complement the main paper.
Specifically, it includes:
(i) full planner performance tables for all evaluated motion models (Unicycle, Second Order Car, and Double Integrator) across environments and robot counts,
(ii) distributional visualizations illustrating variability and tail behavior underlying aggregate statistics,
(iii) an analysis of pruning in the KCBS low-level planner and its interaction with KiTE-Extend, and
(iv) ablation studies examining the effect of edge-bundle size and skip factor on KiTE-Extend performance.

Unless otherwise stated, trends discussed here are consistent with those reported in the main paper and are intended to provide additional insight rather than introduce new claims.

The supplementary material additionally includes videos demonstrating side-by-side executions of baseline planners and their KiTE-Extend variants for selected environments and robot counts.

\clearpage

\subsection{Results for the Unicycle model}

\begin{figure*}[t!]
\centering

\imgrow{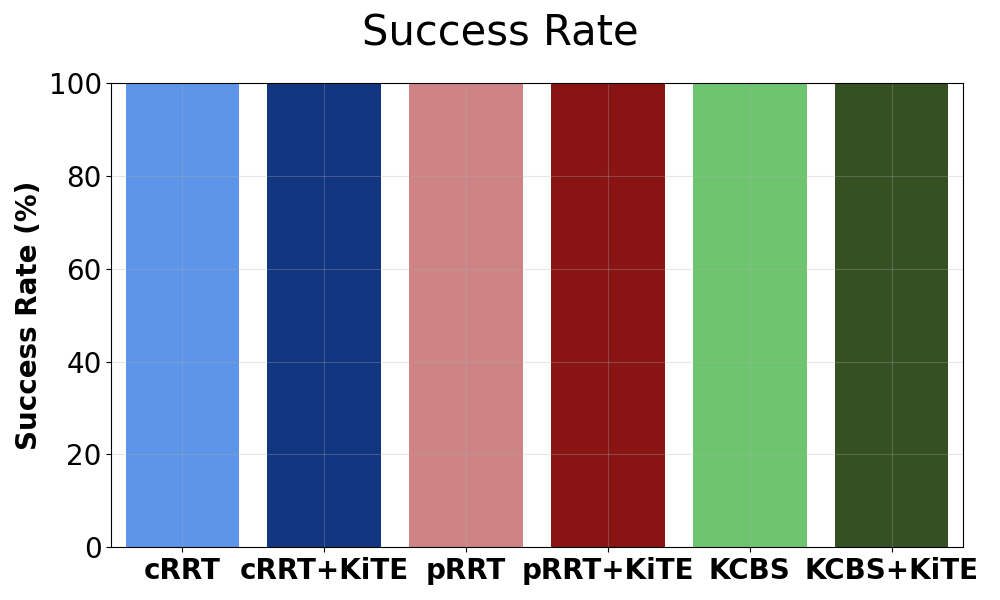}{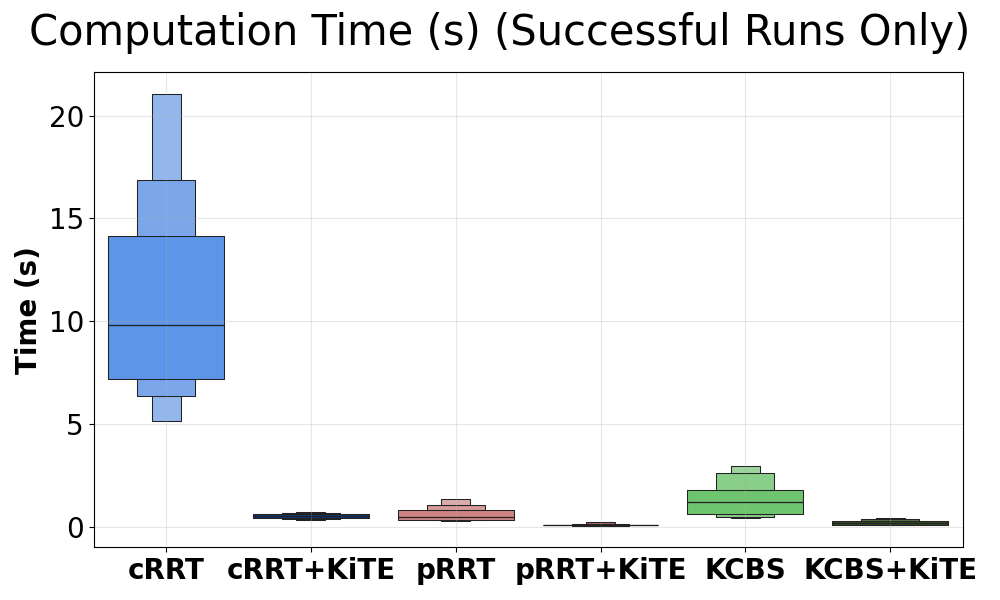}{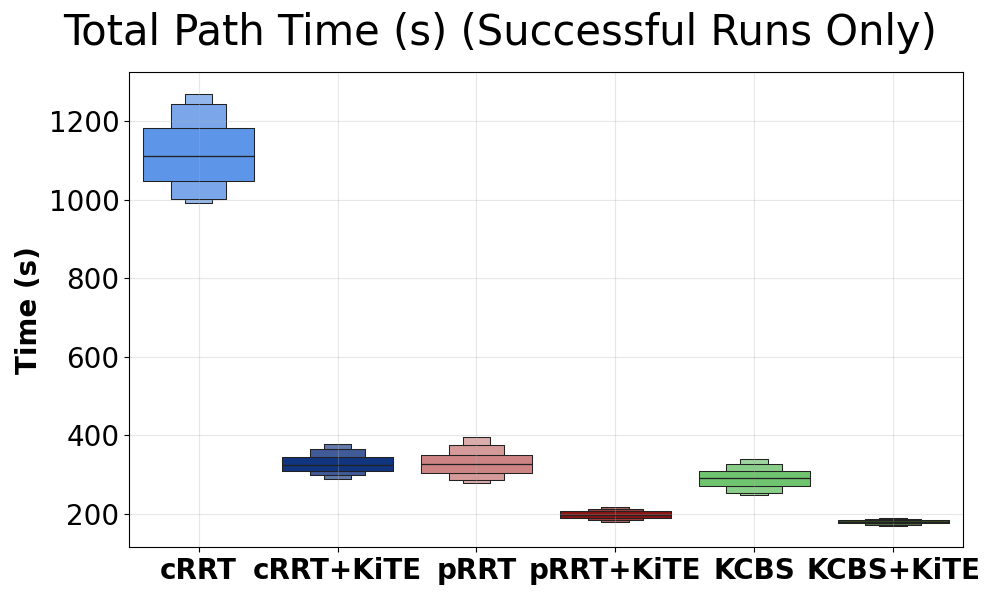}
\rowsubcap{SWAP}{\(N=4\) robots}

\imgrow{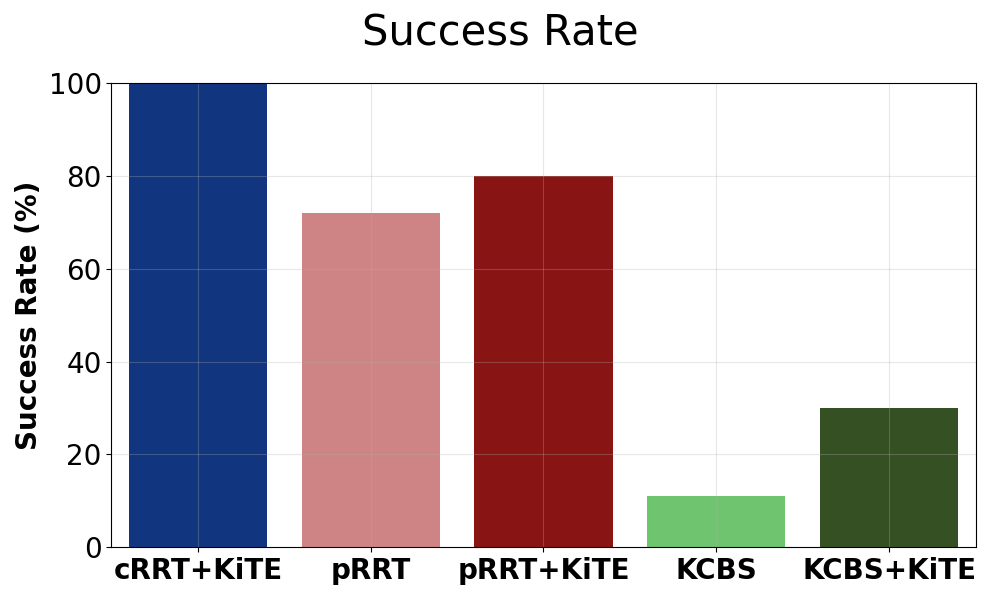}{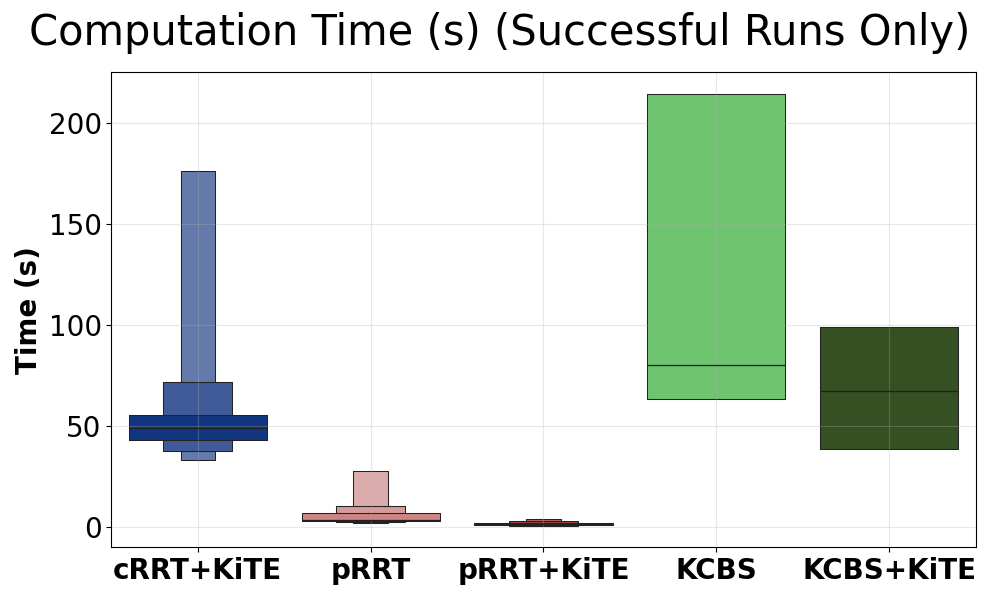}{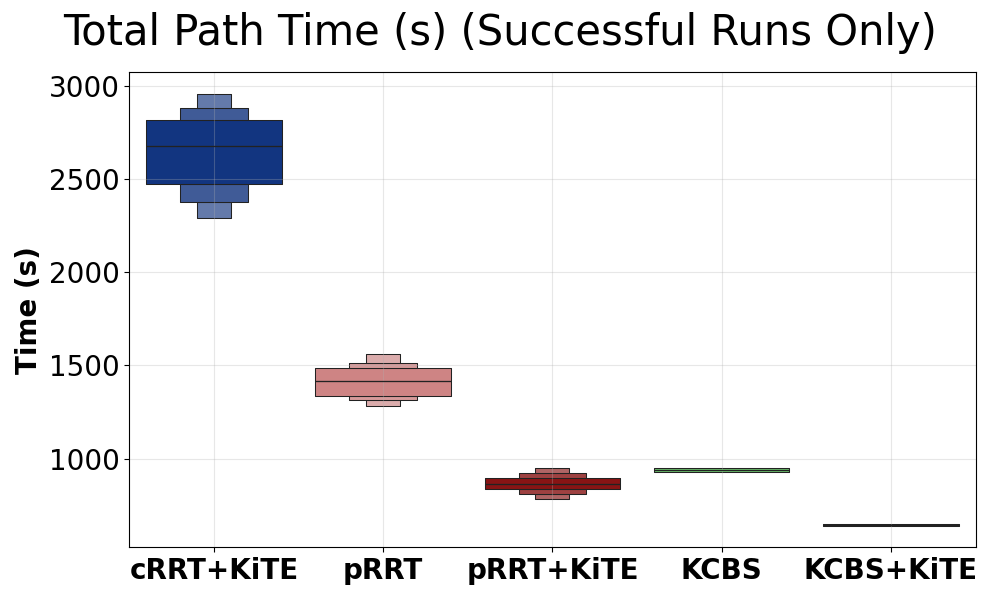}
\rowsubcap{SWAP}{\(N=15\) robots}

\imgrow{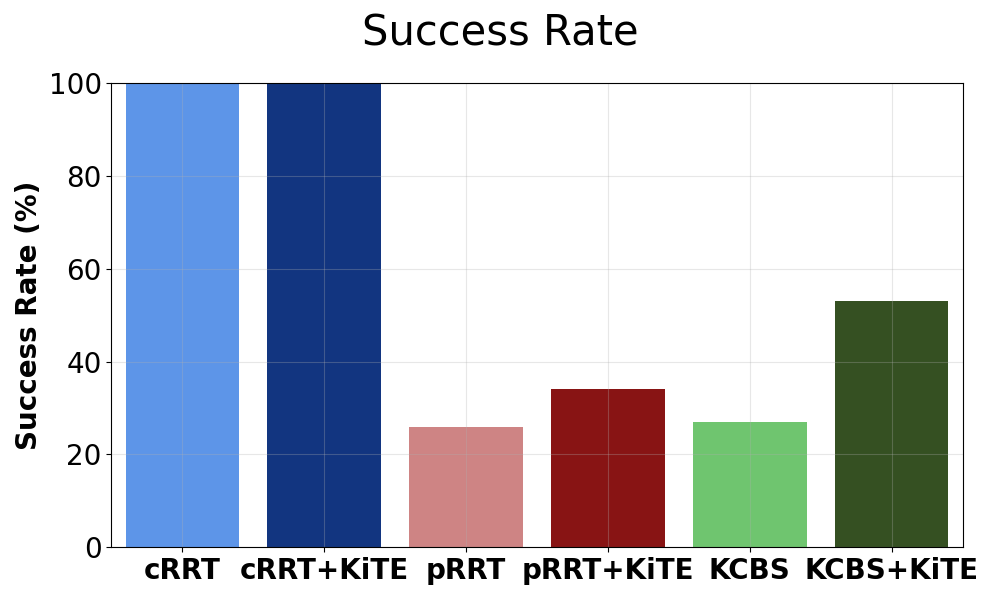}{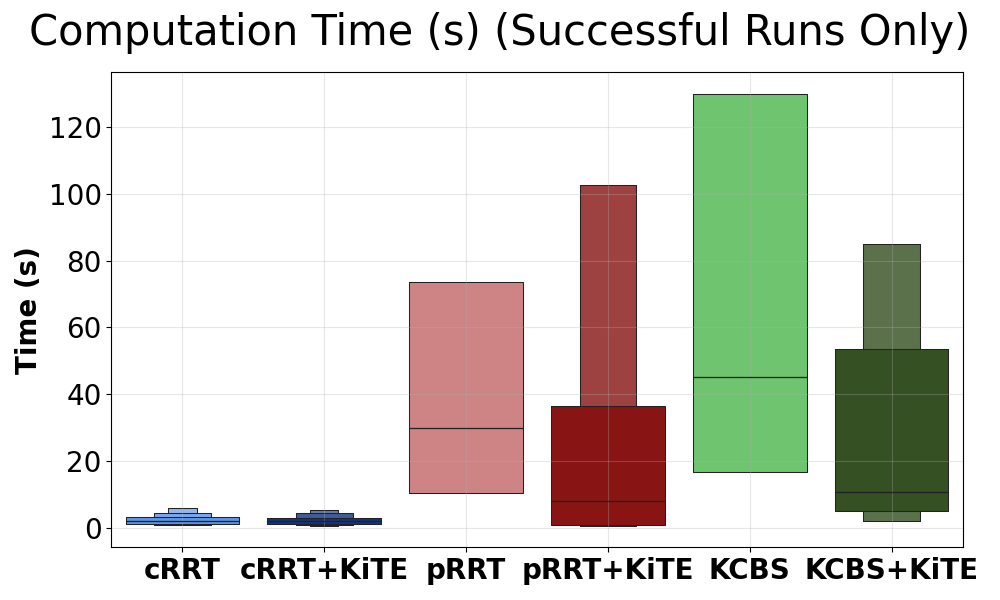}{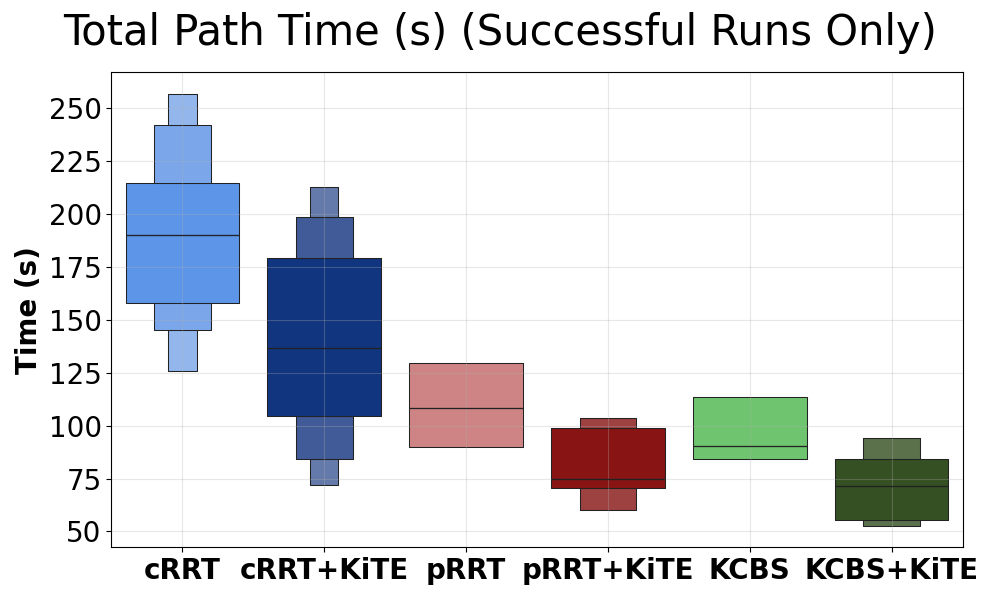}
\rowsubcap{Narrow Corridor}{\(N=3\) robots}

\imgrow{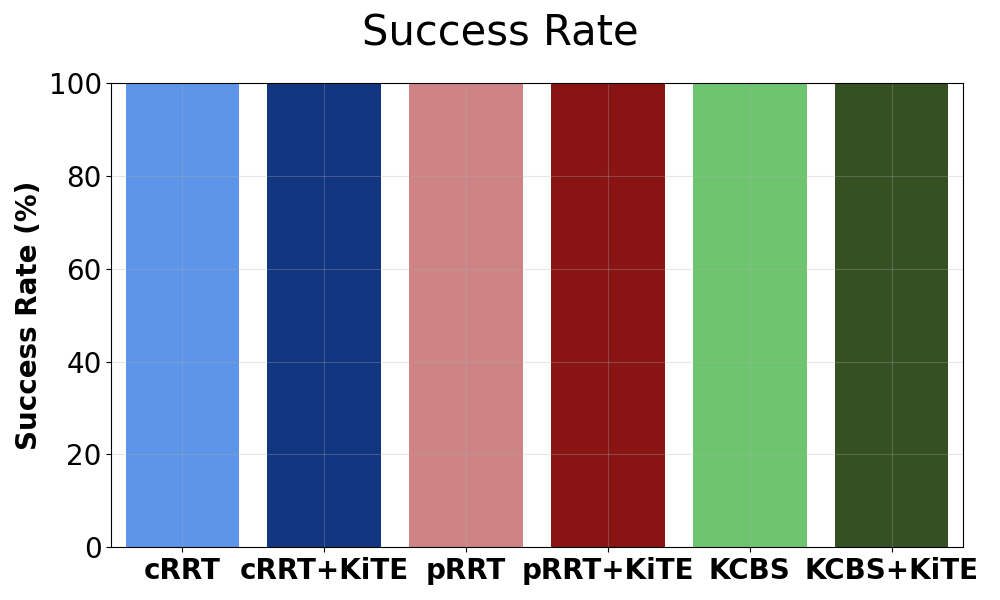}{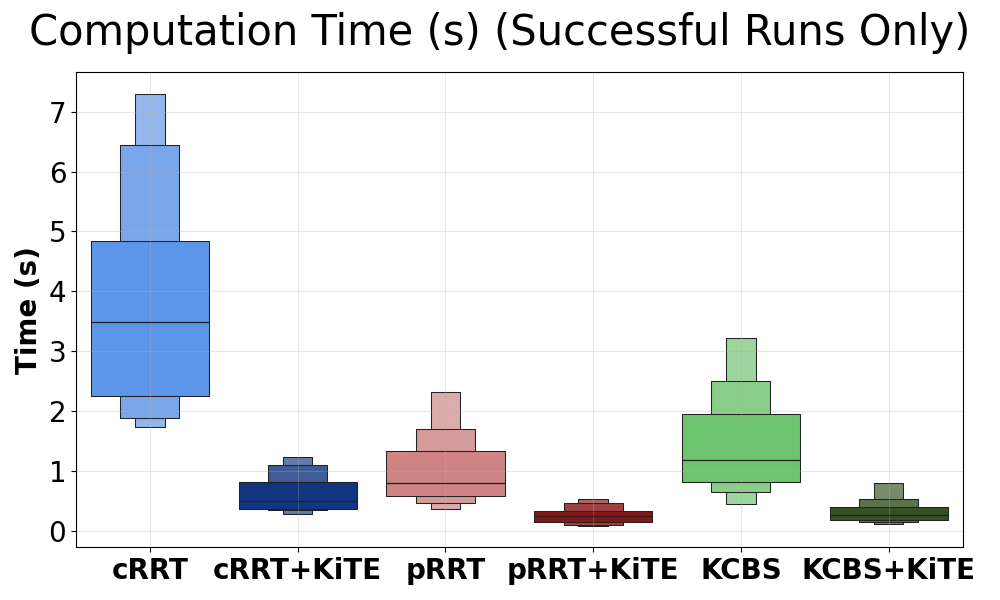}{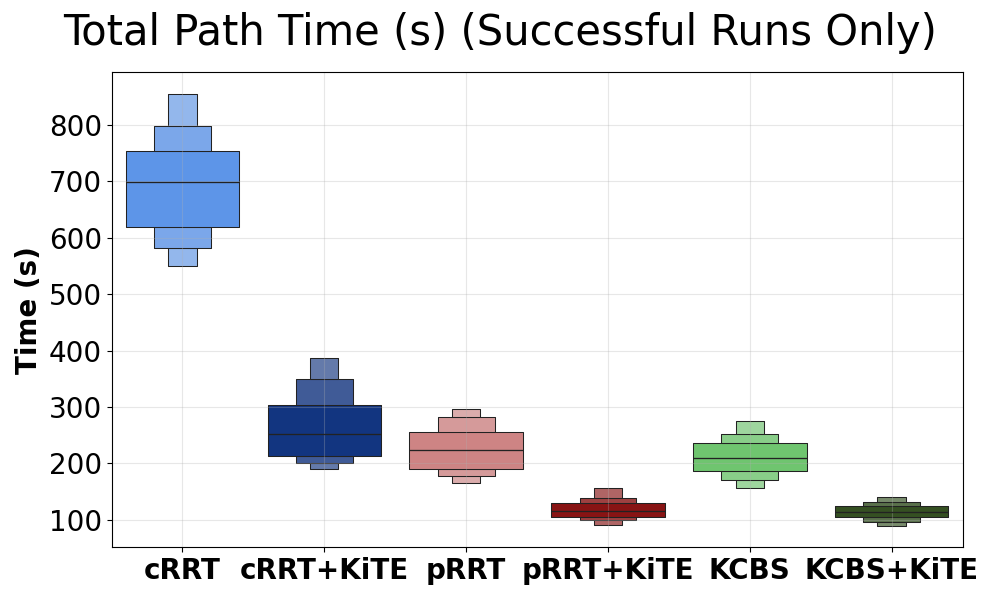}
\rowsubcap{Small Cluttered}{\(N=4\) robots}

\imgrow{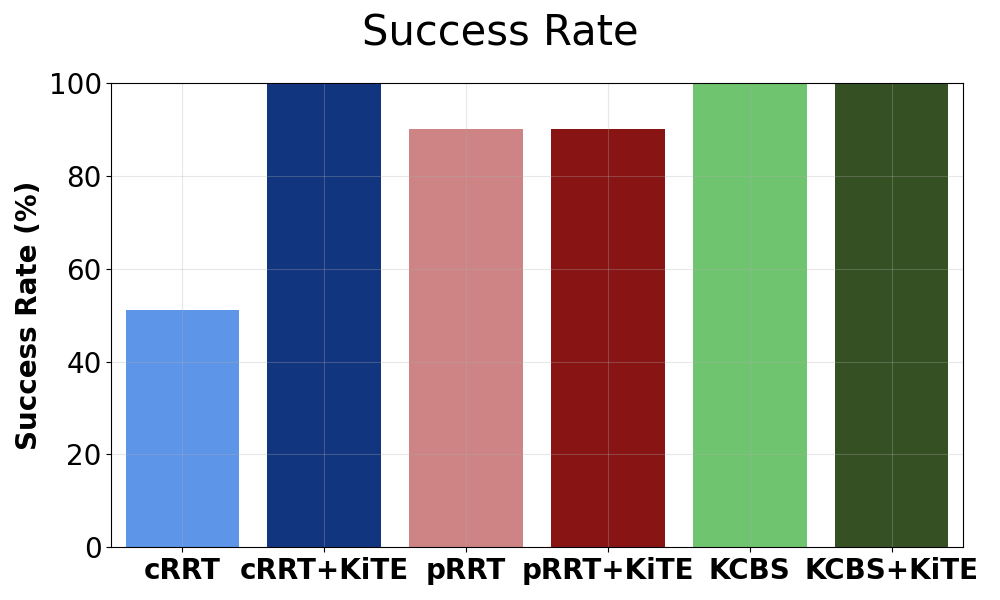}{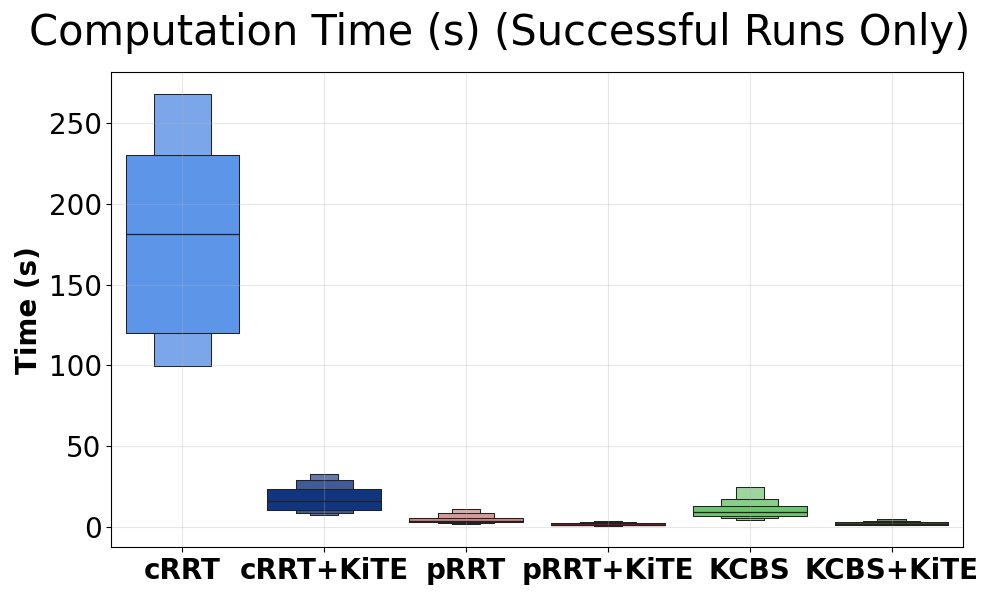}{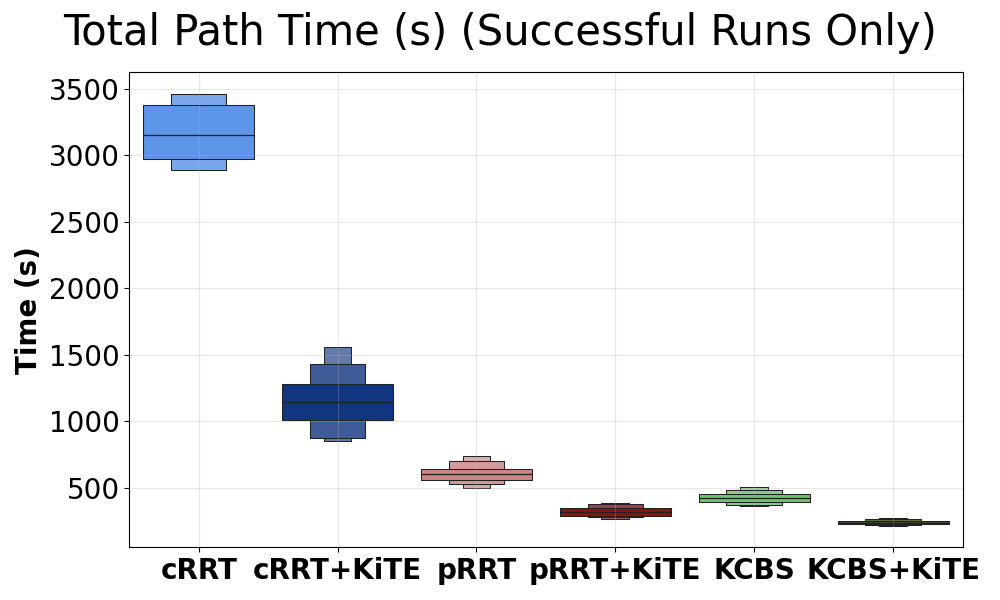}
\rowsubcap{Small Cluttered}{\(N=10\) robots}

\caption{
Detailed results for selected environments in experiments with the Unicycle model.
Each row corresponds to one experimental setting (environment and agent count \(N\)) and contains, from left to right:
(i) success rate (SR; bar chart), (ii) computation time (CT; boxen plot), and (iii) total path time (PT; boxen plot),
comparing each baseline planner to its KiTE-Extend variant.
These plots visualize variability and tail behavior underlying the aggregate results in Table~\ref{table_unicycle_complete_results}.
}
\label{fig:unicycle_rows_1_5}
\end{figure*}

\begin{figure*}[t!]
\centering

\imgrow{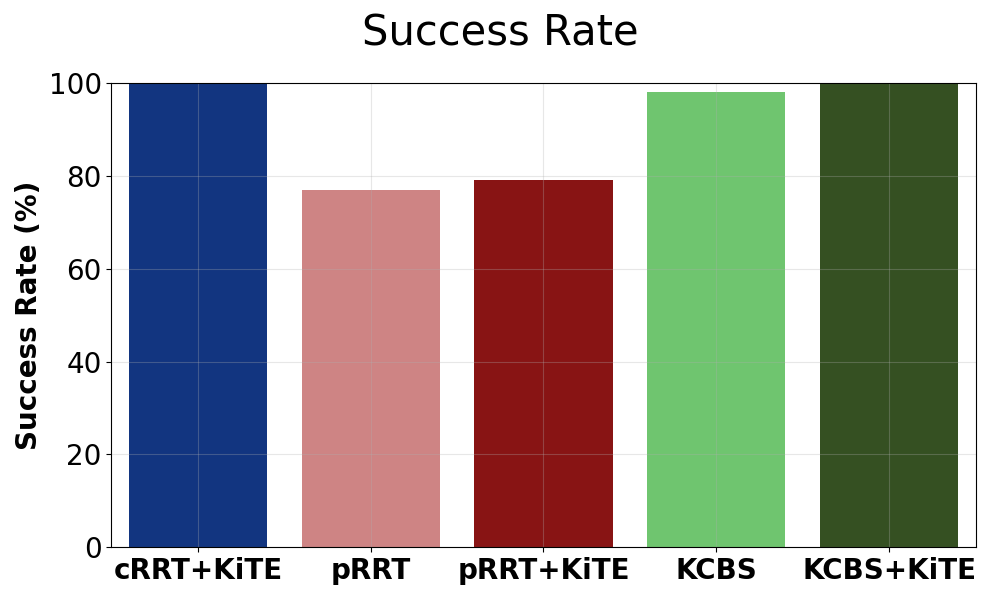}{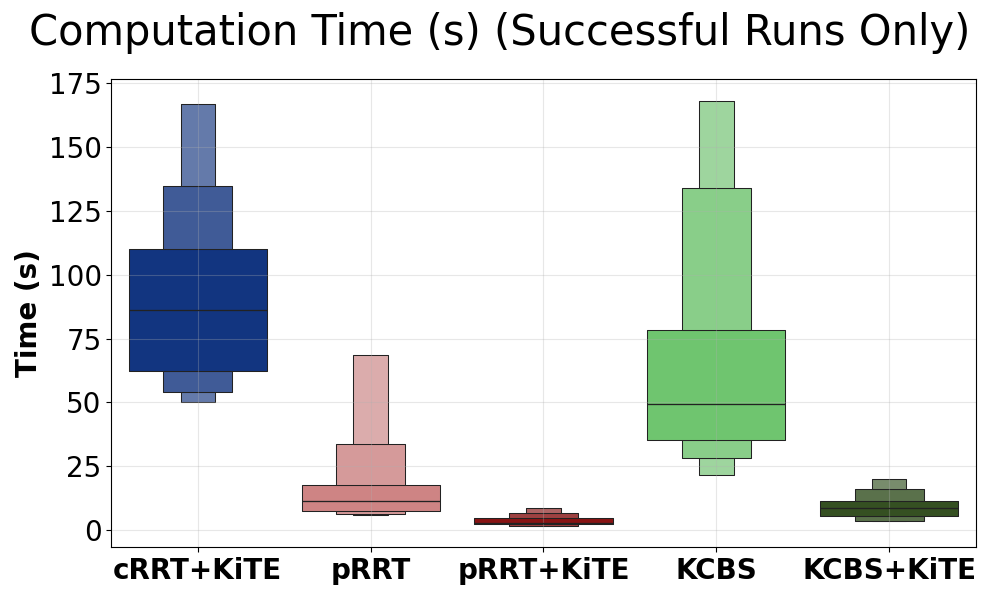}{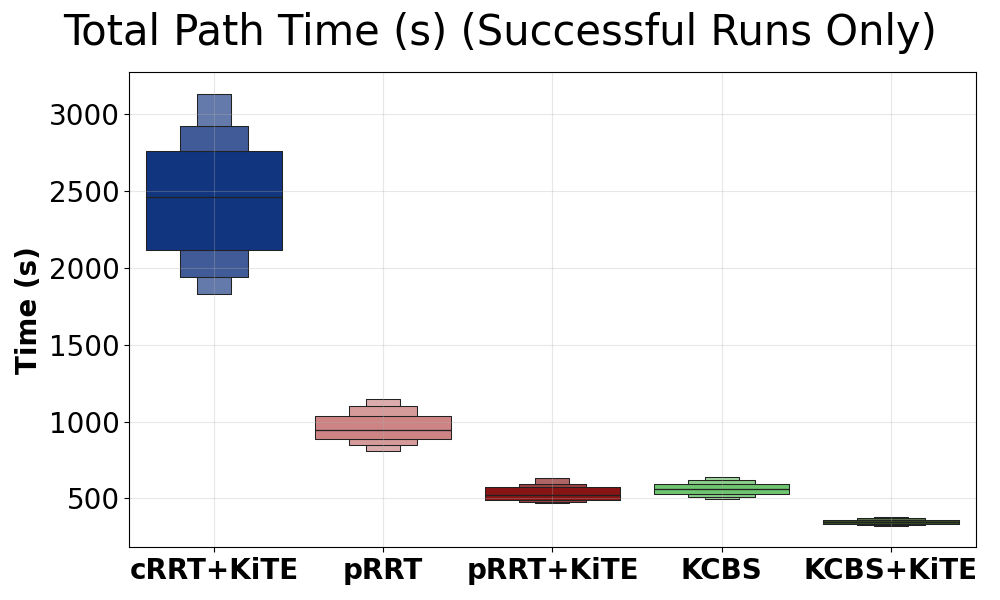}
\rowsubcap{Small Cluttered}{\(N=15\) robots}

\imgrow{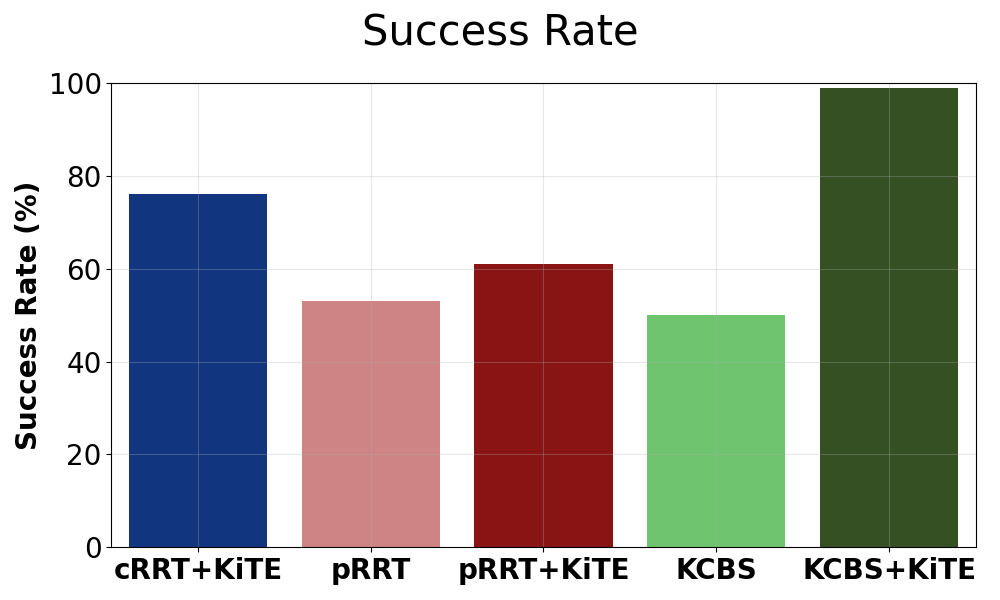}{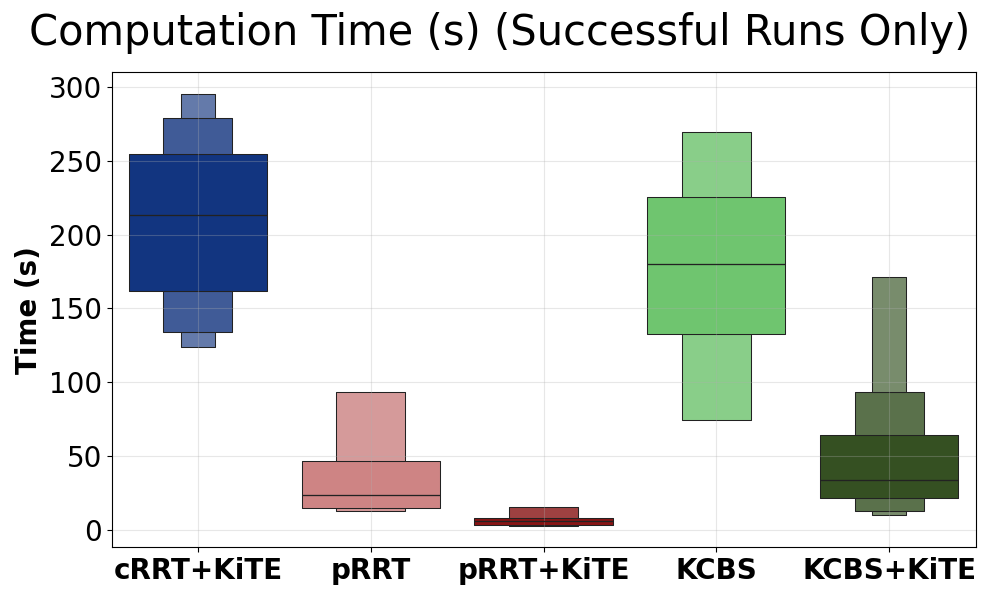}{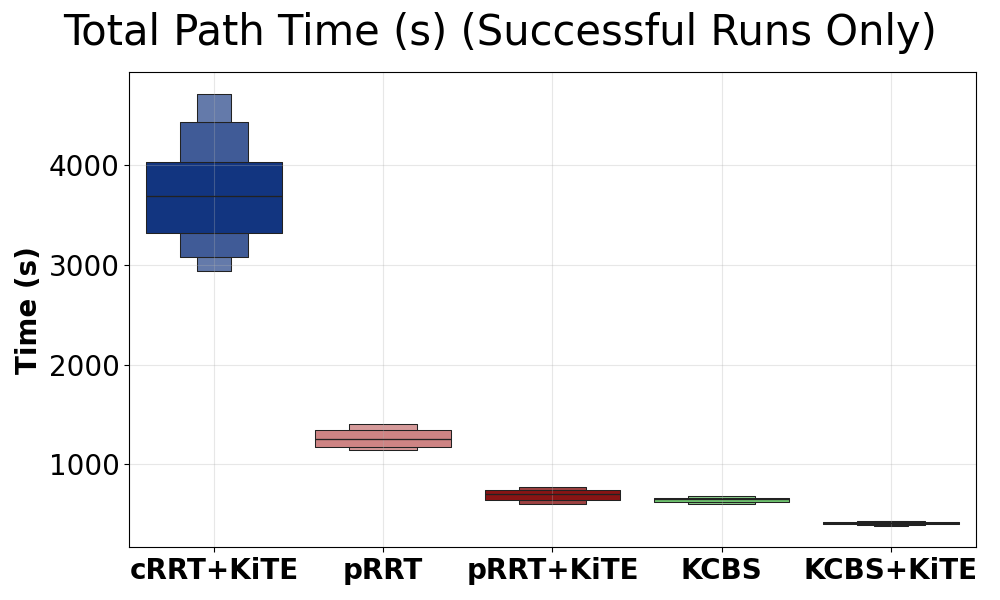}
\rowsubcap{Small Cluttered}{\(N=18\) robots}

\imgrow{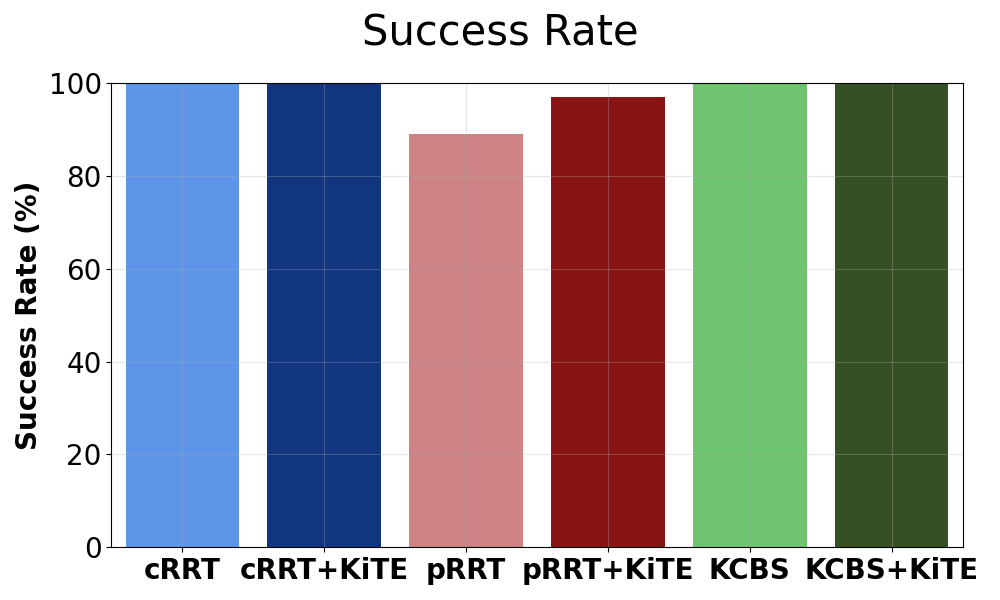}{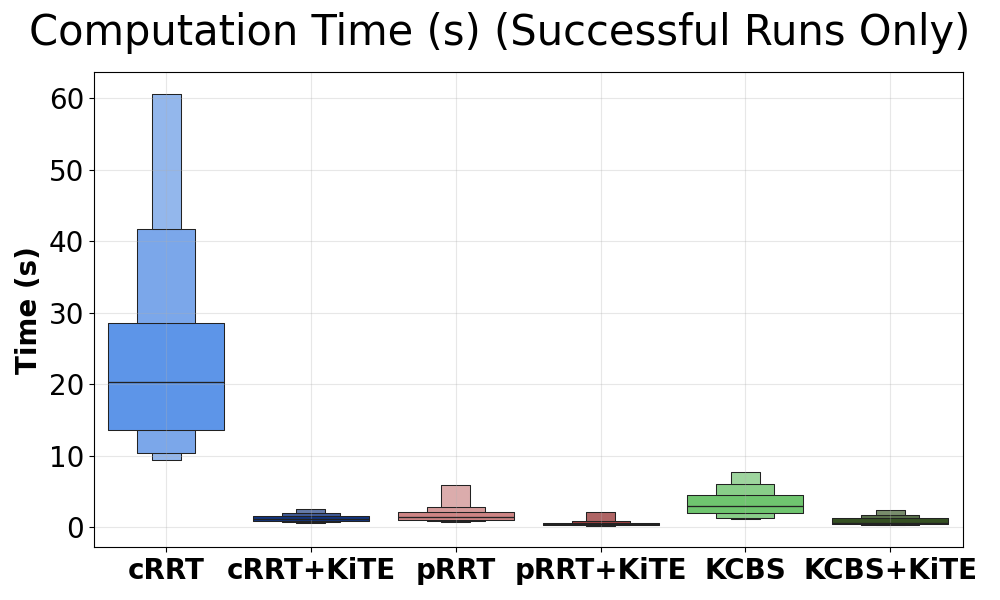}{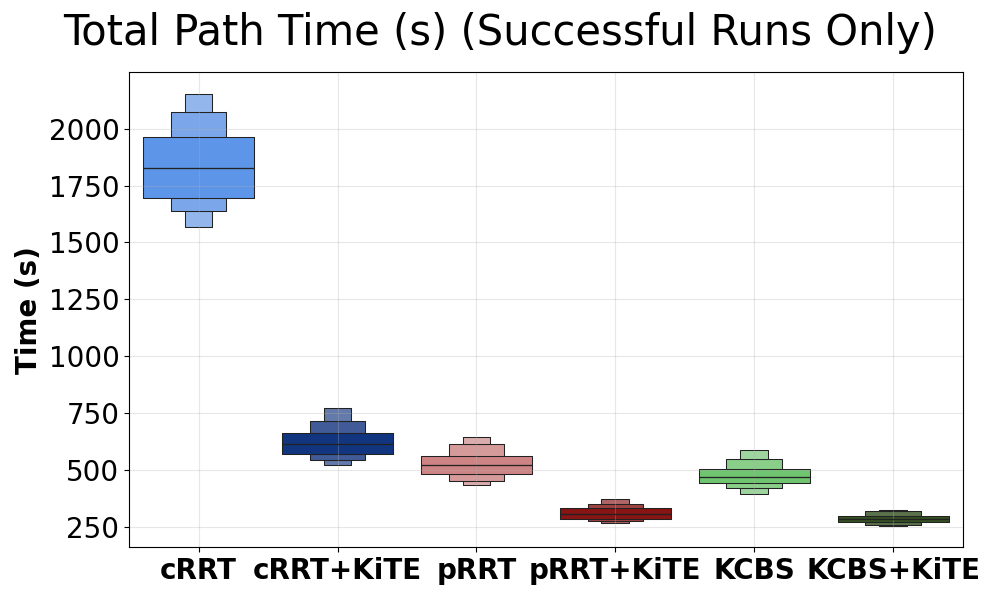}
\rowsubcap{Large Cluttered}{\(N=4\) robots}

\imgrow{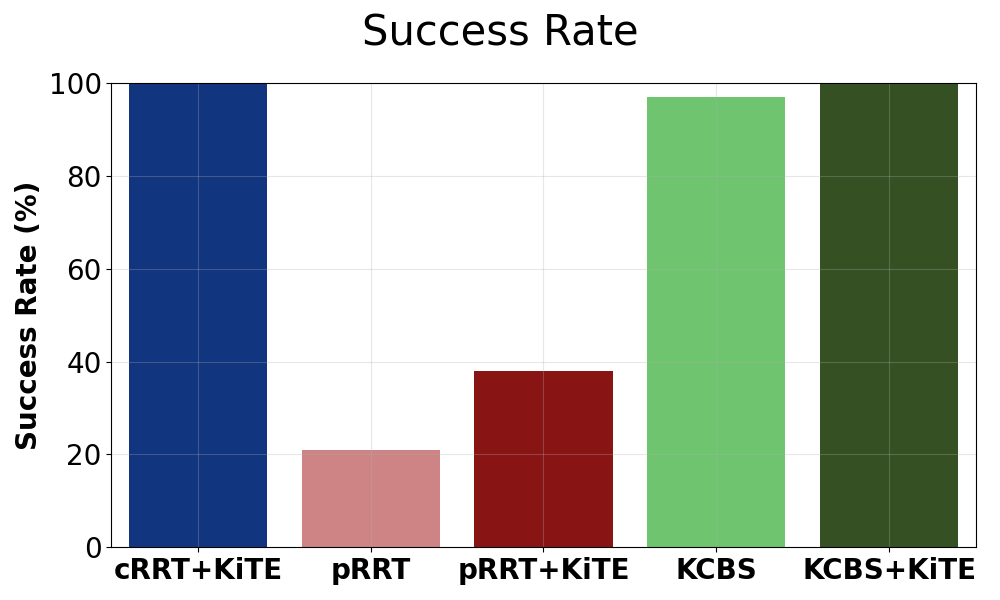}{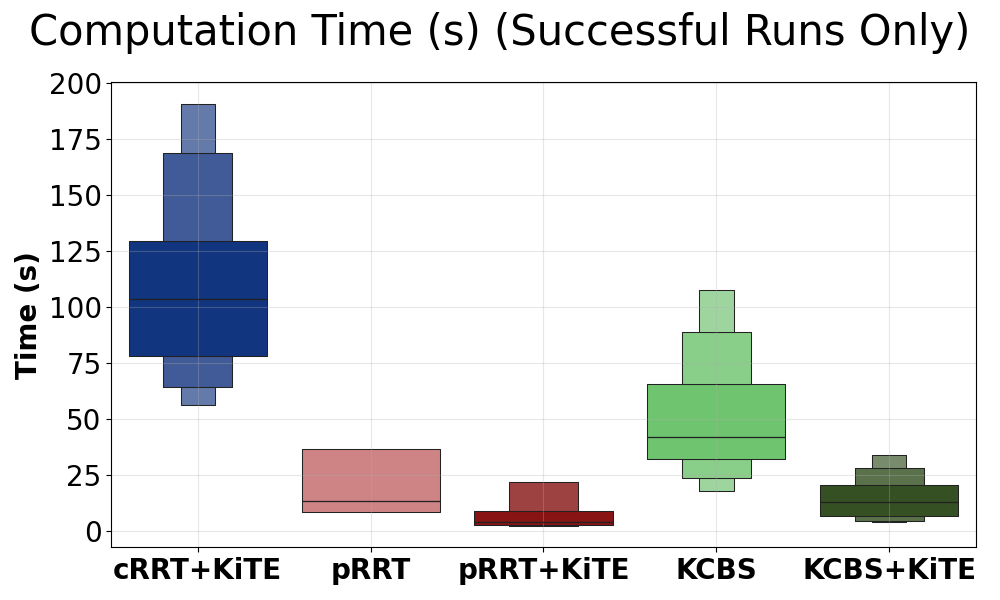}{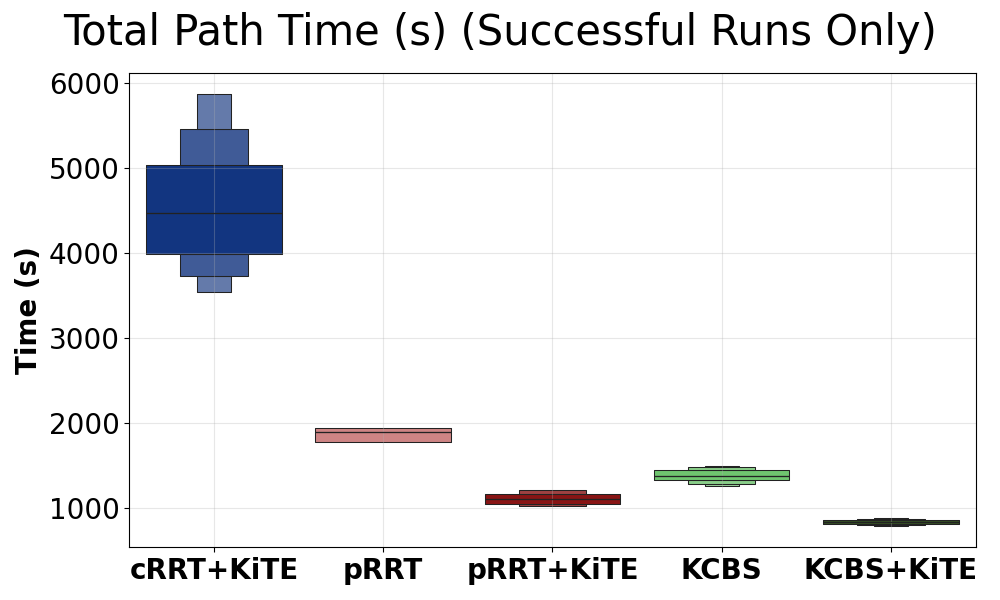}
\rowsubcap{Large Cluttered}{\(N=15\) robots}

\imgrow{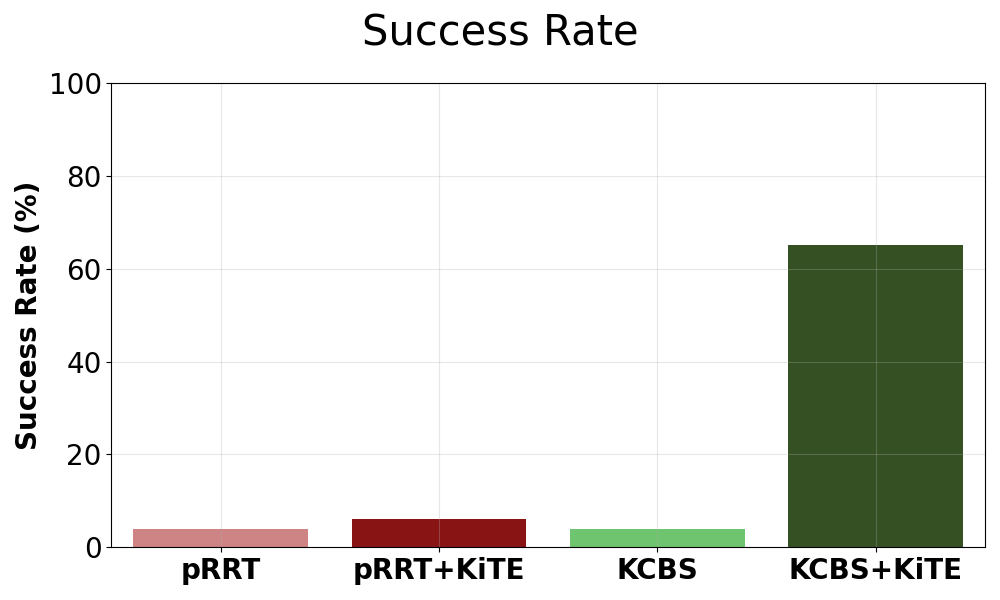}{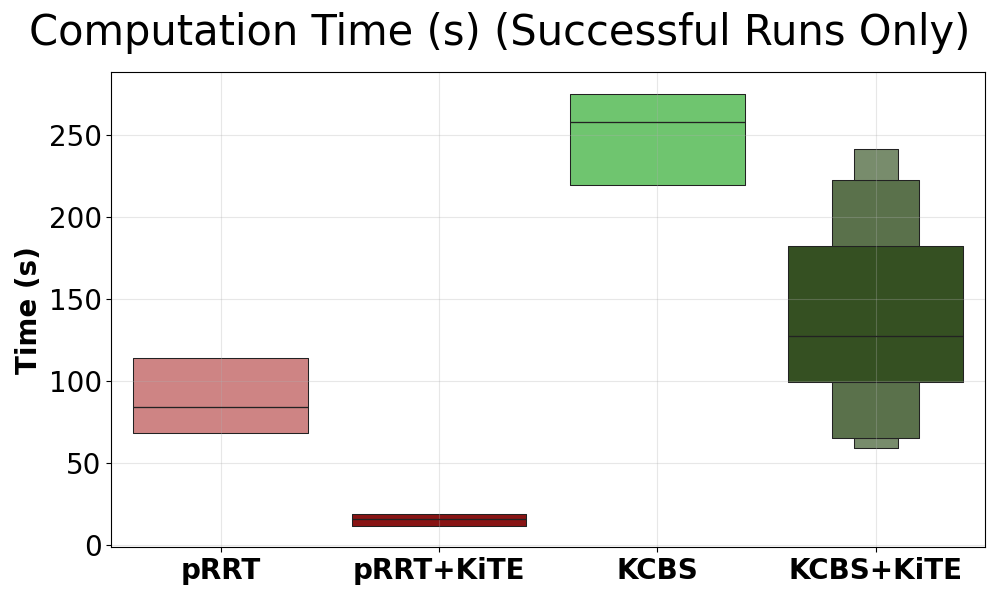}{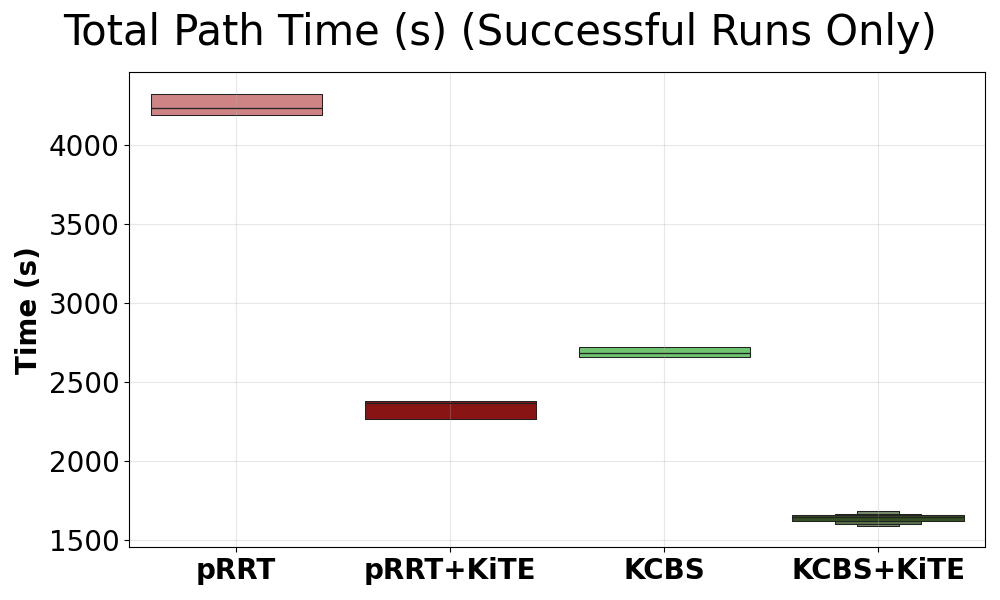}
\rowsubcap{Large Cluttered}{\(N=30\) robots}

\caption{
Detailed results for selected environments in experiments with the Unicycle model (continued).
Each row corresponds to one experimental setting (environment and agent count \(N\)) and contains, from left to right:
(i) success rate (SR; bar chart), (ii) computation time (CT; boxen plot), and (iii) total path time (PT; boxen plot),
comparing each baseline planner to its KiTE-Extend variant.
These plots visualize variability and tail behavior underlying the aggregate results in Table~\ref{table_unicycle_complete_results}.
}
\label{fig:unicycle_rows_6_10}
\end{figure*}

Table \ref{table_unicycle_complete_results} reports complete results for the Unicycle model across all environments and planners.
KiTE-Extend consistently improves feasibility and scalability for sampling-based planners, most notably for cRRT, where it converts widespread failure at higher robot counts into consistent success across swap, and cluttered settings.
For pRRT and KCBS, KiTE-Extend yields substantial reductions in computation time and path duration while largely preserving success rates, with the largest gains occurring in cluttered and high-density settings.
Across all environments, these trends mirror those reported in the main paper and indicate that KiTE’s benefits are robust to both environment structure and agent count.
To complement the aggregate statistics reported in Table~\ref{table_unicycle_complete_results}, 
Fig.~\ref{fig:unicycle_rows_1_5} and Fig.~\ref{fig:unicycle_rows_6_10} visualize performance distributions for representative environments in experiments with the Unicycle model, including success rate (bar charts) as well as computation time and total path duration (boxen plots).

\begin{table*}[!ht]
\caption{
\small
\textbf{Planner performance for the Unicycle (UC) model across environments and robot counts.}
Results compare baseline planners and their KiTE-Extend variants for cRRT, pRRT, KCBS, and dbCBS, reported in terms of success rate (SR), computation time (CT), and total path time (PT).
Bold entries indicate the best value for each metric within a given planner and environment; for success rate, all entries achieving the maximum value are bolded.}
\begin{center}

\resizebox{1\linewidth}{!}{%
\begin{tabular}{l|c|ccc|ccc|ccc|ccc}
\toprule
\textbf{Domain} & \textbf{Variant}
& \multicolumn{3}{c|}{\textbf{cRRT}}
& \multicolumn{3}{c|}{\textbf{pRRT}}
& \multicolumn{3}{c|}{\textbf{KCBS}}
& \multicolumn{3}{c}{\textbf{dbCBS}} \\

& 
& SR (\%) & CT (s) & PT (s)
& SR (\%) & CT (s) & PT (s)
& SR (\%) & CT (s) & PT (s)
& SR (\%) & CT (s) & PT (s) \\[1mm]
\hline
\hline

{Narrow\;Corridor} & Base & \textbf{100} & 2.40 $\pm$ 0.17 & 192 $\pm$ 4.4 & 26 & 51.9 $\pm$ 12 & 110 $\pm$ 5.6 & 27 & 81.4 $\pm$ 15 & 96.9 $\pm$ 5.1 & 100 & 5.08 $\pm$ 0.033 & 28.6 $\pm$ 0.048 \\
{(N=3)} & KiTE & \textbf{100} & \textbf{2.33 $\pm$ 0.17} & \textbf{140 $\pm$ 4.8} & \textbf{34} & \textbf{34.0 $\pm$ 8.6} & \textbf{81.2 $\pm$ 3.3} & \textbf{53} & \textbf{36.1 $\pm$ 7.1} & \textbf{74.6 $\pm$ 3.2} & - & - & - \\
\hline
{Swap} & Base & \textbf{100} & 11.7 $\pm$ 0.67 & 1120 $\pm$ 10.0 & \textbf{100} & 0.645 $\pm$ 0.038 & 330 $\pm$ 3.9 & \textbf{100} & 1.42 $\pm$ 0.10 & 292 $\pm$ 3.2 & 100 & 49.2 $\pm$ 0.088 & 162 $\pm$ 0.0 \\
{(N=4)} & KiTE & \textbf{100} & \textbf{0.539 $\pm$ 0.014} & \textbf{329 $\pm$ 2.9} & \textbf{100} & \textbf{0.107 $\pm$ 0.0063} & \textbf{197 $\pm$ 1.3} & \textbf{100} & \textbf{0.218 $\pm$ 0.013} & \textbf{179 $\pm$ 0.68} & - & - & - \\
\hline
{Swap} & Base & \textbf{100} & 29.0 $\pm$ 2.4 & 1540 $\pm$ 14 & \textbf{100} & 0.539 $\pm$ 0.022 & 401 $\pm$ 4.5 & \textbf{100} & 1.25 $\pm$ 0.067 & 364 $\pm$ 3.8 & 100 & 23.6 $\pm$ 0.099 & 233 $\pm$ 0.0 \\
{(N=5)} & KiTE & \textbf{100} & \textbf{0.933 $\pm$ 0.024} & \textbf{438 $\pm$ 4.0} & 100 & \textbf{0.104 $\pm$ 0.013} & \textbf{230 $\pm$ 1.8} & \textbf{100} & \textbf{0.192 $\pm$ 0.011} & \textbf{212 $\pm$ 0.73} & - & - & - \\
\hline
{Swap} & Base & 20 & 210 $\pm$ 13 & 3315 $\pm$ 35 & \textbf{100} & 1.30 $\pm$ 0.10 & 700 $\pm$ 5.3 & \textbf{100} & 11.1 $\pm$ 1.0 & 544 $\pm$ 3.2 & 0 & - & - \\
{(N=8)} & KiTE & \textbf{100} & \textbf{5.73 $\pm$ 0.23} & \textbf{932 $\pm$ 8.7} & 99 & \textbf{0.370 $\pm$ 0.024} & \textbf{420 $\pm$ 2.2} & \textbf{100} & \textbf{6.86 $\pm$ 0.72} & \textbf{350 $\pm$ 0.66} & - & - & - \\
\hline
{Swap} & Base & 7 & 237 $\pm$ 19 & 4837 $\pm$ 126 & 98 & 1.61 $\pm$ 0.12 & 882 $\pm$ 6.6 & 96 & 56.7 $\pm$ 5.7 & 652 $\pm$ 3.1 & 0 & - & - \\
{(N=10)} & KiTE & \textbf{100} & \textbf{9.63 $\pm$ 0.37} & \textbf{1333 $\pm$ 12} & \textbf{99} & \textbf{0.51 $\pm$ 0.04} & \textbf{547 $\pm$ 2.5} & \textbf{97} & \textbf{35.11 $\pm$ 4.3} & \textbf{437 $\pm$ 0.63} & - & - & - \\
\hline
{Swap} & Base & 0 & - & - & 72 & 8.89 $\pm$ 2.0 & 1409 $\pm$ 11 & 11 & 127 $\pm$ 25 & 943 $\pm$ 6.3 & 0 & - & - \\
{(N=15)} & KiTE & \textbf{100} & \textbf{63.9 $\pm$ 4.8} & \textbf{2661 $\pm$ 23} & \textbf{80} & \textbf{1.74 $\pm$ 0.19} & \textbf{865 $\pm$ 5.6} & \textbf{30} & \textbf{77.4 $\pm$ 9.8} & \textbf{645 $\pm$ 1.7} & - & - & - \\
\hline
{Swap} & Base & 0 & - & - & 48 & 15.3 $\pm$ 2.4 & 1968 $\pm$ 14 & 0 & - & - & 0 & - & - \\
{(N=20)} & KiTE & \textbf{14} & \textbf{268 $\pm$ 5.0} & \textbf{4638 $\pm$ 91} & \textbf{62} & \textbf{14.9 $\pm$ 6.1} & \textbf{1213 $\pm$ 7.5} & 0 & - & - & - & - & - \\
\hline
{Swap} & Base & 0 & - & - & \textbf{20} & 37.0 $\pm$ 5.3 & 2510 $\pm$ 30 & 0 & - & - & 0 & - & - \\
{(N=25)} & KiTE & 0 & - & - & 19 & \textbf{10.1 $\pm$ 1.8} & \textbf{1612 $\pm$ 18} & 0 & - & - & - & - & - \\
\hline
{Swap} & Base & 0 & - & - & 2 & 66.3 $\pm$ 20 & 3042 $\pm$ 63 & 0 & - & - & 0 & - & - \\
{(N=30)} & KiTE & 0 & - & - & \textbf{6} & \textbf{15.3 $\pm$ 5.6} & \textbf{2019 $\pm$ 27} & 0 & - & - & - & - & - \\
\hline
{Small\;Cluttered} & Base & \textbf{100} & 1.34 $\pm$ 0.090 & 422 $\pm$ 8.4 & \textbf{100} & 0.730 $\pm$ 0.049 & 163 $\pm$ 2.8 & \textbf{100} & 1.07 $\pm$ 0.087 & 158 $\pm$ 2.6 & 100 & 4.25 $\pm$ 0.028 & 56.2 $\pm$ 0.0 \\
{(N=3)} & KiTE & \textbf{100} & \textbf{0.287 $\pm$ 0.023} & \textbf{166 $\pm$ 3.5} & \textbf{100} & \textbf{0.190 $\pm$ 0.014} & \textbf{88.1 $\pm$ 1.3} & \textbf{100} & \textbf{0.217 $\pm$ 0.012} & \textbf{88.2 $\pm$ 1.3} & - & - & - \\
\hline
{Small\;Cluttered} & Base & \textbf{100} & 3.97 $\pm$ 0.24 & 696 $\pm$ 10 & \textbf{100} & 1.06 $\pm$ 0.075 & 226 $\pm$ 4.5 & \textbf{100} & 1.50 $\pm$ 0.10 & 212 $\pm$ 3.7 & 100 & 6.25 $\pm$ 0.18 & 68.5 $\pm$ 0.0 \\
{(N=4)} & KiTE & \textbf{100} & \textbf{0.646 $\pm$ 0.040} & \textbf{267 $\pm$ 6.7} & \textbf{100} & \textbf{0.264 $\pm$ 0.016} & \textbf{118 $\pm$ 2.0} & \textbf{100} & \textbf{0.326 $\pm$ 0.023} & \textbf{116 $\pm$ 1.7} & - & - & - \\
\hline
{Small\;Cluttered} & Base & \textbf{100} & 9.00 $\pm$ 0.64 & 958 $\pm$ 13 & \textbf{100} & 1.22 $\pm$ 0.061 & 290 $\pm$ 5.2 & \textbf{100} & 2.06 $\pm$ 0.11 & 260 $\pm$ 3.7 & 100 & 12.0 $\pm$ 0.067 & 92.9 $\pm$ 0.0 \\
{(N=5)} & KiTE & \textbf{100} & \textbf{1.02 $\pm$ 0.049} & \textbf{350 $\pm$ 6.6} & \textbf{100} & \textbf{0.332 $\pm$ 0.020} & \textbf{154 $\pm$ 2.6} & \textbf{100} & \textbf{0.415 $\pm$ 0.022} & \textbf{140 $\pm$ 1.6} & - & - & - \\
\hline
{Small\;Cluttered} & Base & \textbf{100} & 17.3 $\pm$ 0.81 & 1343 $\pm$ 18 & \textbf{100} & 1.43 $\pm$ 0.070 & 339 $\pm$ 5.8 & \textbf{100} & 2.35 $\pm$ 0.11 & 311 $\pm$ 4.3 & 100 & 13.3 $\pm$ 0.092 & 111 $\pm$ 0.0 \\
{(N=6)} & KiTE & \textbf{100} & \textbf{1.71 $\pm$ 0.067} & \textbf{489 $\pm$ 9.1} & \textbf{100} & \textbf{0.381 $\pm$ 0.023} & \textbf{184 $\pm$ 3.3} & \textbf{100} & \textbf{0.527 $\pm$ 0.028} & \textbf{165 $\pm$ 2.0} & - & - & - \\
\hline
{Small\;Cluttered} & Base & \textbf{100} & 38.2 $\pm$ 3.3 & 1756 $\pm$ 22 & 98 & 1.94 $\pm$ 0.16 & 403 $\pm$ 6.3 & \textbf{100} & 3.34 $\pm$ 0.18 & 339 $\pm$ 4.2 & 100 & 18.3 $\pm$ 0.067 & 127 $\pm$ 0.0 \\
{(N=7)} & KiTE & \textbf{100} & \textbf{2.91 $\pm$ 0.13} & \textbf{629 $\pm$ 12} & \textbf{100} & \textbf{1.08 $\pm$ 0.38} & \textbf{220 $\pm$ 3.3} & \textbf{100} & \textbf{0.677 $\pm$ 0.035} & \textbf{188 $\pm$ 1.8} & - & - & - \\
\hline
{Small\;Cluttered} & Base & 98 & 49.6 $\pm$ 3.3 & 2080 $\pm$ 25 & \textbf{95} & 4.64 $\pm$ 1.1 & 460 $\pm$ 6.8 & \textbf{100} & 5.49 $\pm$ 0.33 & 350 $\pm$ 4.3 & 100 & 21.2 $\pm$ 0.24 & 141 $\pm$ 0.0 \\
{(N=8)} & KiTE & \textbf{100} & \textbf{4.37 $\pm$ 0.17} & \textbf{736 $\pm$ 13} & \textbf{95} & \textbf{0.908 $\pm$ 0.095} & \textbf{252 $\pm$ 4.1} & \textbf{100} & \textbf{1.20 $\pm$ 0.068} & \textbf{197 $\pm$ 1.8} & - & - & - \\
\hline
{Small\;Cluttered} & Base & 51 & 179 $\pm$ 9.1 & 3183 $\pm$ 43 & \textbf{90} & 5.15 $\pm$ 0.61 & 607 $\pm$ 8.5 & \textbf{100} & 10.9 $\pm$ 0.66 & 429 $\pm$ 4.6 & 100 & 22.3 $\pm$ 0.83 & 183 $\pm$ 0.0 \\
{(N=10)} & KiTE & \textbf{100} & \textbf{17.6 $\pm$ 0.87} & \textbf{1162 $\pm$ 24} & \textbf{90} & \textbf{1.68 $\pm$ 0.17} & \textbf{321 $\pm$ 4.3} & \textbf{100} & \textbf{2.04 $\pm$ 0.13} & \textbf{244 $\pm$ 2.1} & - & - & - \\
\hline
{Small\;Cluttered} & Base & 1 & 204 $\pm$ 0.0 & 4206 $\pm$ 0.0 & 86 & 11.9 $\pm$ 1.6 & 745 $\pm$ 12 & \textbf{100} & 19.4 $\pm$ 1.4 & 496 $\pm$ 5.3 & 0 & - & - \\
{(N=12)} & KiTE & \textbf{100} & \textbf{31.9 $\pm$ 1.6} & \textbf{1578 $\pm$ 26} & \textbf{89} & \textbf{4.88 $\pm$ 1.6} & \textbf{402 $\pm$ 6.5} & \textbf{100} & \textbf{3.43 $\pm$ 0.18} & \textbf{288 $\pm$ 1.8} & - & - & - \\
\hline
{Small\;Cluttered} & Base & 0 & - & - & 77 & 24.2 $\pm$ 4.9 & 963 $\pm$ 13 & 98 & 66.0 $\pm$ 4.9 & 564 $\pm$ 4.7 & 100 & 224 $\pm$ 0.74 & 287 $\pm$ 0.0 \\
{(N=15)} & KiTE & \textbf{100} & \textbf{94.5 $\pm$ 4.5} & \textbf{2460 $\pm$ 45} & \textbf{79} & \textbf{7.45 $\pm$ 2.8} & \textbf{534 $\pm$ 7.0} & \textbf{100} & \textbf{9.59 $\pm$ 0.58} & \textbf{348 $\pm$ 1.9} & - & - & - \\
\hline
{Small\;Cluttered} & Base & 0 & - & - & 53 & 41.0 $\pm$ 5.8 & 1274 $\pm$ 17 & 50 & 177 $\pm$ 10 & 645 $\pm$ 5.7 & 0 & - & - \\
{(N=18)} & KiTE & \textbf{76} & \textbf{209 $\pm$ 6.4} & \textbf{3726 $\pm$ 64} & \textbf{61} & \textbf{9.81 $\pm$ 2.1} & \textbf{694 $\pm$ 9.2} & \textbf{99} & \textbf{54.6 $\pm$ 5.9} & \textbf{408 $\pm$ 1.5} & - & - & - \\
\hline
{Small\;Cluttered} & Base & 0 & - & - & 51 & 53.0 $\pm$ 7.1 & 1436 $\pm$ 20 & 15 & 192 $\pm$ 15 & 705 $\pm$ 8.9 & 0 & - & - \\
{(N=20)} & KiTE & \textbf{7} & \textbf{261 $\pm$ 5.1} & \textbf{3809 $\pm$ 139} & \textbf{61} & \textbf{31.1 $\pm$ 8.6} & \textbf{808 $\pm$ 11} & \textbf{87} & \textbf{97.6 $\pm$ 7.8} & \textbf{450 $\pm$ 1.4} & - & - & - \\
\hline
{Large\;Cluttered} & Base & \textbf{100} & 25.8 $\pm$ 1.8 & 1837 $\pm$ 18 & 89 & 7.45 $\pm$ 3.9 & 527 $\pm$ 7.1 & \textbf{100} & 3.59 $\pm$ 0.25 & 482 $\pm$ 6.8 & 100 & 78.5 $\pm$ 0.14 & 209 $\pm$ 0.0 \\
{(N=4)} & KiTE & \textbf{100} & \textbf{1.27 $\pm$ 0.060} & \textbf{624 $\pm$ 8.1} & \textbf{97} & \textbf{1.06 $\pm$ 0.41} & \textbf{313 $\pm$ 3.6} & \textbf{100} & \textbf{0.934 $\pm$ 0.086} & \textbf{287 $\pm$ 2.4} & - & - & - \\
\hline
{Large\;Cluttered} & Base & 95 & 82.4 $\pm$ 6.1 & 2476 $\pm$ 23 & 87 & 4.67 $\pm$ 1.2 & 644 $\pm$ 8.0 & \textbf{100} & 3.89 $\pm$ 0.22 & 588 $\pm$ 6.0 & 100 & 96.5 $\pm$ 0.26 & 266 $\pm$ 0.0 \\
{(N=5)} & KiTE & \textbf{100} & \textbf{2.97 $\pm$ 0.18} & \textbf{853 $\pm$ 13} & \textbf{89} & \textbf{2.84 $\pm$ 1.9} & \textbf{379 $\pm$ 4.1} & \textbf{100} & \textbf{1.23 $\pm$ 0.097} & \textbf{340 $\pm$ 3.1} & - & - & - \\
\hline
{Large\;Cluttered} & Base & 3 & 279 $\pm$ 5.6 & 4805 $\pm$ 128 & 67 & 12.5 $\pm$ 4.1 & 985 $\pm$ 12 & \textbf{100} & 10.3 $\pm$ 0.73 & 833 $\pm$ 6.7 & 100 & 149 $\pm$ 0.98 & 400 $\pm$ 0.0 \\
{(N=8)} & KiTE & \textbf{100} & \textbf{10.9 $\pm$ 0.72} & \textbf{1642 $\pm$ 23} & \textbf{85} & \textbf{2.35 $\pm$ 0.37} & \textbf{587 $\pm$ 5.5} & \textbf{100} & \textbf{2.78 $\pm$ 0.31} & \textbf{500 $\pm$ 3.1} & - & - & - \\
\hline
{Large\;Cluttered} & Base & 0 & - & - & 70 & 8.19 $\pm$ 1.8 & 1355 $\pm$ 16 & \textbf{100} & 18.6 $\pm$ 1.2 & 1044 $\pm$ 8.0 & 100 & 202 $\pm$ 0.22 & 517 $\pm$ 0.0 \\
{(N=10)} & KiTE & \textbf{100} & \textbf{27.6 $\pm$ 1.3} & \textbf{2501 $\pm$ 34} & \textbf{83} & \textbf{2.38 $\pm$ 0.36} & \textbf{788 $\pm$ 7.2} & \textbf{100} & \textbf{6.01 $\pm$ 0.68} & \textbf{621 $\pm$ 3.5} & - & - & - \\
\hline
{Large\;Cluttered} & Base & 0 & - & - & 21 & 35.1 $\pm$ 12 & 1855 $\pm$ 25 & 97 & 52.5 $\pm$ 3.3 & 1374 $\pm$ 8.6 & 0 & - & - \\
{(N=15)} & KiTE & \textbf{100} & \textbf{110 $\pm$ 4.4} & \textbf{4565 $\pm$ 78} & \textbf{38} & \textbf{11.0 $\pm$ 3.1} & \textbf{1114 $\pm$ 14} & \textbf{100} & \textbf{16.1 $\pm$ 1.3} & \textbf{829 $\pm$ 2.9} & - & - & - \\
\hline
{Large\;Cluttered} & Base & 0 & - & - & 6 & 30.3 $\pm$ 4.1 & 2557 $\pm$ 132 & 90 & 106 $\pm$ 6.9 & 1753 $\pm$ 10 & 0 & - & - \\
{(N=20)} & KiTE & \textbf{43} & \textbf{248 $\pm$ 4.3} & \textbf{7353 $\pm$ 146} & \textbf{13} & \textbf{28.3 $\pm$ 9.1} & \textbf{1477 $\pm$ 30} & \textbf{100} & \textbf{38.7 $\pm$ 3.8} & \textbf{1070 $\pm$ 4.6} & - & - & - \\
\hline
{Large\;Cluttered} & Base & 0 & - & - & 3 & 40.5 $\pm$ 16 & 3357 $\pm$ 121 & 46 & 180 $\pm$ 11 & 2232 $\pm$ 10 & 0 & - & - \\
{(N=25)} & KiTE & 0 & - & - & \textbf{8} & \textbf{29.7 $\pm$ 9.7} & \textbf{2024 $\pm$ 56} & \textbf{94} & \textbf{74.5 $\pm$ 5.6} & \textbf{1384 $\pm$ 3.9} & - & - & - \\
\hline
{Large\;Cluttered} & Base & 0 & - & - & 4 & 98.3 $\pm$ 26 & 4284 $\pm$ 64 & 4 & 237 $\pm$ 28 & 2702 $\pm$ 44 & 0 & - & - \\
{(N=30)} & KiTE & 0 & - & - & \textbf{6} & \textbf{17.4 $\pm$ 2.9} & \textbf{2390 $\pm$ 81} & \textbf{65} & \textbf{140 $\pm$ 7.5} & \textbf{1637 $\pm$ 3.6} & - & - & - \\
\hline
\bottomrule
\end{tabular}
}

\vspace{0.5ex}
{\scriptsize SR: Success Rate \textbar{} CT: Computation Time \textbar{} PT: Total Path Time}
\label{table_unicycle_complete_results}
\end{center}
\end{table*}

\FloatBarrier

\FloatBarrier

\clearpage
\subsection{Results for the Second Order Car model}

\begin{figure*}[t!]
\centering

\imgrow{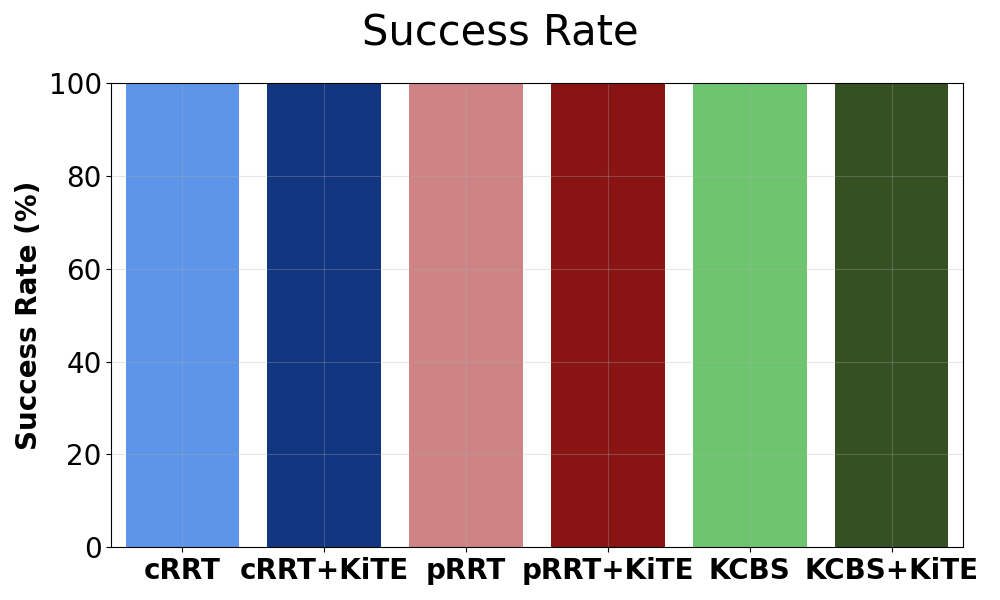}{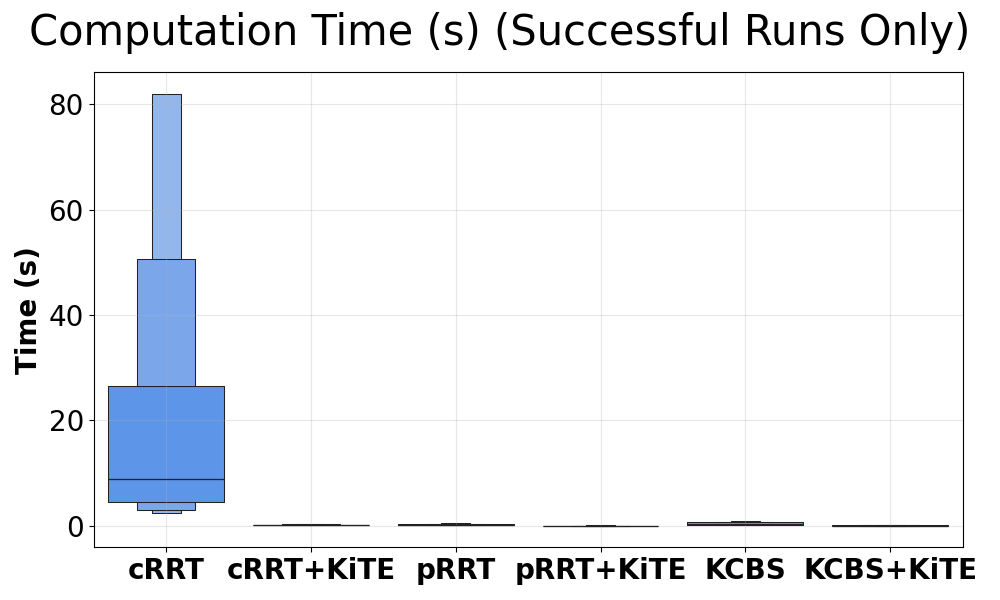}{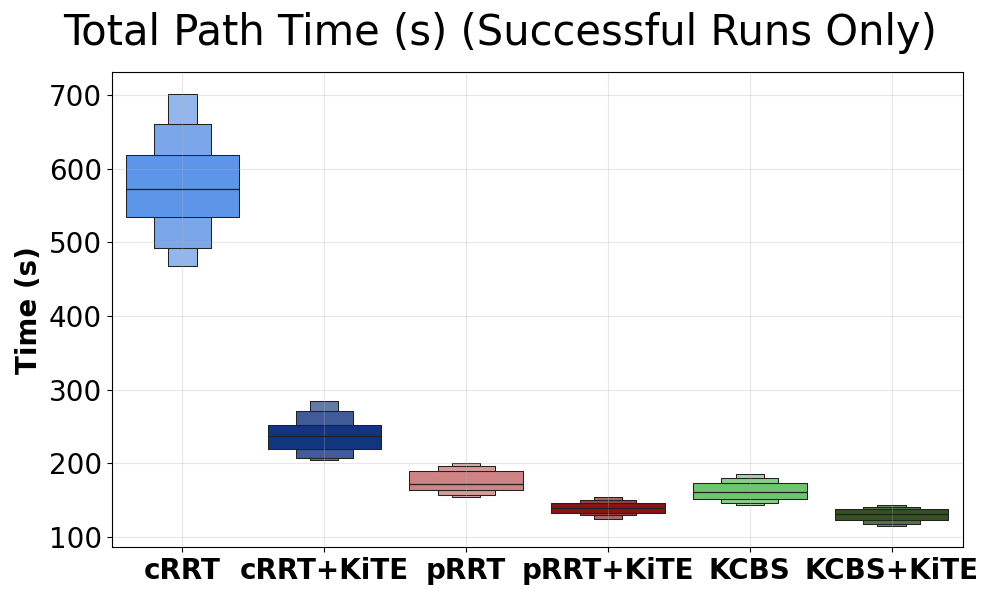}
\rowsubcap{SWAP}{\(N=4\) robots}

\imgrow{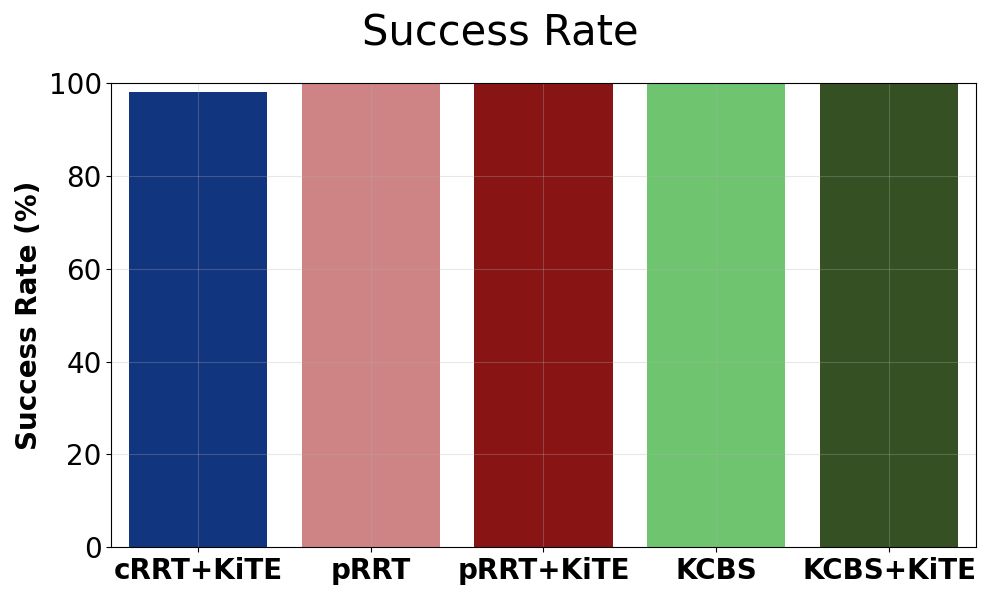}{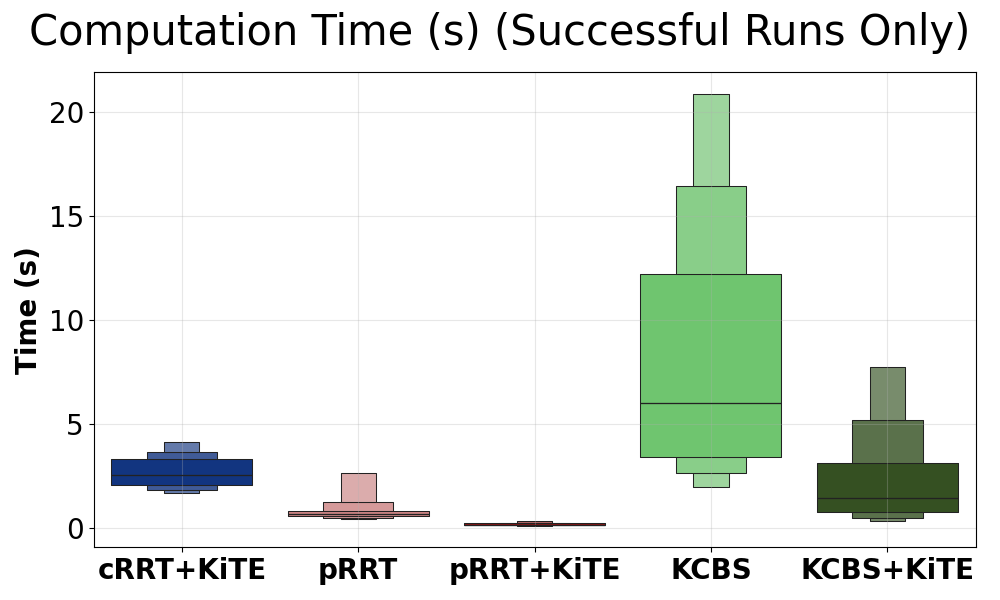}{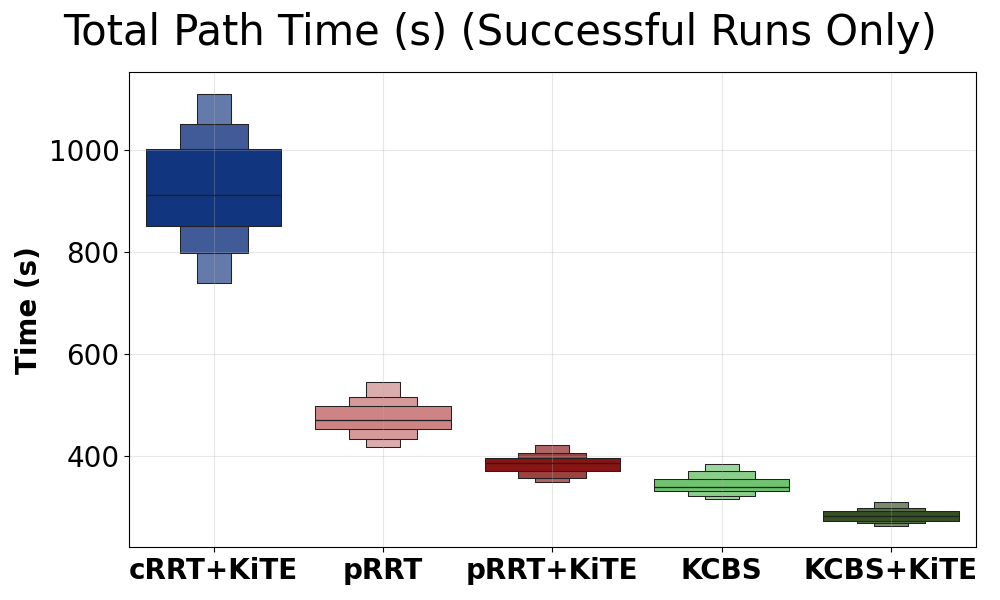}
\rowsubcap{SWAP}{\(N=10\) robots}

\imgrow{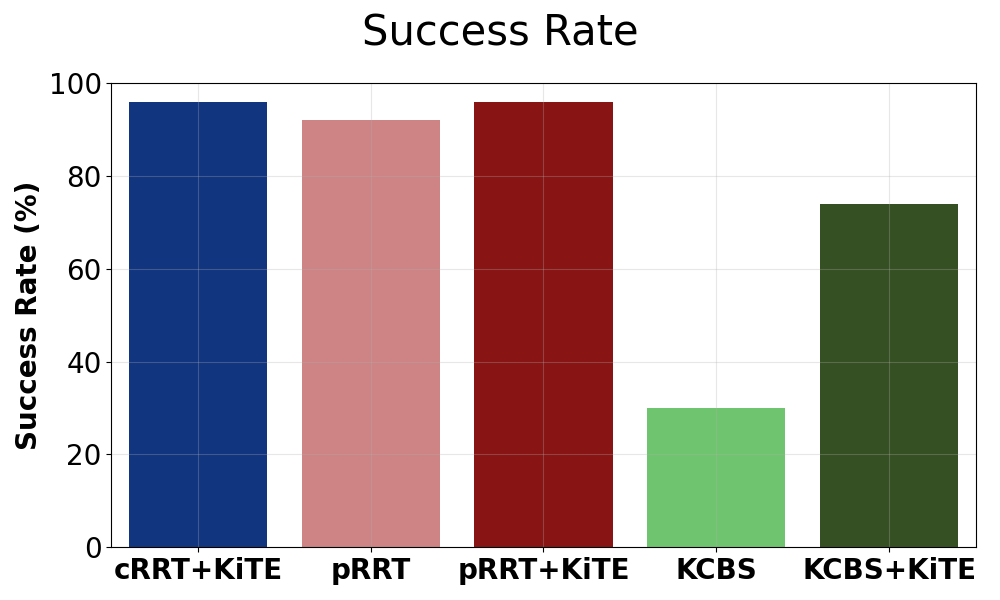}{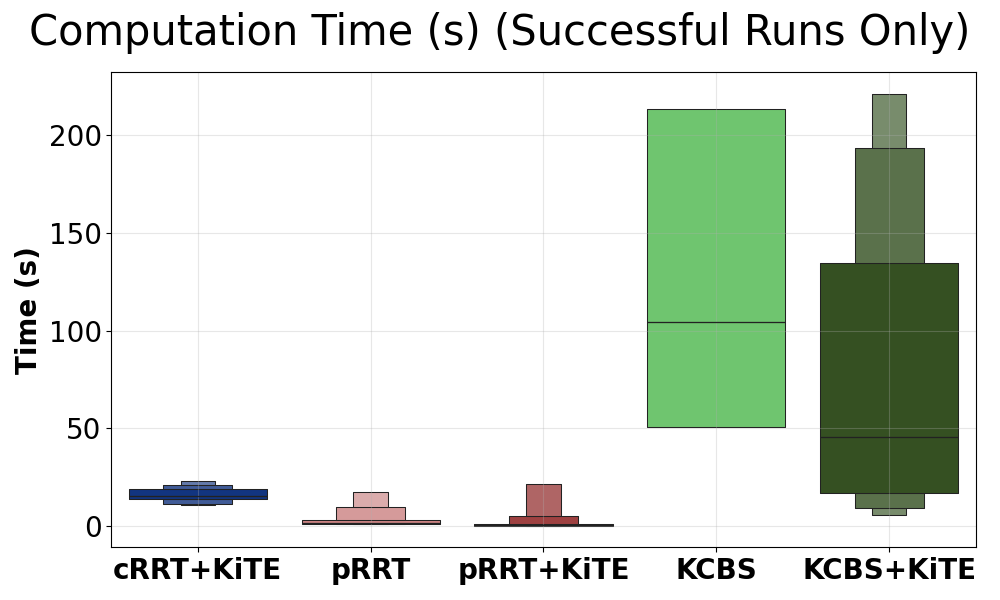}{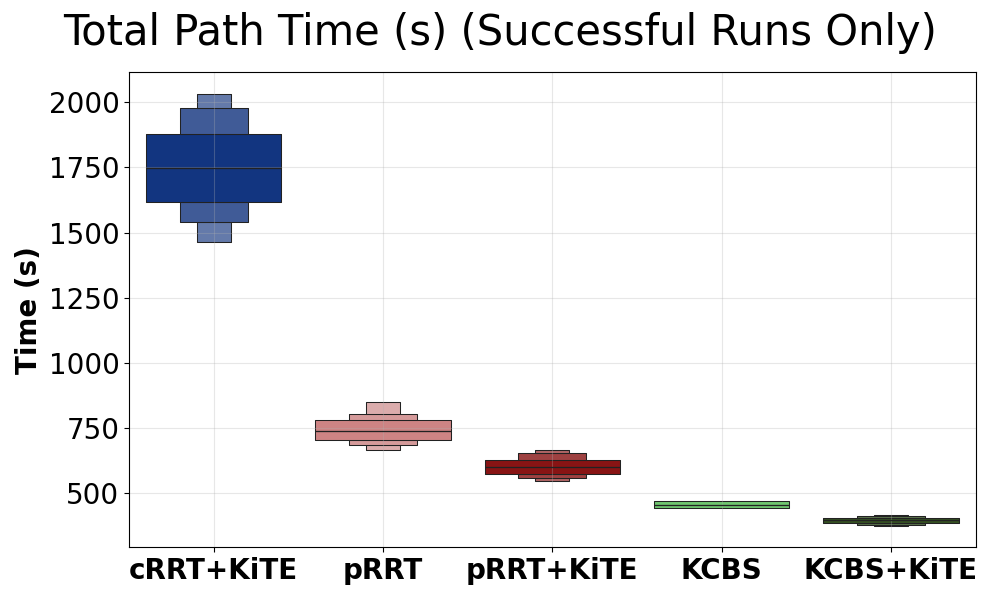}
\rowsubcap{SWAP}{\(N=15\) robots}

\imgrow{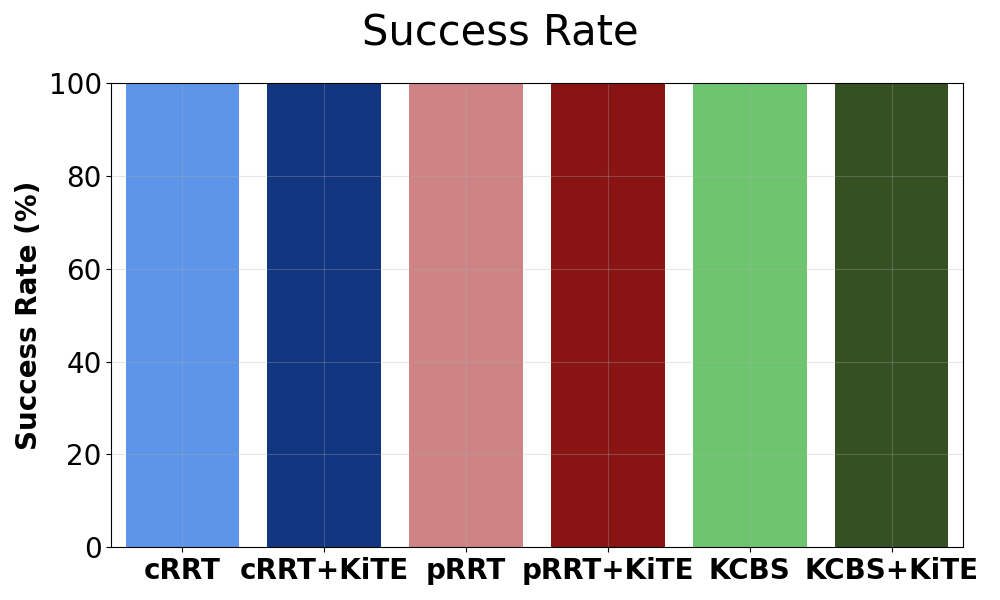}{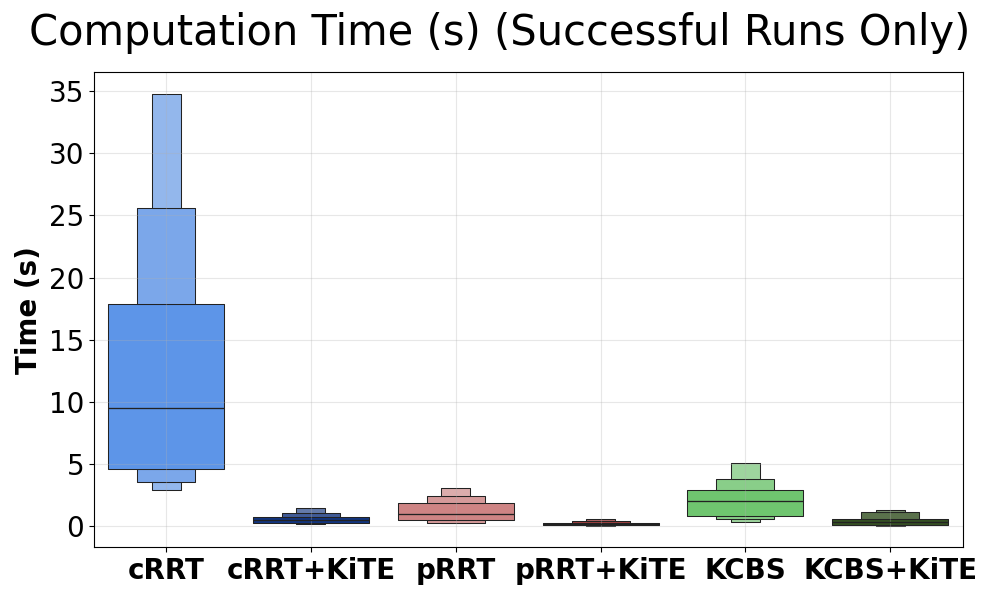}{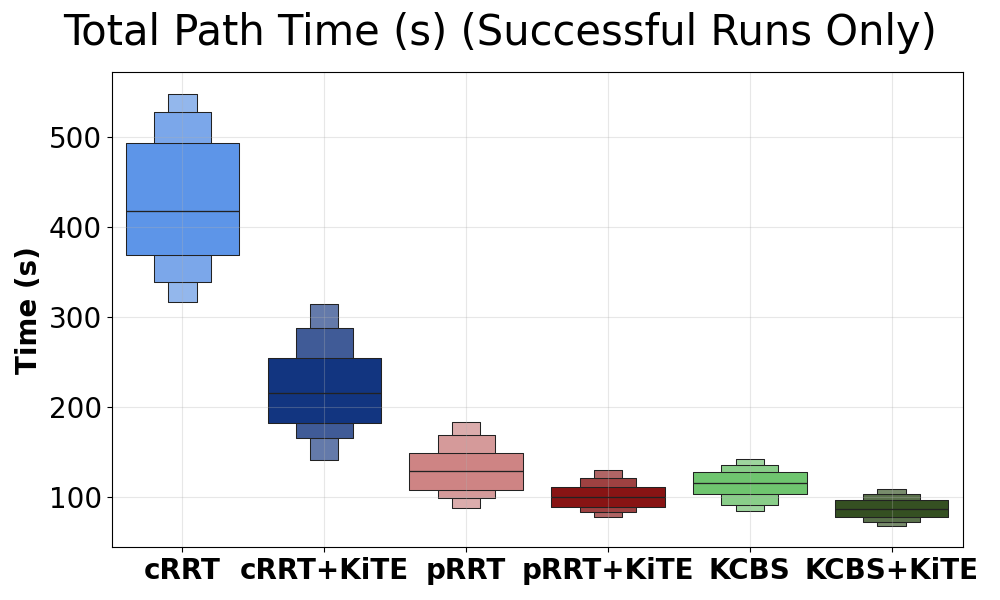}
\rowsubcap{Small Cluttered}{\(N=4\) robots}

\imgrow{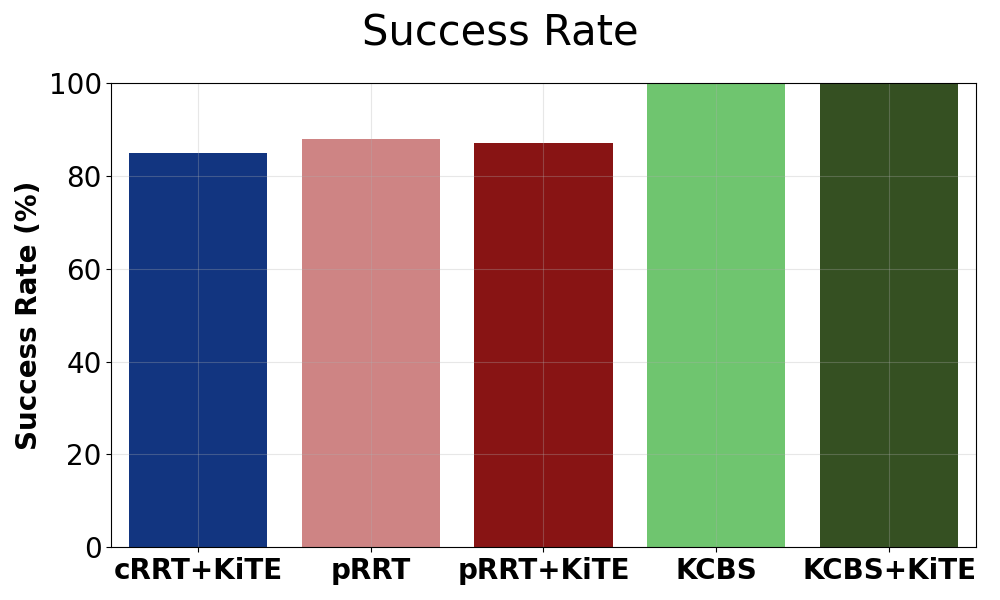}{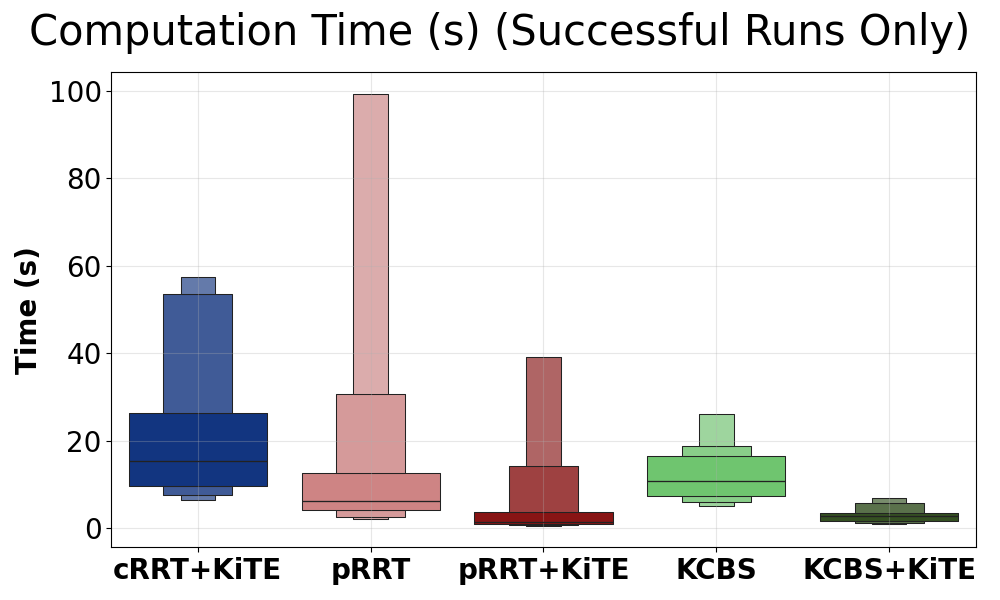}{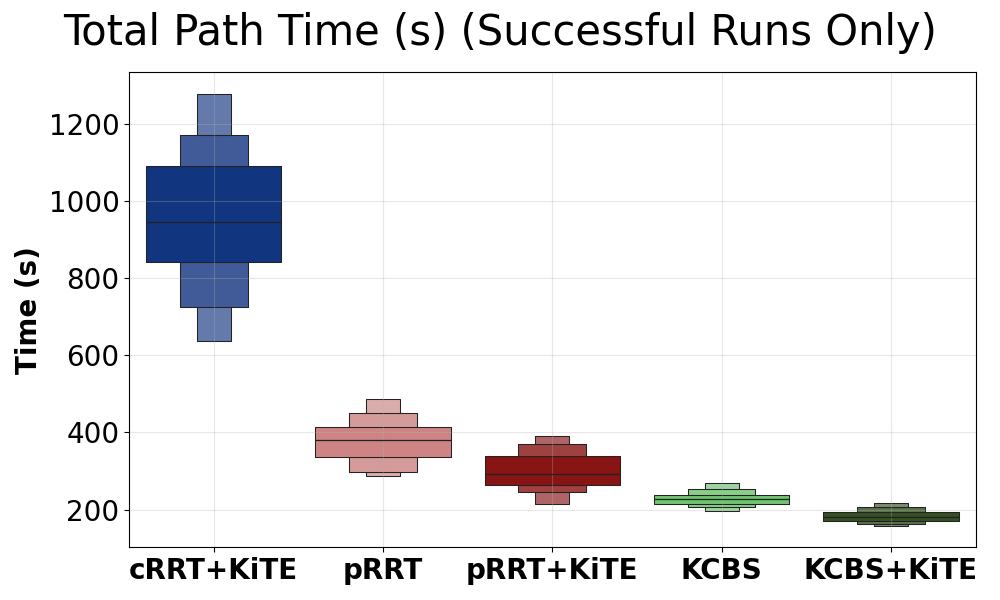}
\rowsubcap{Small Cluttered}{\(N=10\) robots}

\caption{
Detailed results for selected environments in experiments with the Second Order Car model.
Each row corresponds to one experimental setting (environment and agent count \(N\)) and contains, from left to right:
(i) success rate (SR; bar chart), (ii) computation time (CT; boxen plot), and (iii) total path time (PT; boxen plot),
comparing each baseline planner to its KiTE-Extend variant.
These plots visualize variability and tail behavior underlying the aggregate results in Table~\ref{table_soc_complete_results}.
}
\label{fig:soc_rows_1_5}
\end{figure*}

\begin{figure*}[t!]
\centering

\imgrow{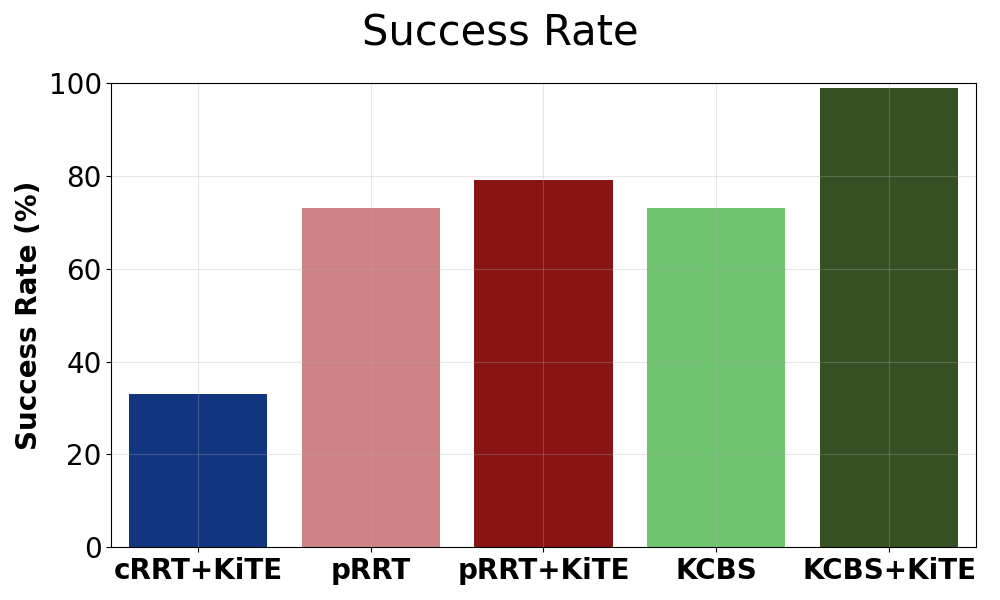}{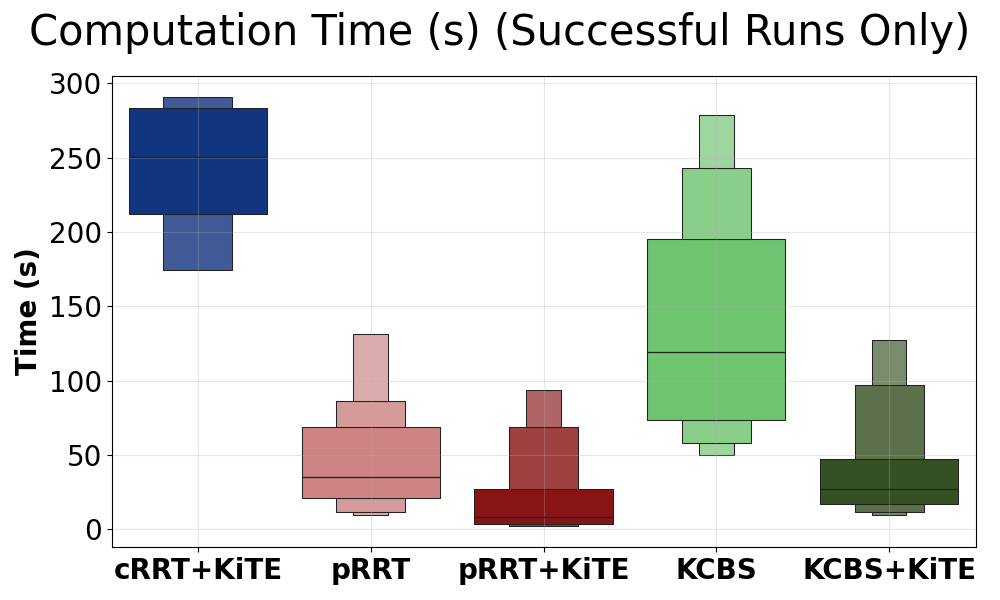}{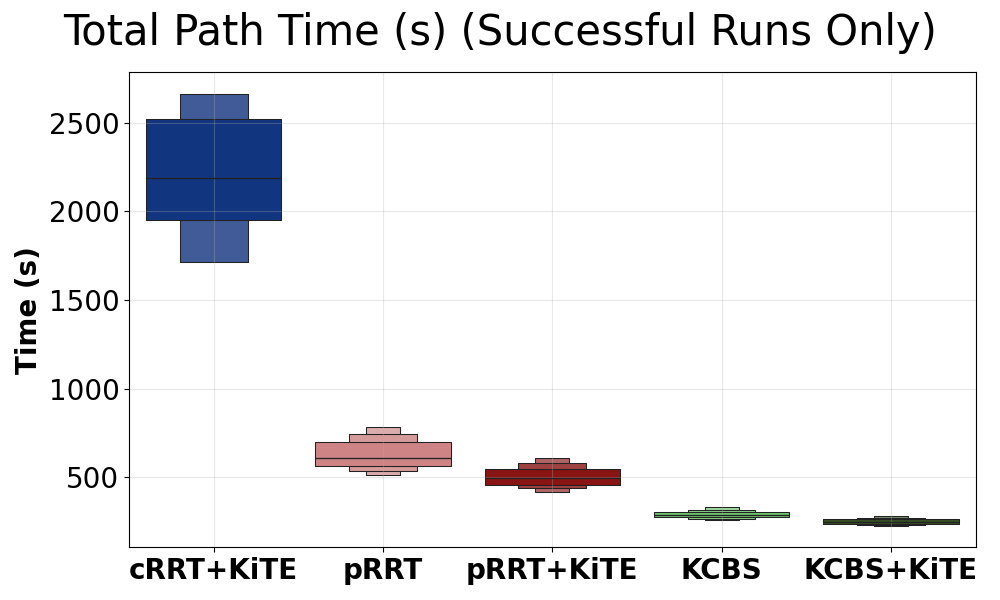}
\rowsubcap{Small Cluttered}{\(N=15\) robots}

\imgrow{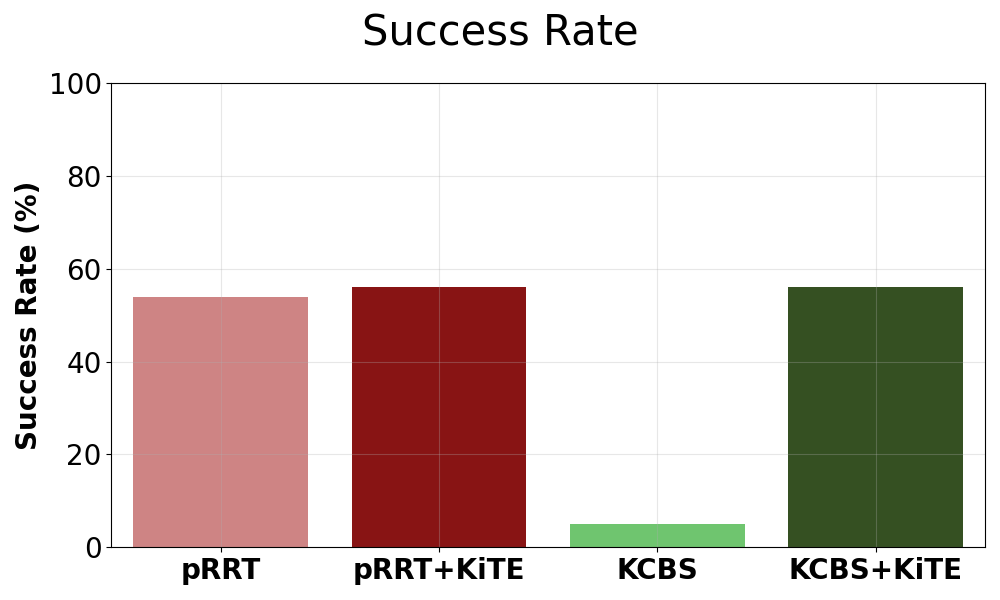}{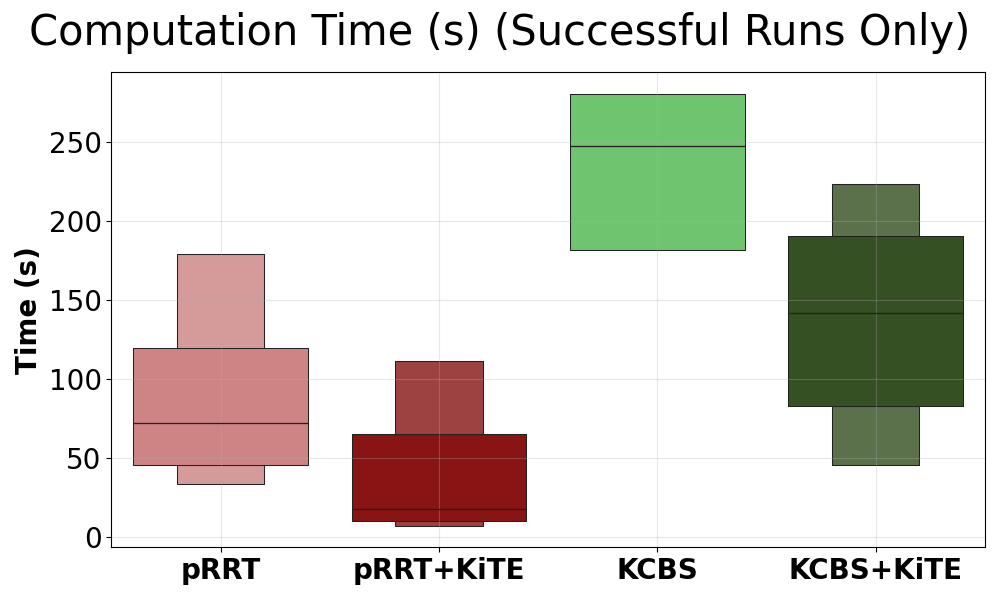}{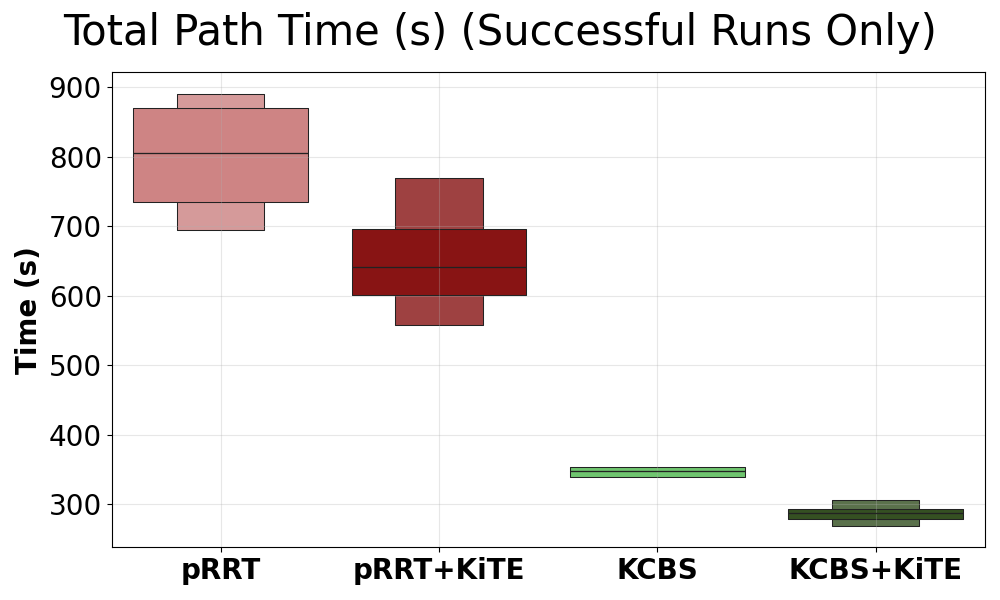}
\rowsubcap{Small Cluttered}{\(N=18\) robots}

\imgrow{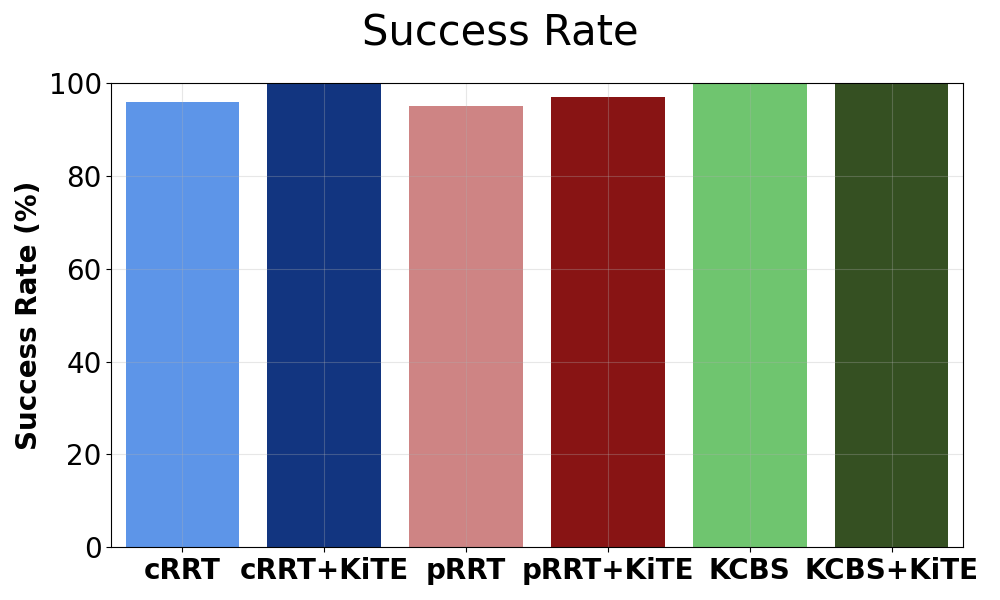}{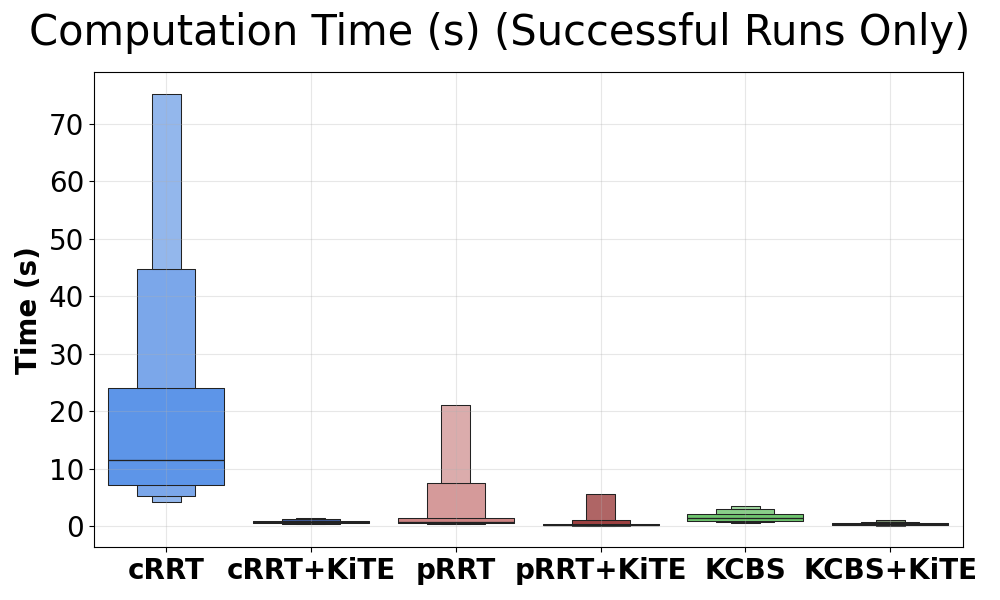}{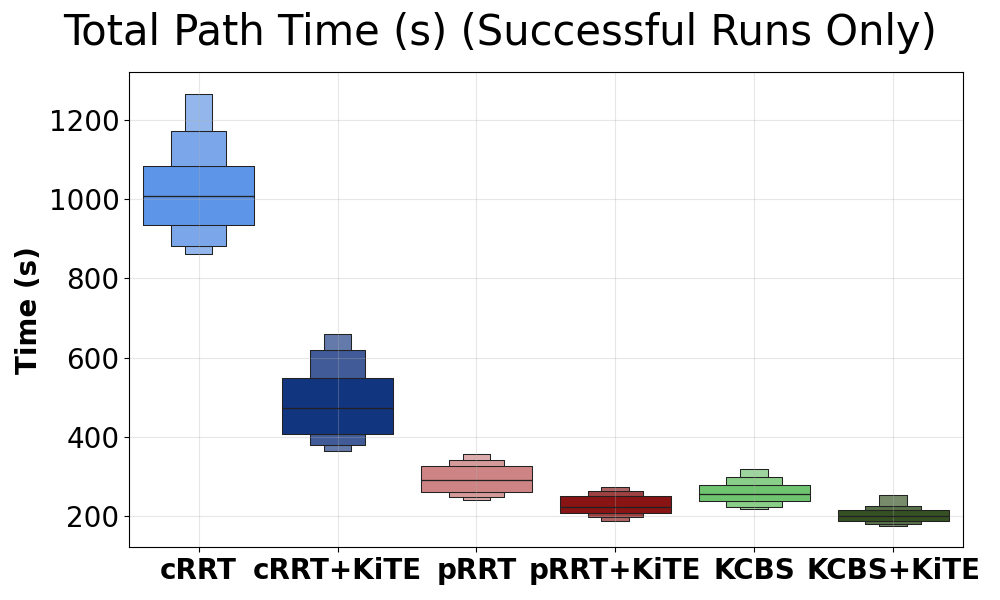}
\rowsubcap{Large Cluttered}{\(N=4\) robots}

\imgrow{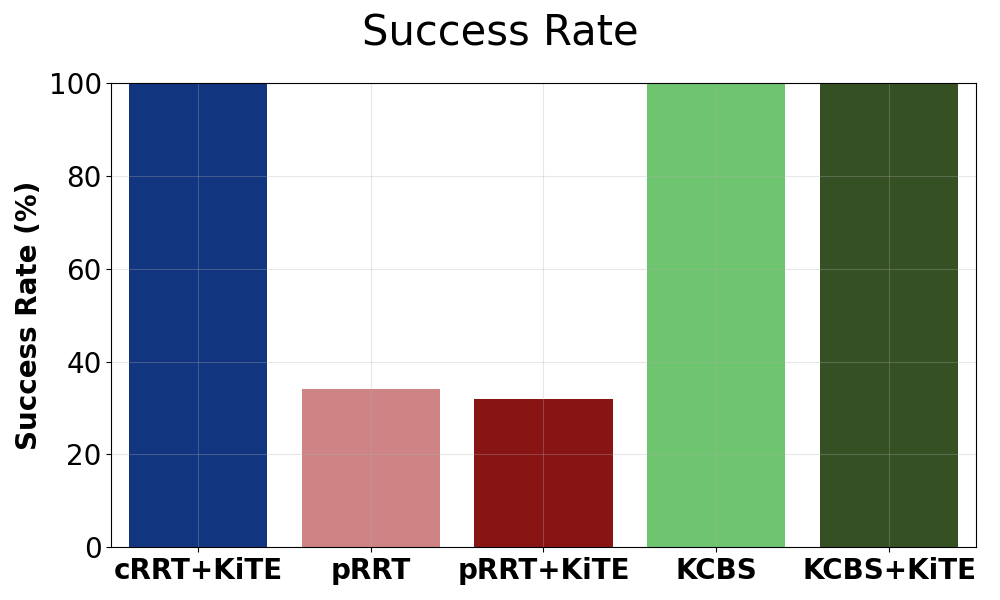}{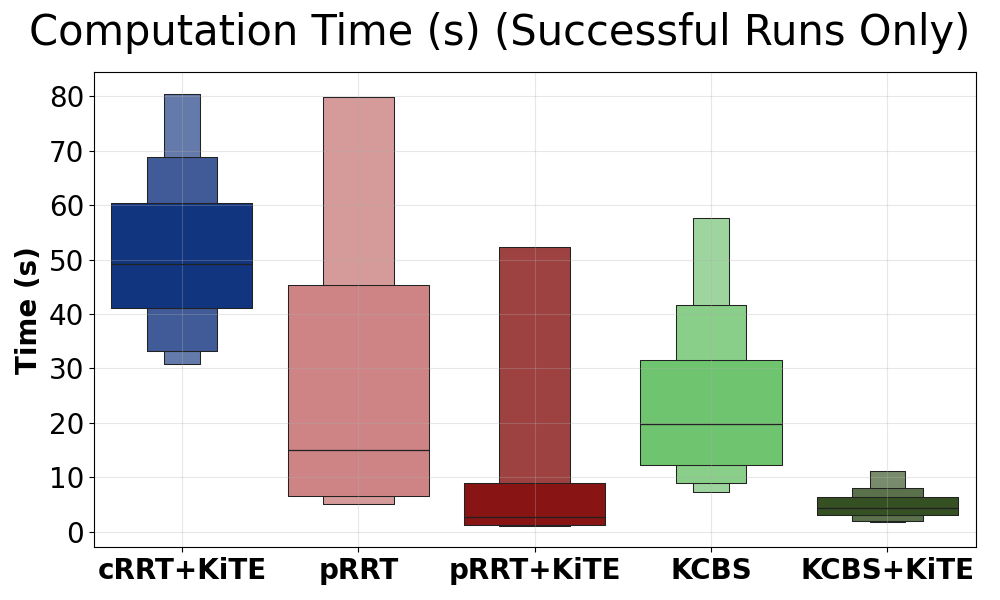}{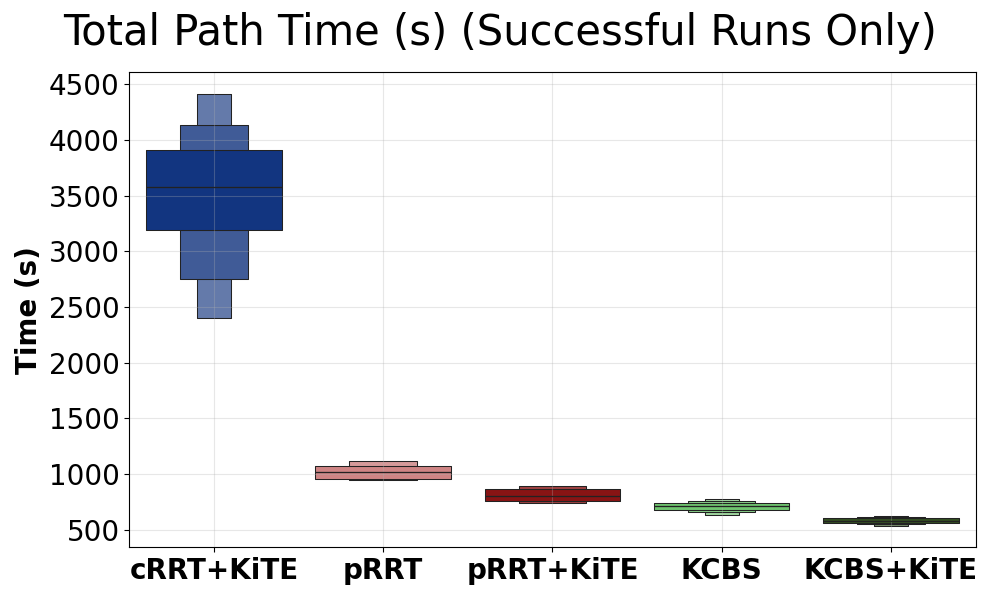}
\rowsubcap{Large Cluttered}{\(N=15\) robots}

\imgrow{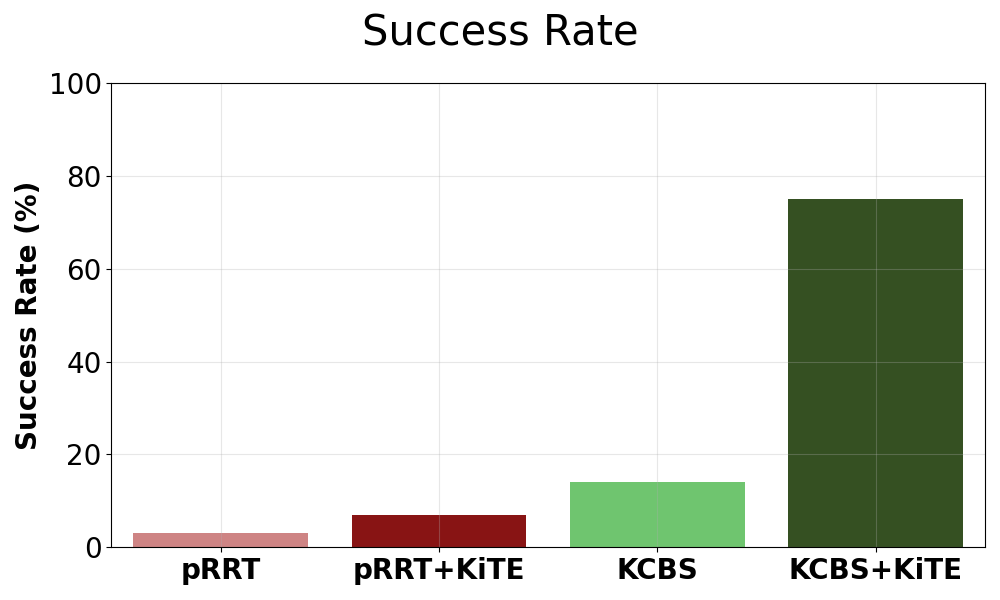}{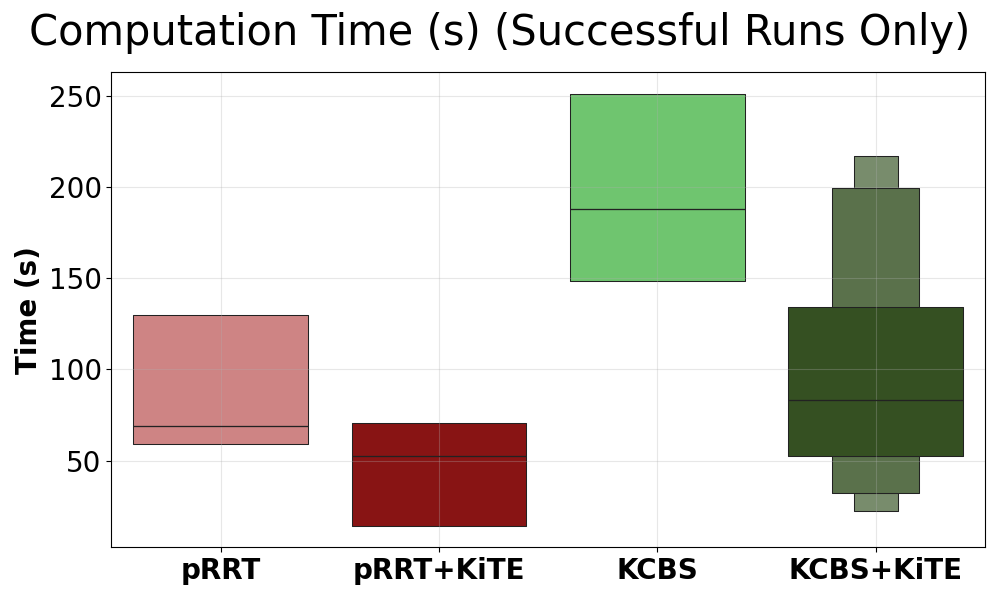}{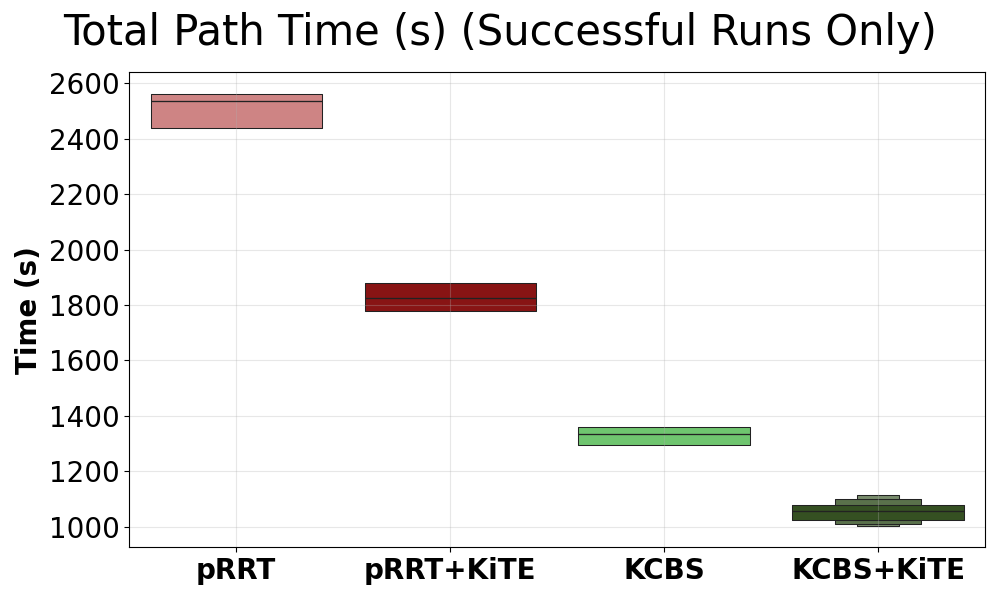}
\rowsubcap{Large Cluttered}{\(N=30\) robots}

\caption{
Detailed results for selected environments in experiments with the Second Order Car model (continued).
Each row corresponds to one experimental setting (environment and agent count \(N\)) and contains, from left to right:
(i) success rate (SR; bar chart), (ii) computation time (CT; boxen plot), and (iii) total path time (PT; boxen plot),
comparing each baseline planner to its KiTE-Extend variant.
These plots visualize variability and tail behavior underlying the aggregate results in Table~\ref{table_soc_complete_results}.
}
\label{fig:soc_rows_6_10}
\end{figure*}

Table~\ref{table_soc_complete_results} reports complete results for the Second Order Car (SOC) model across all environments and planners.
KiTE-Extend consistently extends successful planning to higher robot counts for sampling-based planners, with the most pronounced gains observed for cRRT, where it frequently converts failure into success across swap and cluttered environments.
For pRRT and KCBS, KiTE-Extend yields substantial reductions in computation time and total path duration while largely preserving success rates, and in several high-density settings extends successful planning relative to the baseline.
Overall, the observed trends closely mirror those seen for the Unicycle model, indicating that KiTE’s benefits persist under more challenging second-order dynamics and are robust to both environment complexity and robot count.
To complement the aggregate statistics reported in Table~\ref{table_soc_complete_results}, 
Fig.~\ref{fig:soc_rows_1_5} and Fig.~\ref{fig:soc_rows_6_10} present results for representative environments in experiments with the Second Order Car model, including success rate (bar charts) as well as computation time and total path time (boxen plots).

\begin{table*}[!ht]
\caption{\small
\textbf{Planner performance for the Second Order Car (SOC) model across environments and robot counts.}
Results compare baseline planners and their KiTE-Extend variants for cRRT, pRRT, KCBS, and dbCBS, reported in terms of success rate (SR), computation time (CT), and total path time (PT).
Bold entries indicate the best value for each metric within a given planner and environment; for success rate, all entries achieving the maximum value are bolded.}
\begin{center}

\resizebox{1\linewidth}{!}{%

\begin{tabular}{l|c|ccc|ccc|ccc|ccc}
\toprule
\textbf{Domain} & \textbf{Variant}
& \multicolumn{3}{c|}{\textbf{cRRT}}
& \multicolumn{3}{c|}{\textbf{pRRT}}
& \multicolumn{3}{c|}{\textbf{KCBS}}\\

& 
& SR (\%) & CT (s) & PT (s)
& SR (\%) & CT (s) & PT (s)
& SR (\%) & CT (s) & PT (s)\\[1mm]
\hline
\hline

{Narrow\;Corridor} & Base & 64 & 102 $\pm$ 9.9 & \textbf{106 $\pm$ 3.3} & \textbf{42} & 51.7 $\pm$ 7.8 & \textbf{56.4 $\pm$ 2.3} & 81 & 113 $\pm$ 9.3 & \textbf{48.3 $\pm$ 1.3} \\
{(N=3)} & KiTE & \textbf{92} & \textbf{46.9 $\pm$ 5.4} & 128 $\pm$ 4.3 & 39 & \textbf{32.3 $\pm$ 7.7} & 71.5 $\pm$ 3.2 & \textbf{85} & \textbf{94.8 $\pm$ 9.1} & 53.6 $\pm$ 1.7 \\
\hline
{Swap} & Base & \textbf{100} & 22.3 $\pm$ 3.2 & 579 $\pm$ 7.3 & \textbf{100} & 0.234 $\pm$ 0.017 & 175 $\pm$ 1.6 & \textbf{100} & 0.459 $\pm$ 0.033 & 162 $\pm$ 1.4 \\
{(N=4)} & KiTE & \textbf{100} & \textbf{0.261 $\pm$ 0.043} & \textbf{239 $\pm$ 3.0} & \textbf{100} & \textbf{0.0440 $\pm$ 0.0023} & \textbf{140 $\pm$ 0.97} & \textbf{100} & \textbf{0.0792 $\pm$ 0.0069} & \textbf{130 $\pm$ 0.99} \\
\hline
{Swap} & Base & 86 & 53.2 $\pm$ 6.2 & 852 $\pm$ 11 & \textbf{100} & 0.251 $\pm$ 0.015 & 211 $\pm$ 2.4 & \textbf{100} & 0.451 $\pm$ 0.026 & 190 $\pm$ 1.7 \\
{(N=5)} & KiTE & \textbf{100} & \textbf{0.342 $\pm$ 0.025} & \textbf{323 $\pm$ 4.2} & \textbf{100} & \textbf{0.0492 $\pm$ 0.0024} & \textbf{165 $\pm$ 1.4} & \textbf{100} & \textbf{0.0891 $\pm$ 0.0057} & \textbf{154 $\pm$ 1.3} \\
\hline
{Swap} & Base & 1 & 298 $\pm$ 0.0 & 1730 $\pm$ 0.0 & \textbf{100} & \textbf{0.515 $\pm$ 0.024} & 372 $\pm$ 3.1 & \textbf{100} & 2.60 $\pm$ 0.17 & 292 $\pm$ 2.1 \\
{(N=8)} & KiTE & \textbf{100} & \textbf{1.22 $\pm$ 0.072} & \textbf{663 $\pm$ 7.5} & \textbf{100} & 1.00 $\pm$ 0.84 & \textbf{301 $\pm$ 2.1} & \textbf{100} & \textbf{0.651 $\pm$ 0.062} & \textbf{235 $\pm$ 1.5} \\
\hline
{Swap} & Base & 0 & - & - & \textbf{100} & 4.73 $\pm$ 2.9 & 474 $\pm$ 3.8 & \textbf{100} & 9.13 $\pm$ 0.98 & 344 $\pm$ 2.4 \\
{(N=10)} & KiTE & \textbf{98} & \textbf{2.80 $\pm$ 0.14} & \textbf{922 $\pm$ 11} & \textbf{100} & \textbf{0.809 $\pm$ 0.44} & \textbf{384 $\pm$ 2.4} & \textbf{100} & \textbf{2.55 $\pm$ 0.28} & \textbf{283 $\pm$ 1.4} \\
\hline
{Swap} & Base & 0 & - & - & 92 & \textbf{6.64 $\pm$ 2.0} & 744 $\pm$ 5.8 & 30 & 128 $\pm$ 17 & 457 $\pm$ 4.1 \\
{(N=15)} & KiTE & \textbf{96} & \textbf{16.3 $\pm$ 0.44} & \textbf{1754 $\pm$ 21} & \textbf{96} & 7.61 $\pm$ 3.4 & \textbf{603 $\pm$ 4.2} & \textbf{74} & \textbf{80.0 $\pm$ 9.0} & \textbf{395 $\pm$ 1.7} \\
\hline
{Swap} & Base & 0 & - & - & 72 & 26.1 $\pm$ 6.4 & 1080 $\pm$ 9.3 & 0 & - & - \\
{(N=20)} & KiTE & \textbf{38} & \textbf{229 $\pm$ 6.3} & \textbf{2746 $\pm$ 55} & \textbf{91} & \textbf{17.0 $\pm$ 4.5} & \textbf{878 $\pm$ 4.9} & 0 & - & - \\
\hline
{Swap} & Base & 0 & - & - & 48 & 66.7 $\pm$ 11 & 1418 $\pm$ 10.0 & 0 & - & - \\
{(N=25)} & KiTE & 0 & - & - & \textbf{56} & \textbf{42.2 $\pm$ 8.4} & \textbf{1165 $\pm$ 8.9} & 0 & - & - \\
\hline
{Swap} & Base & 0 & - & - & \textbf{25} & 118 $\pm$ 17 & 1868 $\pm$ 22 & 0 & - & - \\
{(N=30)} & KiTE & 0 & - & - & 23 & \textbf{79.5 $\pm$ 19} & \textbf{1507 $\pm$ 19} & 0 & - & - \\
\hline
{Small\;Cluttered} & Base & \textbf{100} & 2.03 $\pm$ 0.17 & 243 $\pm$ 5.4 & \textbf{100} & 0.861 $\pm$ 0.089 & 92.5 $\pm$ 1.6 & \textbf{100} & 1.32 $\pm$ 0.14 & 86.2 $\pm$ 1.6 \\
{(N=3)} & KiTE & \textbf{100} & \textbf{0.254 $\pm$ 0.018} & \textbf{138 $\pm$ 4.6} & \textbf{100} & \textbf{0.159 $\pm$ 0.018} & \textbf{67.1 $\pm$ 1.1} & \textbf{100} & \textbf{0.293 $\pm$ 0.040} & \textbf{64.3 $\pm$ 1.1} \\
\hline
{Small\;Cluttered} & Base & \textbf{100} & 13.2 $\pm$ 1.2 & 428 $\pm$ 8.0 & \textbf{100} & 1.34 $\pm$ 0.11 & 132 $\pm$ 3.1 & \textbf{100} & 2.17 $\pm$ 0.16 & 116 $\pm$ 2.0 \\
{(N=4)} & KiTE & \textbf{100} & \textbf{0.639 $\pm$ 0.049} & \textbf{223 $\pm$ 5.5} & \textbf{100} & \textbf{0.258 $\pm$ 0.019} & \textbf{102 $\pm$ 1.8} & \textbf{100} & \textbf{0.479 $\pm$ 0.042} & \textbf{87.6 $\pm$ 1.4} \\
\hline
{Small\;Cluttered} & Base & 95 & 55.9 $\pm$ 6.5 & 635 $\pm$ 11 & \textbf{100} & 1.41 $\pm$ 0.098 & 173 $\pm$ 3.1 & \textbf{100} & 2.53 $\pm$ 0.18 & 149 $\pm$ 2.3 \\
{(N=5)} & KiTE & \textbf{100} & \textbf{4.54 $\pm$ 1.8} & \textbf{321 $\pm$ 8.6} & \textbf{100} & \textbf{0.349 $\pm$ 0.050} & \textbf{133 $\pm$ 2.6} & \textbf{100} & \textbf{0.649 $\pm$ 0.047} & \textbf{113 $\pm$ 1.9} \\
\hline
{Small\;Cluttered} & Base & 52 & 83.4 $\pm$ 9.7 & 856 $\pm$ 23 & \textbf{100} & 3.82 $\pm$ 1.7 & 224 $\pm$ 4.9 & \textbf{100} & 2.90 $\pm$ 0.16 & 176 $\pm$ 2.5 \\
{(N=6)} & KiTE & \textbf{100} & \textbf{4.96 $\pm$ 1.0} & \textbf{439 $\pm$ 12} & 99 & \textbf{0.533 $\pm$ 0.055} & \textbf{168 $\pm$ 3.2} & \textbf{100} & \textbf{0.751 $\pm$ 0.046} & \textbf{134 $\pm$ 2.3} \\
\hline
{Small\;Cluttered} & Base & 17 & 137 $\pm$ 18 & 1149 $\pm$ 26 & \textbf{100} & 4.30 $\pm$ 1.1 & 259 $\pm$ 5.1 & \textbf{100} & 4.09 $\pm$ 0.23 & 187 $\pm$ 2.5 \\
{(N=7)} & KiTE & \textbf{96} & \textbf{11.9 $\pm$ 3.3} & \textbf{520 $\pm$ 14} & 99 & \textbf{3.79 $\pm$ 1.5} & \textbf{198 $\pm$ 3.7} & \textbf{100} & \textbf{1.01 $\pm$ 0.050} & \textbf{140 $\pm$ 1.5} \\
\hline
{Small\;Cluttered} & Base & 1 & 221 $\pm$ 0.0 & 1668 $\pm$ 0.0 & 95 & 7.27 $\pm$ 1.2 & 278 $\pm$ 5.3 & \textbf{100} & 7.13 $\pm$ 0.39 & 188 $\pm$ 2.4 \\
{(N=8)} & KiTE & \textbf{93} & \textbf{13.0 $\pm$ 2.3} & \textbf{643 $\pm$ 16} & \textbf{97} & \textbf{5.68 $\pm$ 2.1} & \textbf{219 $\pm$ 3.5} & \textbf{100} & \textbf{1.72 $\pm$ 0.12} & \textbf{146 $\pm$ 1.6} \\
\hline
{Small\;Cluttered} & Base & 0 & - & - & \textbf{88} & 22.3 $\pm$ 5.1 & 378 $\pm$ 6.8 & \textbf{100} & 12.6 $\pm$ 0.68 & 228 $\pm$ 2.3 \\
{(N=10)} & KiTE & \textbf{85} & \textbf{25.9 $\pm$ 3.4} & \textbf{959 $\pm$ 23} & 87 & \textbf{9.45 $\pm$ 2.6} & \textbf{302 $\pm$ 5.9} & \textbf{100} & \textbf{3.15 $\pm$ 0.19} & \textbf{183 $\pm$ 1.9} \\
\hline
{Small\;Cluttered} & Base & 0 & - & - & \textbf{86} & 30.4 $\pm$ 4.1 & 455 $\pm$ 6.4 & \textbf{100} & 36.5 $\pm$ 3.1 & 260 $\pm$ 2.3 \\
{(N=12)} & KiTE & \textbf{91} & \textbf{68.9 $\pm$ 5.8} & \textbf{1449 $\pm$ 36} & 85 & \textbf{16.4 $\pm$ 5.0} & \textbf{368 $\pm$ 5.9} & \textbf{100} & \textbf{8.25 $\pm$ 0.69} & \textbf{207 $\pm$ 1.4} \\
\hline
{Small\;Cluttered} & Base & 0 & - & - & 73 & 50.4 $\pm$ 5.5 & 632 $\pm$ 11 & 73 & 141 $\pm$ 9.2 & 290 $\pm$ 2.6 \\
{(N=15)} & KiTE & \textbf{33} & \textbf{244 $\pm$ 8.1} & \textbf{2211 $\pm$ 71} & \textbf{79} & \textbf{26.0 $\pm$ 4.5} & \textbf{504 $\pm$ 6.7} & \textbf{99} & \textbf{43.2 $\pm$ 4.1} & \textbf{250 $\pm$ 1.9} \\
\hline
{Small\;Cluttered} & Base & 0 & - & - & 54 & 94.9 $\pm$ 8.9 & 802 $\pm$ 11 & 5 & 229 $\pm$ 27 & 343 $\pm$ 5.7 \\
{(N=18)} & KiTE & 0 & - & - & \textbf{56} & \textbf{46.1 $\pm$ 7.5} & \textbf{652 $\pm$ 11} & \textbf{56} & \textbf{140 $\pm$ 9.7} & \textbf{287 $\pm$ 1.9} \\
\hline
{Small\;Cluttered} & Base & 0 & - & - & 43 & 116 $\pm$ 11 & 936 $\pm$ 15 & 0 & - & - \\
{(N=20)} & KiTE & 0 & - & - & \textbf{53} & \textbf{59.0 $\pm$ 7.5} & \textbf{758 $\pm$ 15} & \textbf{15} & \textbf{156 $\pm$ 17} & \textbf{317 $\pm$ 4.3} \\
\hline
{Large\;Cluttered} & Base & 96 & 26.8 $\pm$ 4.5 & 1018 $\pm$ 13 & 95 & 9.38 $\pm$ 3.5 & 293 $\pm$ 4.0 & \textbf{100} & 1.69 $\pm$ 0.10 & 260 $\pm$ 3.4 \\
{(N=4)} & KiTE & \textbf{100} & \textbf{0.874 $\pm$ 0.095} & \textbf{483 $\pm$ 9.3} & \textbf{97} & \textbf{5.38 $\pm$ 3.0} & \textbf{229 $\pm$ 3.3} & \textbf{100} & \textbf{0.424 $\pm$ 0.036} & \textbf{205 $\pm$ 2.7} \\
\hline
{Large\;Cluttered} & Base & 63 & 70.9 $\pm$ 9.0 & 1371 $\pm$ 19 & \textbf{97} & 7.36 $\pm$ 2.3 & 340 $\pm$ 4.1 & \textbf{100} & 2.03 $\pm$ 0.13 & 301 $\pm$ 3.2 \\
{(N=5)} & KiTE & \textbf{100} & \textbf{1.15 $\pm$ 0.056} & \textbf{615 $\pm$ 9.9} & 94 & \textbf{3.16 $\pm$ 1.9} & \textbf{270 $\pm$ 3.7} & \textbf{100} & \textbf{0.534 $\pm$ 0.048} & \textbf{244 $\pm$ 2.6} \\
\hline
{Large\;Cluttered} & Base & 1 & 171 $\pm$ 0.0 & 2837 $\pm$ 0.0 & \textbf{84} & 17.4 $\pm$ 4.9 & 534 $\pm$ 6.9 & \textbf{100} & 4.67 $\pm$ 0.32 & 432 $\pm$ 3.4 \\
{(N=8)} & KiTE & \textbf{100} & \textbf{4.04 $\pm$ 0.18} & \textbf{1217 $\pm$ 22} & 83 & \textbf{4.43 $\pm$ 2.0} & \textbf{418 $\pm$ 4.4} & \textbf{100} & \textbf{1.18 $\pm$ 0.082} & \textbf{345 $\pm$ 2.9} \\
\hline
{Large\;Cluttered} & Base & 0 & - & - & \textbf{77} & 18.6 $\pm$ 5.7 & 733 $\pm$ 8.9 & \textbf{100} & 11.9 $\pm$ 0.86 & 527 $\pm$ 4.6 \\
{(N=10)} & KiTE & \textbf{100} & \textbf{10.4 $\pm$ 0.46} & \textbf{1845 $\pm$ 35} & 76 & \textbf{5.60 $\pm$ 2.7} & \textbf{571 $\pm$ 6.9} & \textbf{100} & \textbf{2.59 $\pm$ 0.22} & \textbf{436 $\pm$ 3.2} \\
\hline
{Large\;Cluttered} & Base & 0 & - & - & \textbf{34} & 42.3 $\pm$ 12 & 1031 $\pm$ 17 & \textbf{100} & 26.5 $\pm$ 2.3 & 711 $\pm$ 5.2 \\
{(N=15)} & KiTE & \textbf{100} & \textbf{51.9 $\pm$ 1.7} & \textbf{3529 $\pm$ 63} & 32 & \textbf{27.1 $\pm$ 11} & \textbf{819 $\pm$ 17} & \textbf{100} & \textbf{5.13 $\pm$ 0.31} & \textbf{580 $\pm$ 3.1} \\
\hline
{Large\;Cluttered} & Base & 0 & - & - & \textbf{14} & \textbf{42.7 $\pm$ 19} & 1373 $\pm$ 30 & 95 & 65.2 $\pm$ 5.3 & 890 $\pm$ 5.5 \\
{(N=20)} & KiTE & \textbf{30} & \textbf{228 $\pm$ 8.2} & \textbf{5666 $\pm$ 157} & 12 & 52.7 $\pm$ 24 & \textbf{1084 $\pm$ 25} & \textbf{100} & \textbf{11.3 $\pm$ 0.90} & \textbf{728 $\pm$ 4.1} \\
\hline
{Large\;Cluttered} & Base & 0 & - & - & 9 & 79.1 $\pm$ 27 & 1857 $\pm$ 41 & 71 & 124 $\pm$ 8.0 & 1121 $\pm$ 5.1 \\
{(N=25)} & KiTE & 0 & - & - & \textbf{11} & \textbf{41.8 $\pm$ 21} & \textbf{1466 $\pm$ 33} & \textbf{98} & \textbf{33.7 $\pm$ 3.5} & \textbf{915 $\pm$ 4.0} \\
\hline
{Large\;Cluttered} & Base & 0 & - & - & 3 & 103 $\pm$ 36 & 2487 $\pm$ 62 & 14 & 201 $\pm$ 15 & 1325 $\pm$ 12 \\
{(N=30)} & KiTE & 0 & - & - & \textbf{7} & \textbf{43.5 $\pm$ 11} & \textbf{1821 $\pm$ 31} & \textbf{75} & \textbf{102 $\pm$ 8.0} & \textbf{1055 $\pm$ 4.4} \\
\hline
\bottomrule
\end{tabular}
}

\vspace{0.5ex}
{\scriptsize SR: Success Rate \textbar{} CT: Computation Time \textbar{} PT: Total Path Time
}
\label{table_soc_complete_results}
\end{center}
\end{table*}

\FloatBarrier

\FloatBarrier

\clearpage
\subsection{Results for the Double Integrator model}

\begin{figure*}[t!]
\centering

\imgrow{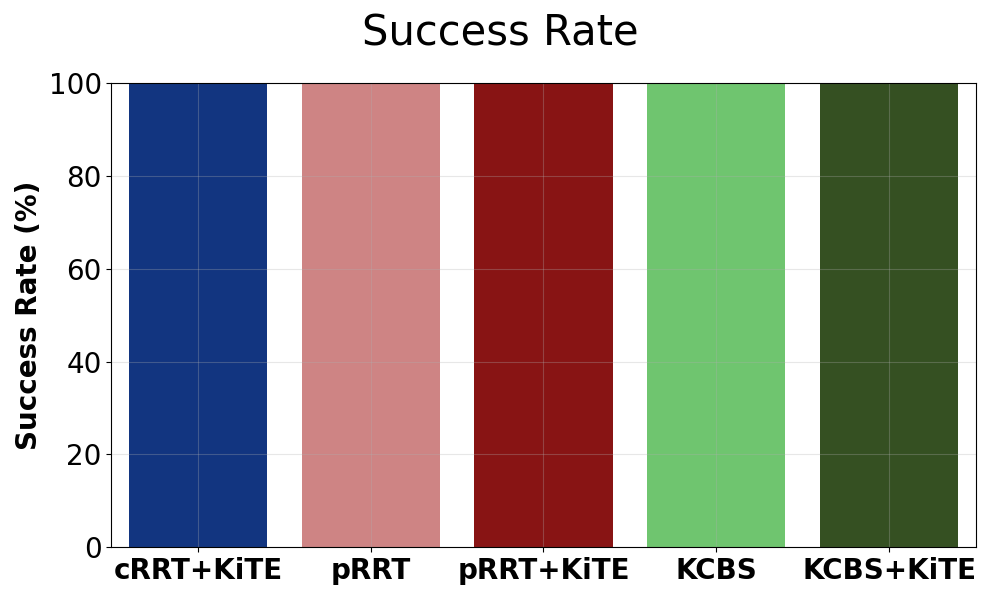}{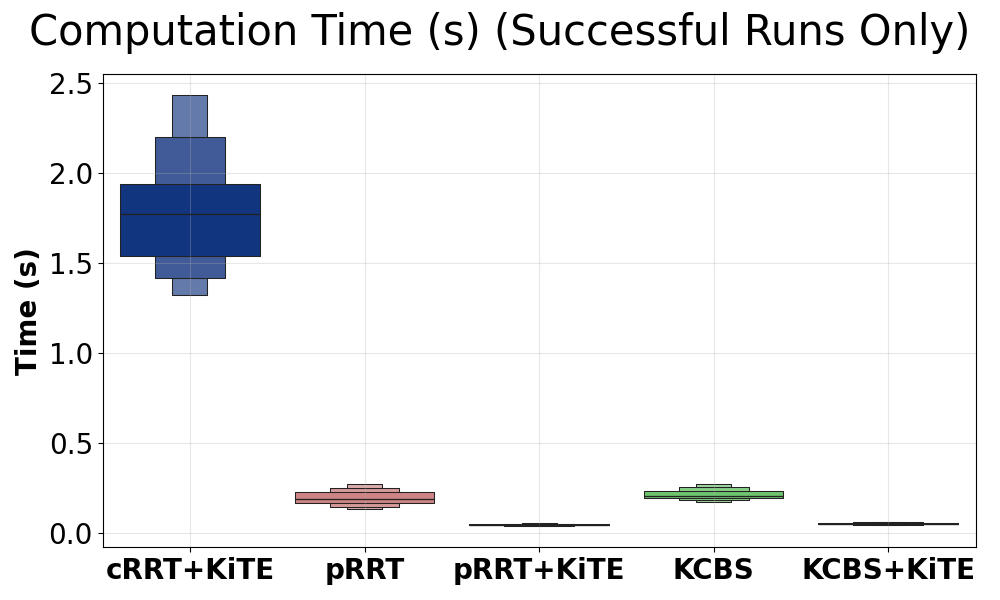}{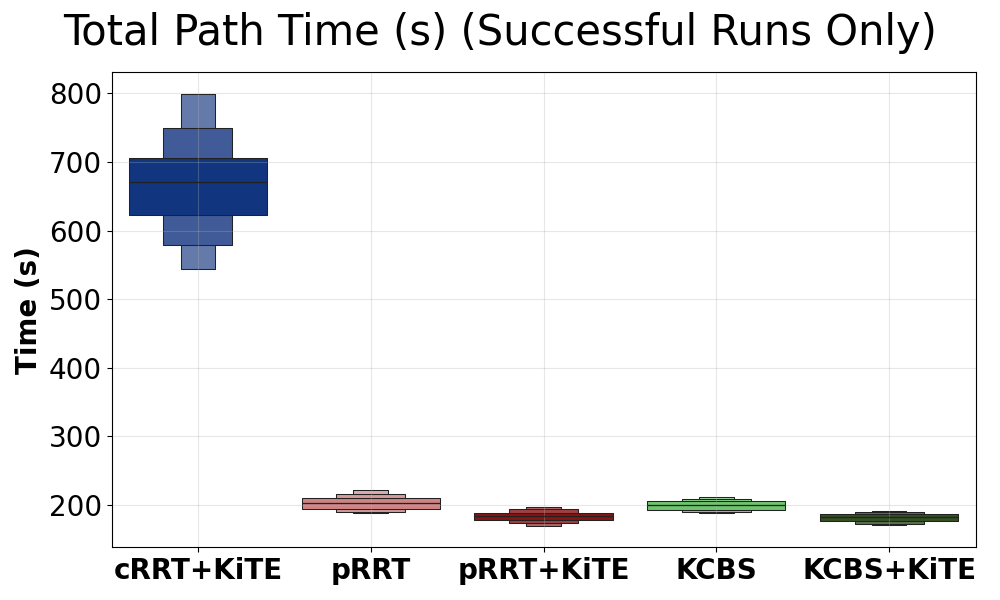}
\rowsubcap{SWAP 3D}{\(N=15\) robots}

\imgrow{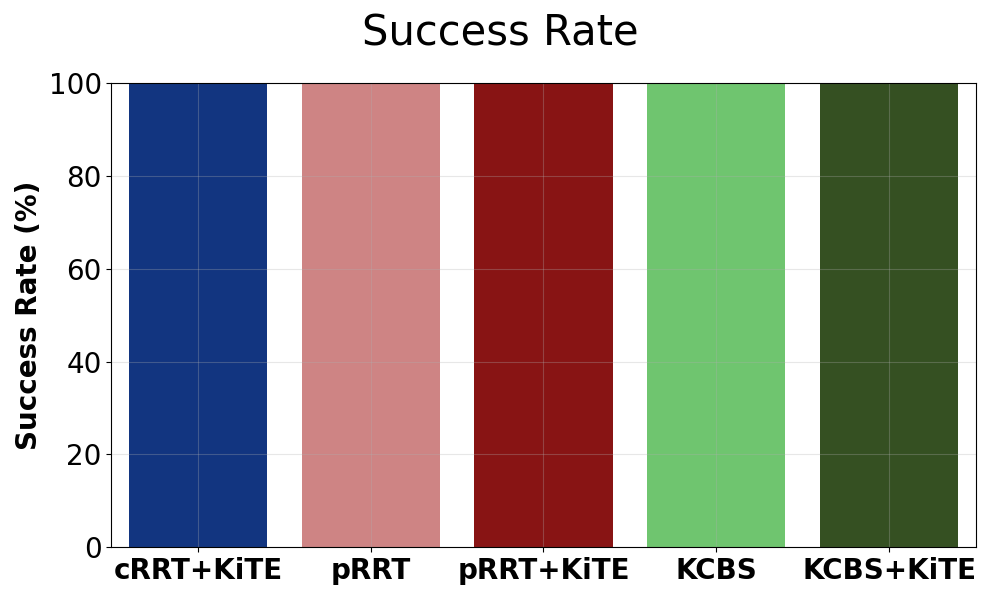}{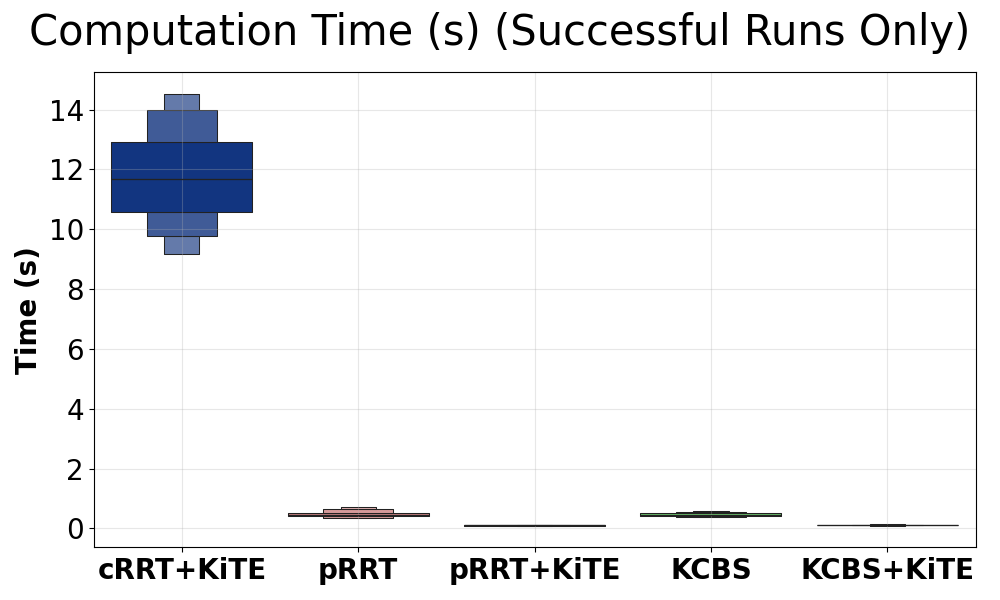}{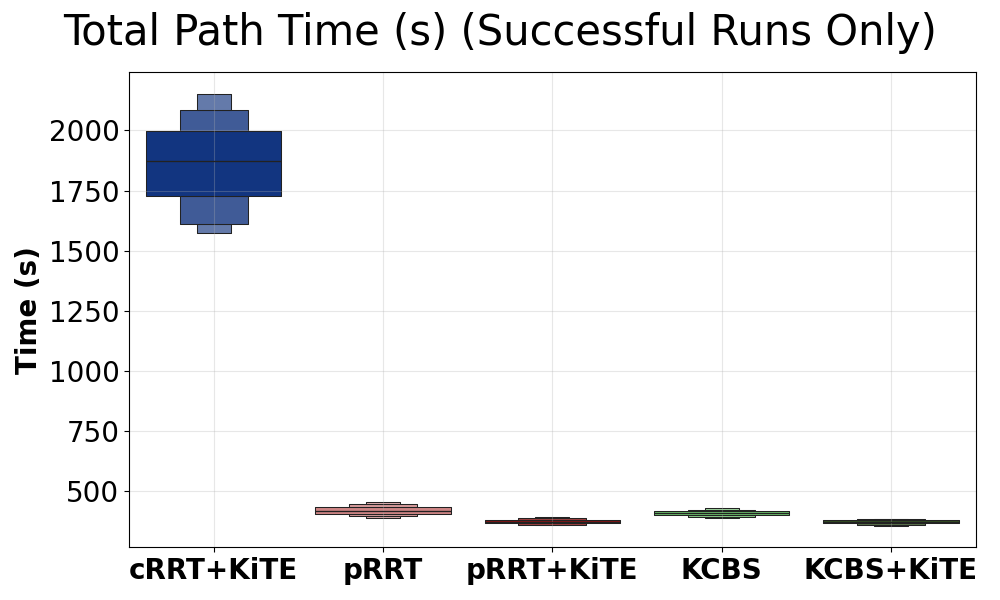}
\rowsubcap{SWAP 3D}{\(N=30\) robots}

\imgrow{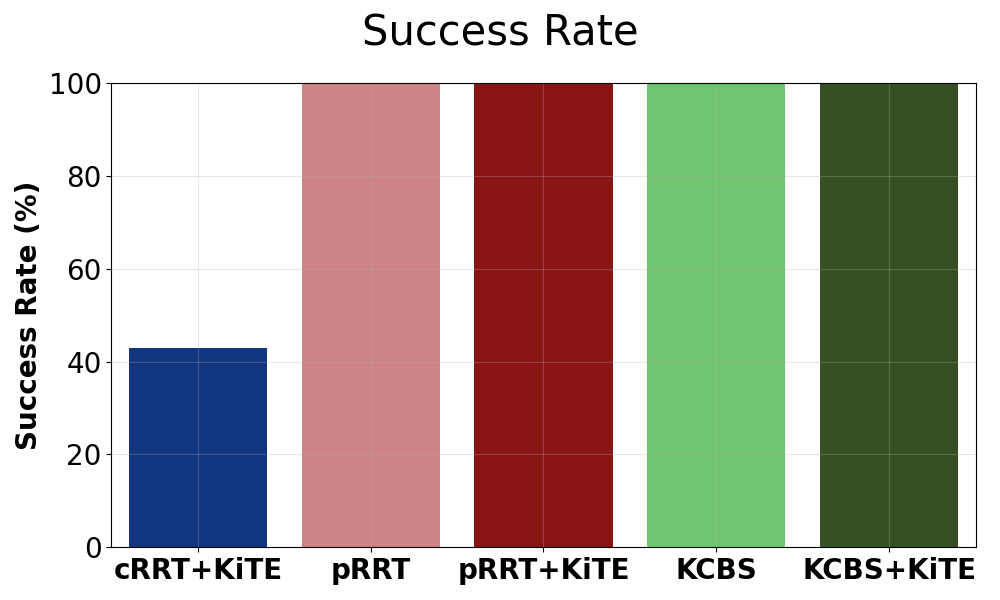}{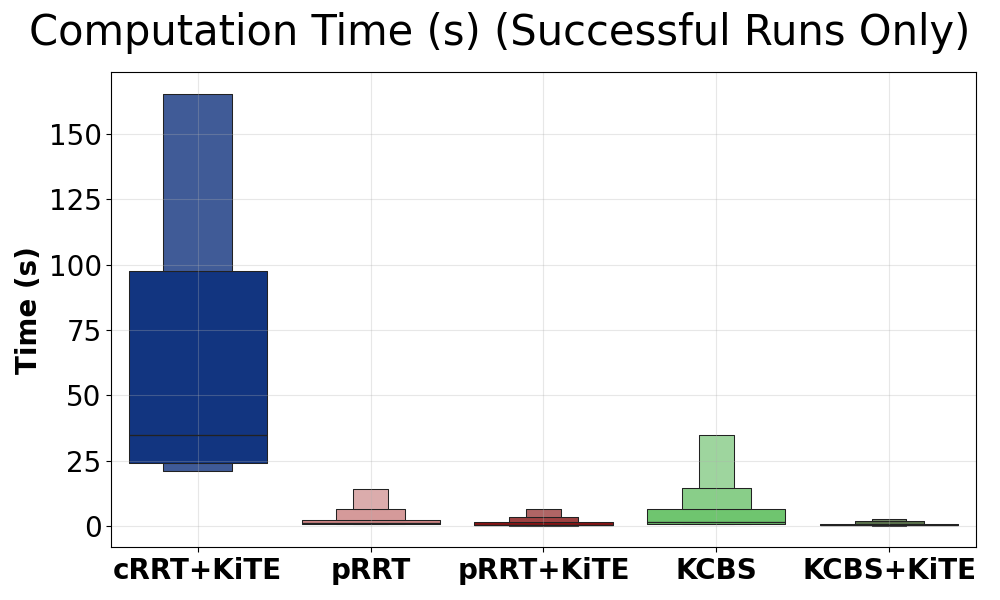}{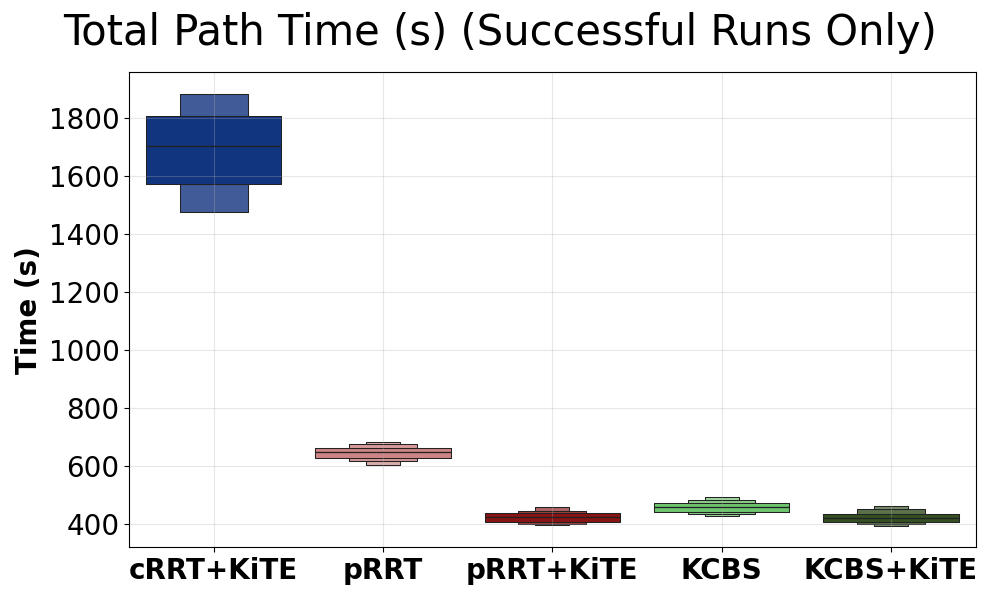}
\rowsubcap{Large Cluttered 3D}{\(N=15\) robots}

\imgrow{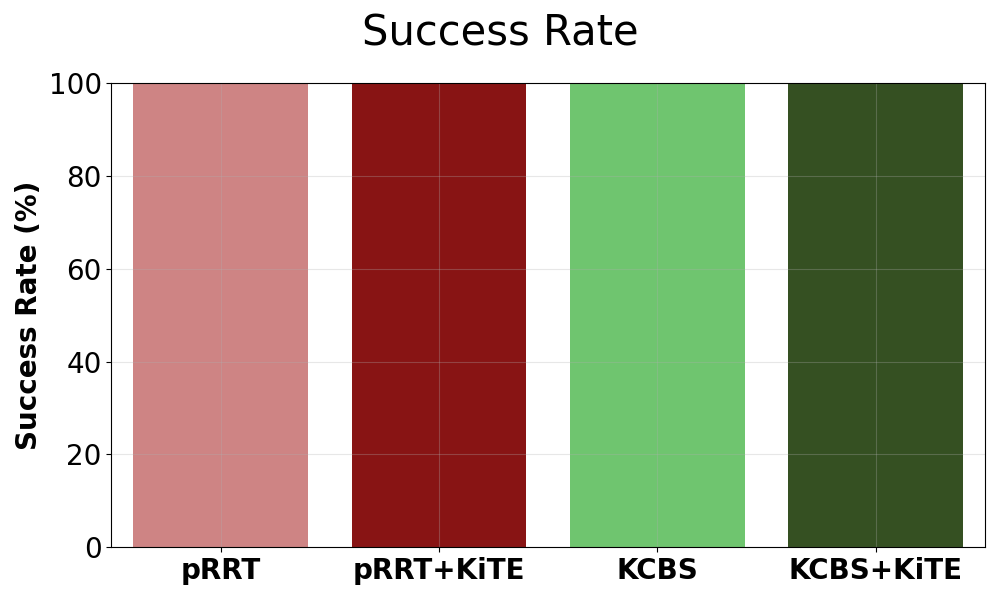}{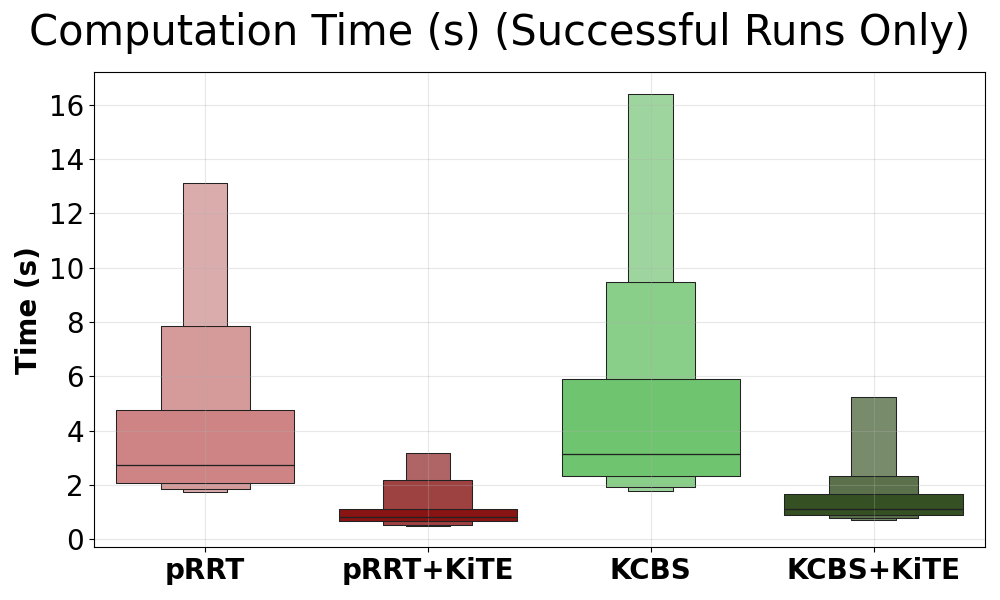}{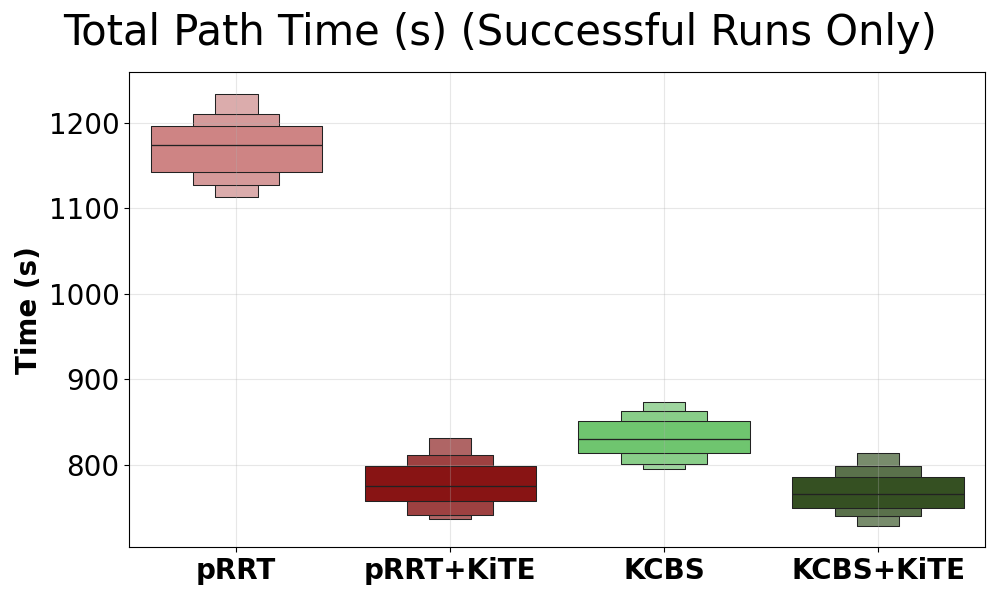}
\rowsubcap{Large Cluttered 3D}{\(N=30\) robots}

\caption{
Detailed results for selected environments in experiments with the Double Integrator model.
Each row corresponds to one experimental setting (environment and agent count \(N\)) and contains, from left to right:
(i) success rate (SR; bar chart), (ii) computation time (CT; boxen plot), and (iii) total path time (PT; boxen plot),
comparing each baseline planner to its KiTE-Extend variant.
These plots visualize variability and tail behavior underlying the aggregate results in Table~\ref{table_di_complete_results}.
}
\label{fig:di_rows}
\end{figure*}

Table~\ref{table_di_complete_results} reports complete results for the Double Integrator (DI) model across all environments and planners.
KiTE-Extend consistently extends successful planning to higher robot counts for sampling-based planners, with particularly pronounced gains for cRRT, where baseline performance degrades rapidly and KiTE-Extend recovers high success rates across both swap and cluttered environments.
For pRRT and KCBS, KiTE-Extend yields substantial and consistent reductions in computation time and total path duration while preserving success rates across nearly all settings.
Taken together with the Unicycle and Second Order Car results, these findings indicate that KiTE’s benefits persist under higher-dimensional and more challenging dynamics, and are not limited to simpler motion models.
To complement the aggregate statistics reported in Table~\ref{table_di_complete_results}, 
Fig.~\ref{fig:di_rows} presents results for representative environments in experiments with the Double Integrator model, including success rate (bar charts) as well as computation time and total path duration (boxen plots).

\begin{table*}[!ht]
\caption{\small
\textbf{Planner performance for the Double Integrator (DI) model across environments and robot counts.}
Results compare baseline planners and their KiTE-Extend variants for cRRT, pRRT, KCBS, and dbCBS, reported in terms of success rate (SR), computation time (CT), and total path time (PT).
Bold entries indicate the best value for each metric within a given planner and environment; for success rate, all entries achieving the maximum value are bolded.}
\begin{center}

\resizebox{1\linewidth}{!}{%
\begin{tabular}{l|c|ccc|ccc|ccc|ccc}
\toprule
\textbf{Domain} & \textbf{Variant}
& \multicolumn{3}{c|}{\textbf{cRRT}}
& \multicolumn{3}{c|}{\textbf{pRRT}}
& \multicolumn{3}{c|}{\textbf{KCBS}}
& \multicolumn{3}{c}{\textbf{dbCBS}} \\

& 
& SR (\%) & CT (s) & PT (s)
& SR (\%) & CT (s) & PT (s)
& SR (\%) & CT (s) & PT (s)
& SR (\%) & CT (s) & PT (s) \\[1mm]
\hline
\hline

{Swap} & Base & \textbf{100} & 3.74 $\pm$ 0.53 & 214 $\pm$ 3.0 & \textbf{100} & 0.0484 $\pm$ 0.0013 & 58.7 $\pm$ 0.47 & \textbf{100} & 0.0808 $\pm$ 0.016 & 57.3 $\pm$ 0.43 & 100 & 231 $\pm$ 0.54 & 51.7 $\pm$ 0.0 \\
{(N=4)} & KiTE & \textbf{100} & \textbf{0.0722 $\pm$ 0.0029} & \textbf{103 $\pm$ 1.6} & \textbf{100} & \textbf{0.0127 $\pm$ 0.00022} & \textbf{53.0 $\pm$ 0.40} & \textbf{100} & \textbf{0.0246 $\pm$ 0.011} & \textbf{53.5 $\pm$ 0.42} & - & - & - \\
\hline
{Swap} & Base & 93 & 27.9 $\pm$ 4.1 & 313 $\pm$ 4.1 & \textbf{100} & 0.0549 $\pm$ 0.0012 & 69.4 $\pm$ 0.52 & \textbf{100} & 0.0632 $\pm$ 0.0012 & 70.7 $\pm$ 0.50 & 60 & 290 $\pm$ 0.69 & 64.6 $\pm$ 0.0 \\
{(N=5)} & KiTE & \textbf{100} & \textbf{0.115 $\pm$ 0.0031} & \textbf{140 $\pm$ 1.9} & \textbf{100} & \textbf{0.0160 $\pm$ 0.00030} & \textbf{64.1 $\pm$ 0.50} & \textbf{100} & \textbf{0.0272 $\pm$ 0.0084} & \textbf{63.8 $\pm$ 0.45} & - & - & - \\
\hline
{Swap} & Base & 2 & 173 $\pm$ 4.7 & 676 $\pm$ 31 & \textbf{100} & 0.105 $\pm$ 0.0027 & 120 $\pm$ 0.73 & \textbf{100} & 0.129 $\pm$ 0.0028 & 120 $\pm$ 0.65 & 0 & - & - \\
{(N=8)} & KiTE & \textbf{100} & \textbf{0.381 $\pm$ 0.0093} & \textbf{279 $\pm$ 3.9} & \textbf{100} & \textbf{0.0263 $\pm$ 0.00062} & \textbf{108 $\pm$ 0.58} & \textbf{100} & \textbf{0.0297 $\pm$ 0.00057} & \textbf{109 $\pm$ 0.56} & - & - & - \\
\hline
{Swap} & Base & 0 & - & - & \textbf{100} & 0.111 $\pm$ 0.0030 & 123 $\pm$ 0.79 & \textbf{100} & 0.184 $\pm$ 0.022 & 123 $\pm$ 0.67 & 0 & - & - \\
{(N=10)} & KiTE & \textbf{100} & \textbf{0.621 $\pm$ 0.014} & \textbf{351 $\pm$ 5.8} & \textbf{100} & \textbf{0.0296 $\pm$ 0.00038} & \textbf{111 $\pm$ 0.60} & \textbf{100} & \textbf{0.0923 $\pm$ 0.022} & \textbf{111 $\pm$ 0.61} & - & - & - \\
\hline
{Swap} & Base & 0 & - & - & \textbf{100} & 0.198 $\pm$ 0.0048 & 203 $\pm$ 1.2 & \textbf{100} & 0.214 $\pm$ 0.0036 & 199 $\pm$ 0.84 & 0 & - & - \\
{(N=15)} & KiTE & \textbf{100} & \textbf{1.79 $\pm$ 0.035} & \textbf{663 $\pm$ 8.1} & \textbf{100} & \textbf{0.0460 $\pm$ 0.00050} & \textbf{183 $\pm$ 0.86} & \textbf{100} & \textbf{0.0511 $\pm$ 0.00064} & \textbf{181 $\pm$ 0.71} & - & - & - \\
\hline
{Swap} & Base & 0 & - & - & \textbf{100} & 0.269 $\pm$ 0.0073 & 253 $\pm$ 1.4 & \textbf{100} & 0.275 $\pm$ 0.0047 & 251 $\pm$ 0.97 & 0 & - & - \\
{(N=20)} & KiTE & \textbf{100} & \textbf{3.57 $\pm$ 0.060} & \textbf{989 $\pm$ 12} & \textbf{100} & \textbf{0.0635 $\pm$ 0.0017} & \textbf{229 $\pm$ 0.90} & \textbf{100} & \textbf{0.0639 $\pm$ 0.00081} & \textbf{226 $\pm$ 0.89} & - & - & - \\
\hline
{Swap} & Base & 0 & - & - & \textbf{100} & 0.347 $\pm$ 0.0081 & 339 $\pm$ 1.7 & \textbf{100} & 0.380 $\pm$ 0.0057 & 335 $\pm$ 1.1 & 0 & - & - \\
{(N=25)} & KiTE & \textbf{100} & \textbf{6.64 $\pm$ 0.096} & \textbf{1444 $\pm$ 18} & \textbf{100} & \textbf{0.0837 $\pm$ 0.0020} & \textbf{307 $\pm$ 0.93} & \textbf{100} & \textbf{0.0918 $\pm$ 0.0013} & \textbf{305 $\pm$ 0.78} & - & - & - \\
\hline
{Swap} & Base & 0 & - & - & \textbf{100} & 0.473 $\pm$ 0.012 & 421 $\pm$ 2.1 & \textbf{100} & 0.461 $\pm$ 0.0066 & 409 $\pm$ 1.4 & 0 & - & - \\
{(N=30)} & KiTE & \textbf{100} & \textbf{11.7 $\pm$ 0.17} & \textbf{1864 $\pm$ 20} & \textbf{100} & \textbf{0.0989 $\pm$ 0.00082} & \textbf{375 $\pm$ 1.2} & \textbf{100} & \textbf{0.111 $\pm$ 0.0013} & \textbf{373 $\pm$ 1.1} & - & - & - \\
\hline
{Large\;Cluttered} & Base & \textbf{100} & 0.633 $\pm$ 0.029 & 179 $\pm$ 2.2 & \textbf{100} & 0.109 $\pm$ 0.0058 & 115 $\pm$ 0.87 & \textbf{100} & 0.113 $\pm$ 0.0042 & 81.1 $\pm$ 0.65 & 0 & - & - \\
{(N=2)} & KiTE & \textbf{100} & \textbf{0.0865 $\pm$ 0.0040} & \textbf{101 $\pm$ 1.2} & \textbf{100} & \textbf{0.0303 $\pm$ 0.0015} & \textbf{72.2 $\pm$ 0.63} & \textbf{100} & \textbf{0.0462 $\pm$ 0.0092} & \textbf{72.7 $\pm$ 0.60} & - & - & - \\
\hline
{Large\;Cluttered} & Base & \textbf{100} & 11.9 $\pm$ 2.9 & 347 $\pm$ 3.5 & \textbf{100} & 0.159 $\pm$ 0.0050 & 172 $\pm$ 1.1 & \textbf{100} & 0.214 $\pm$ 0.015 & 121 $\pm$ 0.80 & 0 & - & - \\
{(N=3)} & KiTE & \textbf{100} & \textbf{0.468 $\pm$ 0.17} & \textbf{182 $\pm$ 2.3} & \textbf{100} & \textbf{0.0517 $\pm$ 0.0018} & \textbf{110 $\pm$ 0.76} & \textbf{100} & \textbf{0.0696 $\pm$ 0.0085} & \textbf{110 $\pm$ 0.79} & - & - & - \\
\hline
{Large\;Cluttered} & Base & 59 & 100 $\pm$ 11 & 573 $\pm$ 7.3 & \textbf{100} & 0.251 $\pm$ 0.0082 & 230 $\pm$ 1.4 & \textbf{100} & 0.300 $\pm$ 0.020 & 161 $\pm$ 0.91 & 0 & - & - \\
{(N=4)} & KiTE & \textbf{100} & \textbf{2.49 $\pm$ 0.54} & \textbf{287 $\pm$ 3.8} & \textbf{100} & \textbf{0.0807 $\pm$ 0.0035} & \textbf{145 $\pm$ 0.91} & \textbf{100} & \textbf{0.118 $\pm$ 0.015} & \textbf{147 $\pm$ 0.88} & - & - & - \\
\hline
{Large\;Cluttered} & Base & 1 & 154 $\pm$ 0.0 & 785 $\pm$ 0.0 & \textbf{100} & 0.290 $\pm$ 0.0078 & 272 $\pm$ 1.6 & \textbf{100} & 0.444 $\pm$ 0.034 & 195 $\pm$ 1.1 & 0 & - & - \\
{(N=5)} & KiTE & \textbf{100} & \textbf{5.32 $\pm$ 1.6} & \textbf{381 $\pm$ 5.1} & \textbf{100} & \textbf{0.0970 $\pm$ 0.0029} & \textbf{176 $\pm$ 1.1} & \textbf{100} & \textbf{0.179 $\pm$ 0.024} & \textbf{176 $\pm$ 1.1} & - & - & - \\
\hline
{Large\;Cluttered} & Base & 0 & - & - & \textbf{100} & 0.434 $\pm$ 0.034 & 299 $\pm$ 1.5 & \textbf{100} & 0.593 $\pm$ 0.052 & 214 $\pm$ 1.1 & 0 & - & - \\
{(N=6)} & KiTE & \textbf{98} & \textbf{8.53 $\pm$ 2.3} & \textbf{466 $\pm$ 6.0} & \textbf{100} & \textbf{0.118 $\pm$ 0.0047} & \textbf{194 $\pm$ 0.95} & \textbf{100} & \textbf{0.218 $\pm$ 0.027} & \textbf{194 $\pm$ 1.1} & - & - & - \\
\hline
{Large\;Cluttered} & Base & 0 & - & - & \textbf{100} & 1.59 $\pm$ 0.23 & 382 $\pm$ 2.1 & \textbf{100} & 3.49 $\pm$ 0.84 & 274 $\pm$ 1.4 & 0 & - & - \\
{(N=8)} & KiTE & \textbf{93} & \textbf{15.7 $\pm$ 3.2} & \textbf{725 $\pm$ 11} & \textbf{100} & \textbf{0.640 $\pm$ 0.12} & \textbf{253 $\pm$ 1.4} & \textbf{100} & \textbf{1.21 $\pm$ 0.38} & \textbf{252 $\pm$ 1.5} & - & - & - \\
\hline
{Large\;Cluttered} & Base & 0 & - & - & 99 & 3.91 $\pm$ 0.73 & 463 $\pm$ 2.3 & \textbf{100} & 7.00 $\pm$ 1.3 & 329 $\pm$ 1.5 & 0 & - & - \\
{(N=10)} & KiTE & \textbf{85} & \textbf{27.3 $\pm$ 5.6} & \textbf{984 $\pm$ 14} & \textbf{100} & \textbf{1.21 $\pm$ 0.31} & \textbf{303 $\pm$ 1.6} & \textbf{100} & \textbf{1.36 $\pm$ 0.28} & \textbf{303 $\pm$ 1.5} & - & - & - \\
\hline
{Large\;Cluttered} & Base & 0 & - & - & \textbf{100} & 3.89 $\pm$ 0.71 & 648 $\pm$ 3.1 & \textbf{100} & 7.63 $\pm$ 1.5 & 460 $\pm$ 2.2 & 0 & - & - \\
{(N=15)} & KiTE & \textbf{43} & \textbf{68.4 $\pm$ 10} & \textbf{1700 $\pm$ 30} & \textbf{100} & \textbf{1.97 $\pm$ 0.43} & \textbf{426 $\pm$ 2.1} & \textbf{100} & \textbf{1.06 $\pm$ 0.17} & \textbf{424 $\pm$ 2.2} & - & - & - \\
\hline
{Large\;Cluttered} & Base & 0 & - & - & 99 & 2.80 $\pm$ 0.41 & 836 $\pm$ 3.0 & \textbf{100} & 5.14 $\pm$ 1.4 & 599 $\pm$ 2.8 & 0 & - & - \\
{(N=20)} & KiTE & \textbf{5} & \textbf{165 $\pm$ 44} & \textbf{2917 $\pm$ 177} & \textbf{100} & \textbf{1.27 $\pm$ 0.23} & \textbf{553 $\pm$ 2.4} & \textbf{100} & \textbf{1.14 $\pm$ 0.14} & \textbf{548 $\pm$ 2.1} & - & - & - \\
\hline
{Large\;Cluttered} & Base & 0 & - & - & 99 & 3.51 $\pm$ 0.37 & 1018 $\pm$ 4.1 & \textbf{100} & 4.26 $\pm$ 0.42 & 727 $\pm$ 2.7 & 0 & - & - \\
{(N=25)} & KiTE & \textbf{5} & \textbf{144 $\pm$ 8.2} & \textbf{3611 $\pm$ 195} & \textbf{100} & \textbf{1.14 $\pm$ 0.13} & \textbf{676 $\pm$ 2.7} & \textbf{100} & \textbf{1.83 $\pm$ 0.21} & \textbf{668 $\pm$ 2.5} & - & - & - \\
\hline
{Large\;Cluttered} & Base & 0 & - & - & \textbf{100} & 4.55 $\pm$ 0.46 & 1171 $\pm$ 3.9 & \textbf{100} & 5.60 $\pm$ 0.66 & 831 $\pm$ 2.8 & 0 & - & - \\
{(N=30)} & KiTE & 0 & - & - & \textbf{100} & \textbf{1.51 $\pm$ 0.25} & \textbf{780 $\pm$ 3.2} & \textbf{100} & \textbf{2.02 $\pm$ 0.39} & \textbf{768 $\pm$ 2.8} & - & - & - \\
\hline

\bottomrule
\end{tabular}
}

\vspace{0.5ex}
{\scriptsize SR: Success Rate \textbar{} CT: Computation Time \textbar{} PT: Total Path Time}
\label{table_di_complete_results}
\end{center}
\end{table*}

\FloatBarrier

\FloatBarrier

\clearpage

\subsection{Effect of pruning in KCBS for the Second-Order Car model.}
Table~\ref{table_kcbs_pruning_results_soc} compares KCBS with and without low-level tree pruning for the SOC model, both with and without KiTE-Extend.
Across both environments, pruning primarily reduces computation time (CT) at moderate to high agent counts by avoiding full replanning when new conflicts are introduced, while largely preserving success rates.

When combined with KiTE-Extend, pruning sometimes yields slightly higher total path time (PT), particularly at higher densities.
This behavior is consistent with KiTE-Extend’s ability to select low-velocity or wait-like trajectory segments that resolve conflicts by slowing down rather than rerouting entirely.
Under pruning, such adaptations are often realized by modifying existing trajectories instead of discovering alternative homotopy classes, improving feasibility and efficiency at the expense of longer execution times.

In lower-density or less constrained settings, pruning provides limited benefit.
In these settings, the overhead of pruning and tree maintenance can offset its advantages, leading to similar performance between KCBS and KCBS+Pruning, for both the base planner and their KiTE-Extend variants, as conflicts are infrequent and replanning is rarely triggered.

\begin{figure*}[t!]
\centering

\imgrow{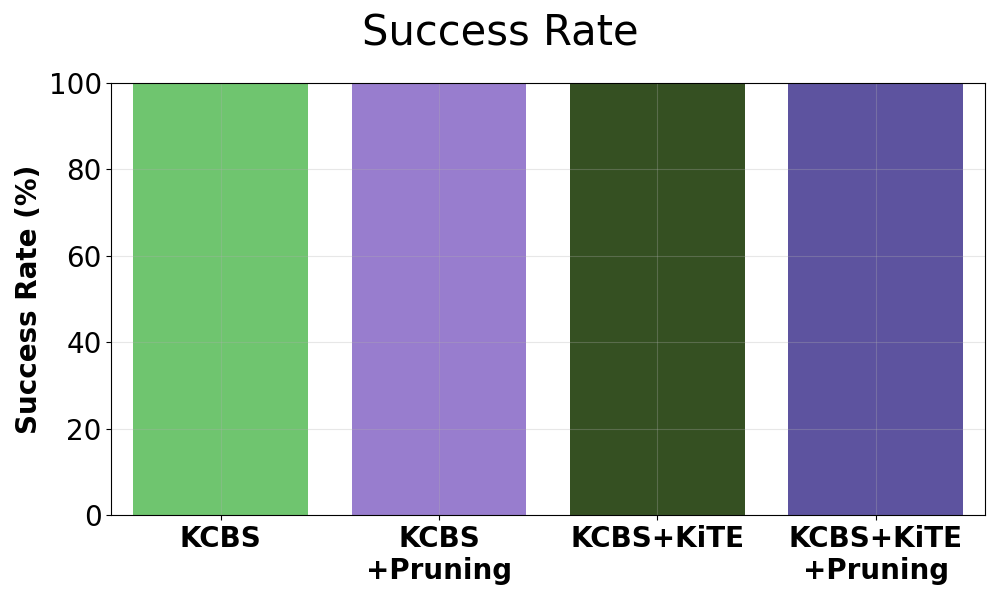}{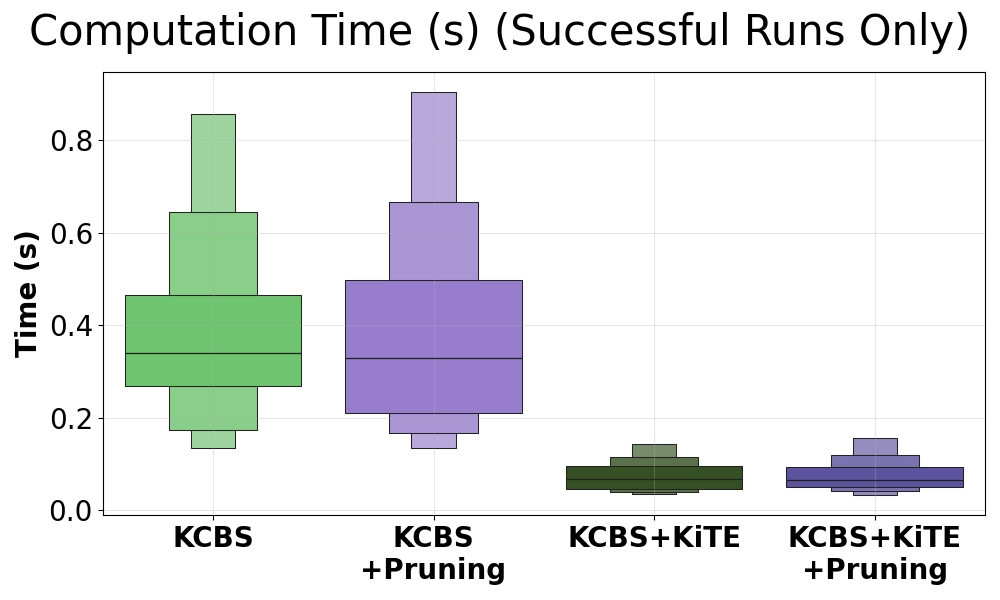}{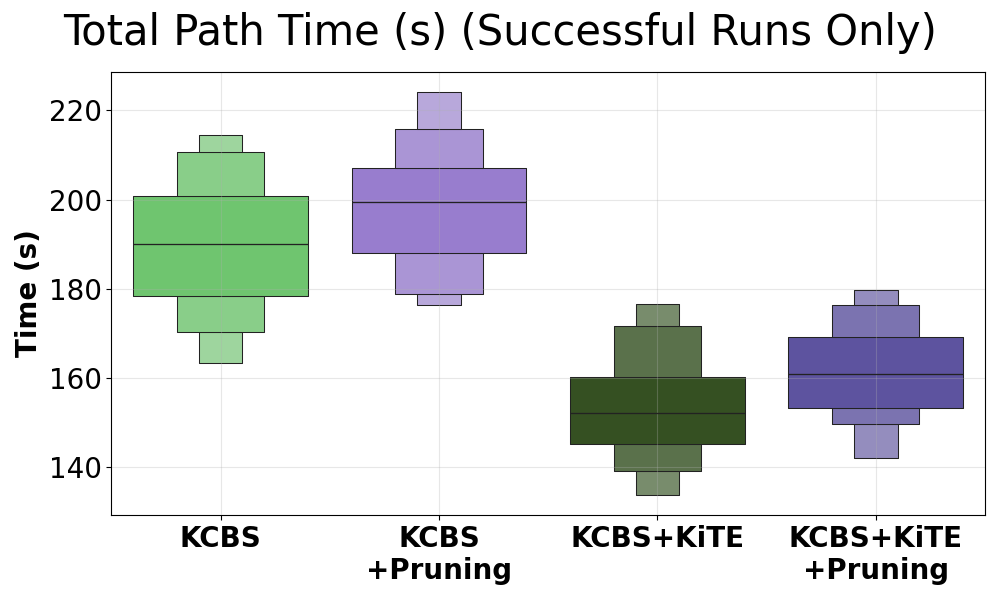}
\rowsubcap{SWAP}{\(N=5\) robots}

\imgrow{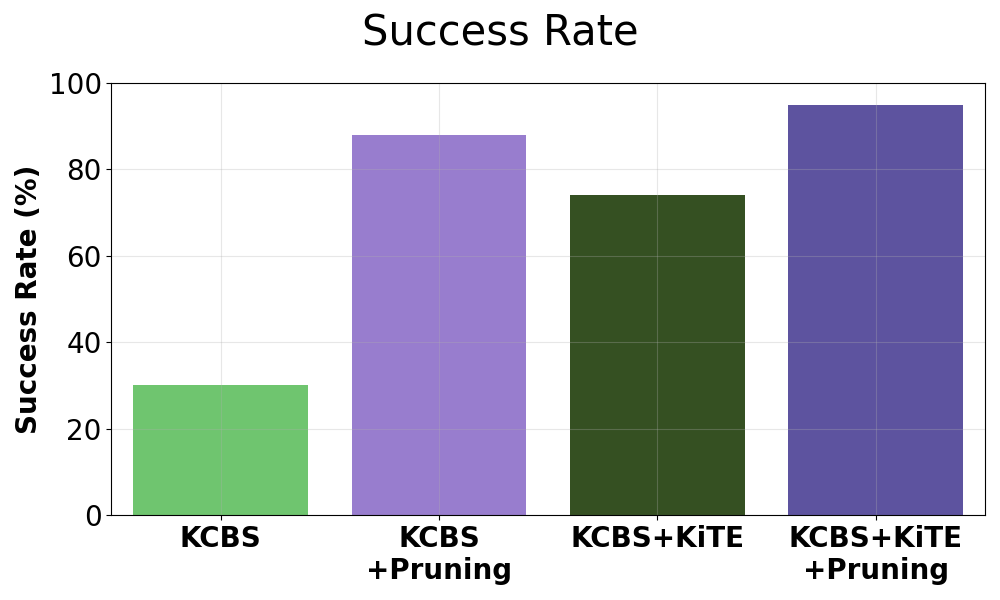}{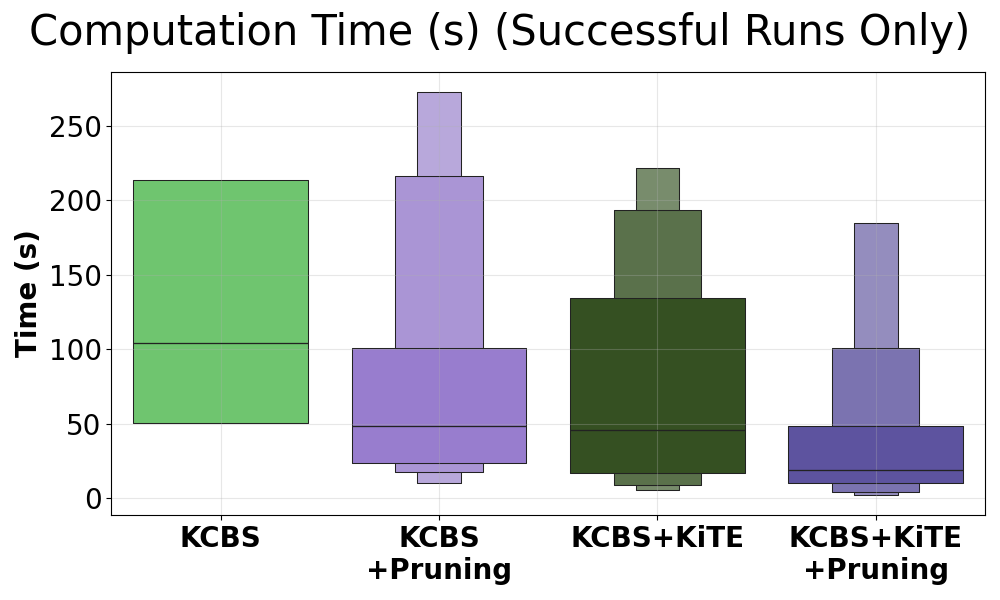}{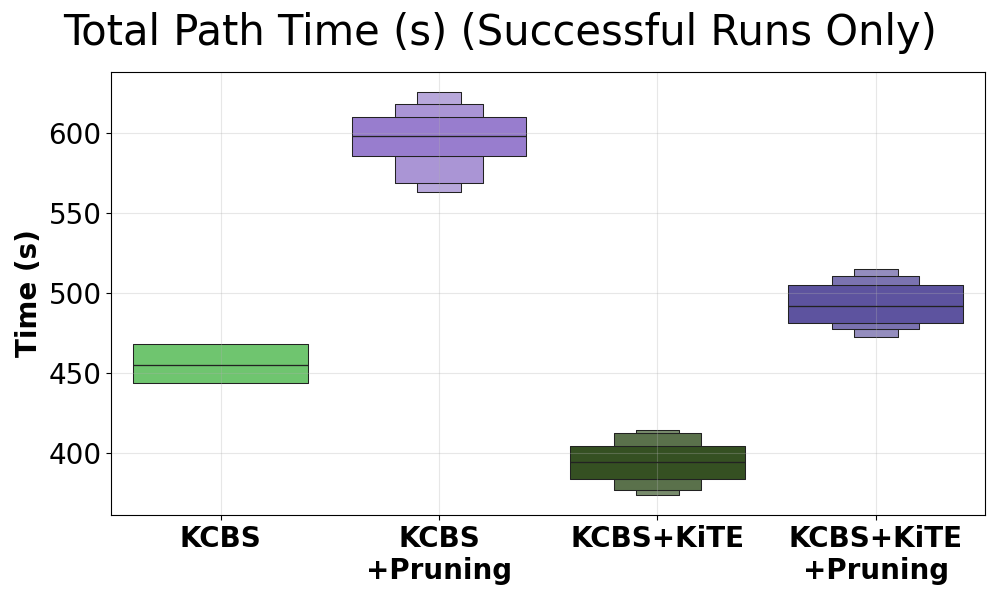}
\rowsubcap{SWAP}{\(N=15\) robots}

\imgrow{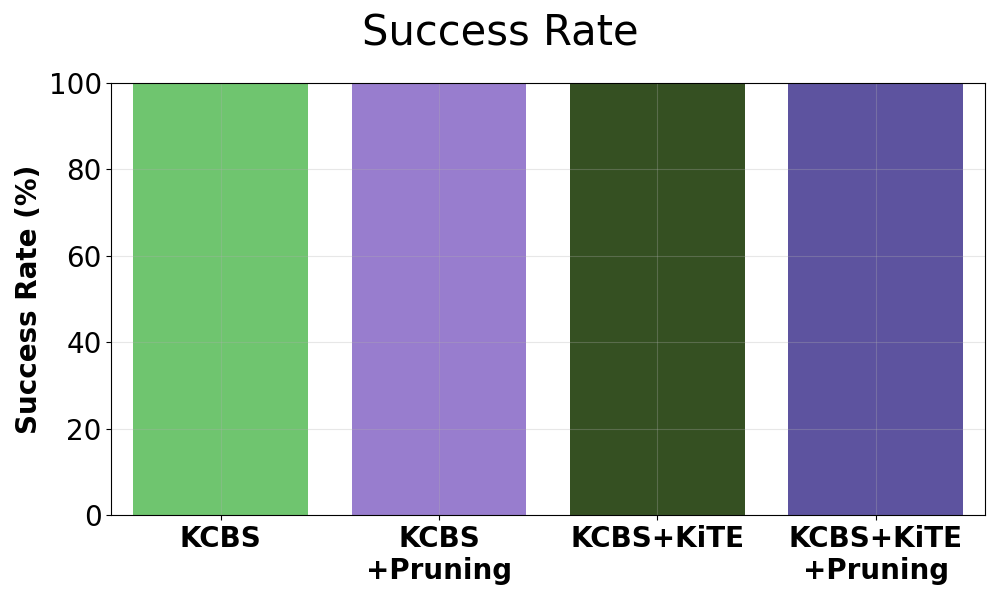}{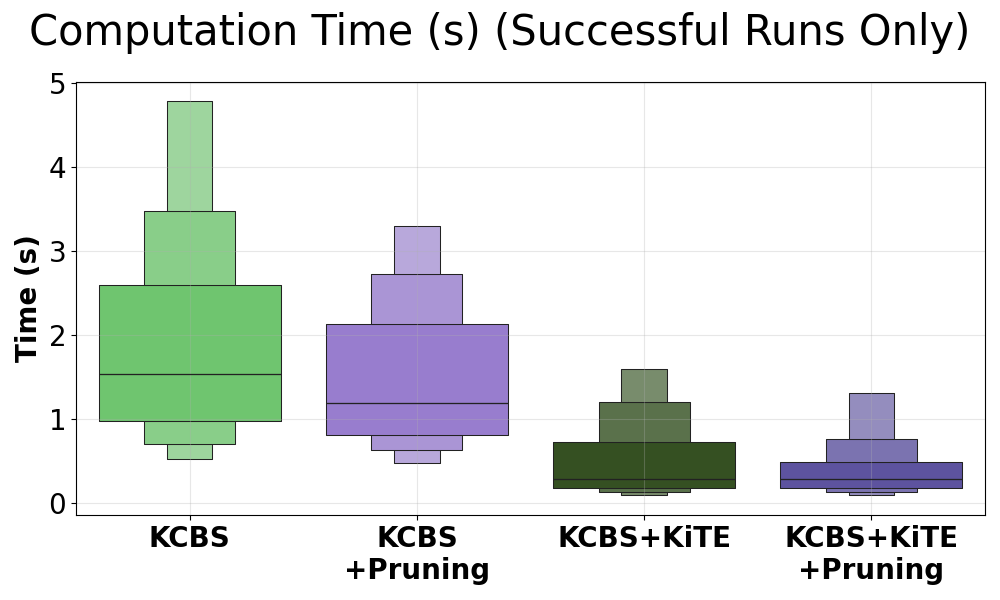}{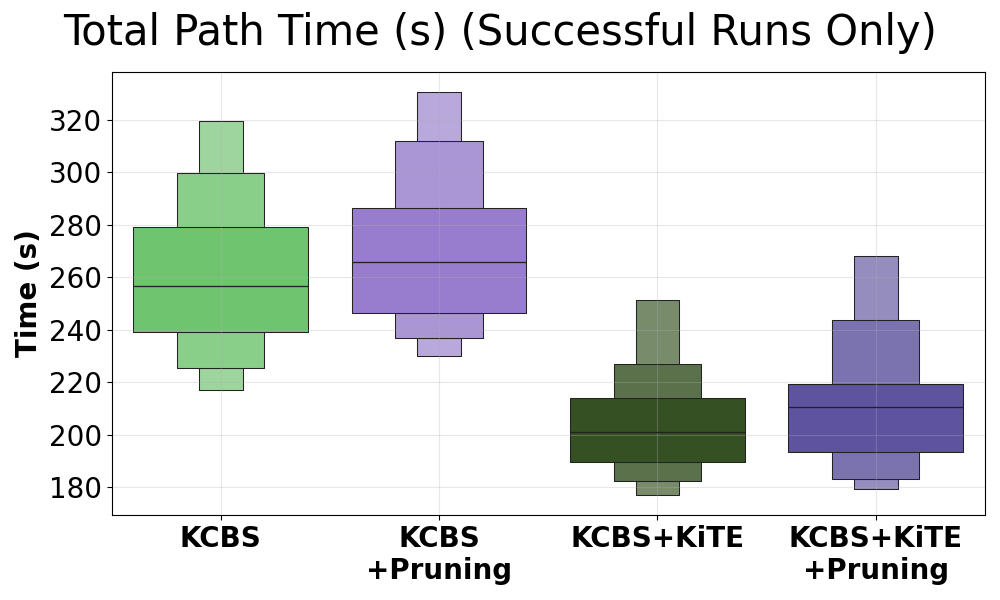}
\rowsubcap{Large Cluttered}{\(N=4\) robots}

\imgrow{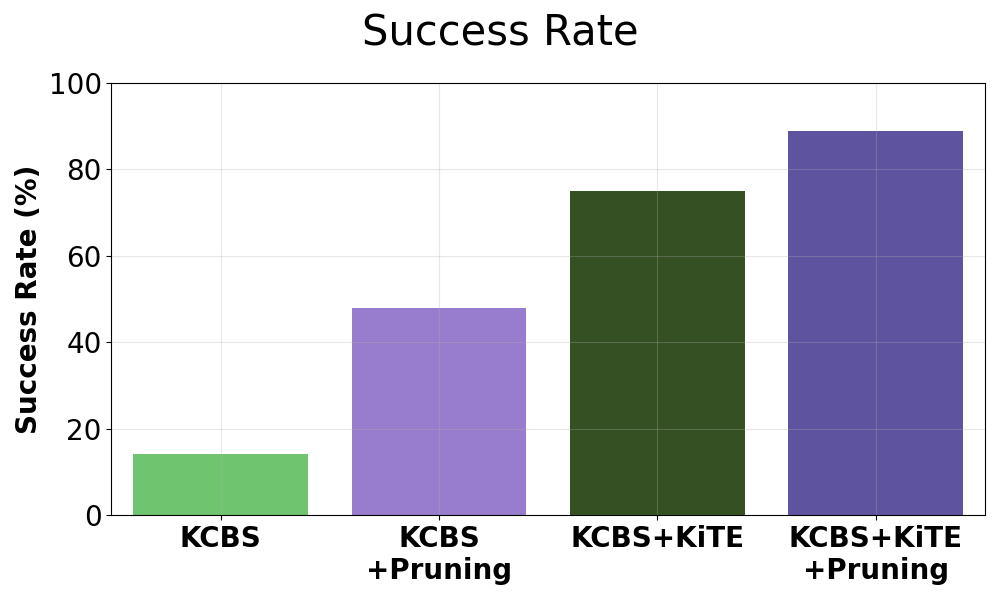}{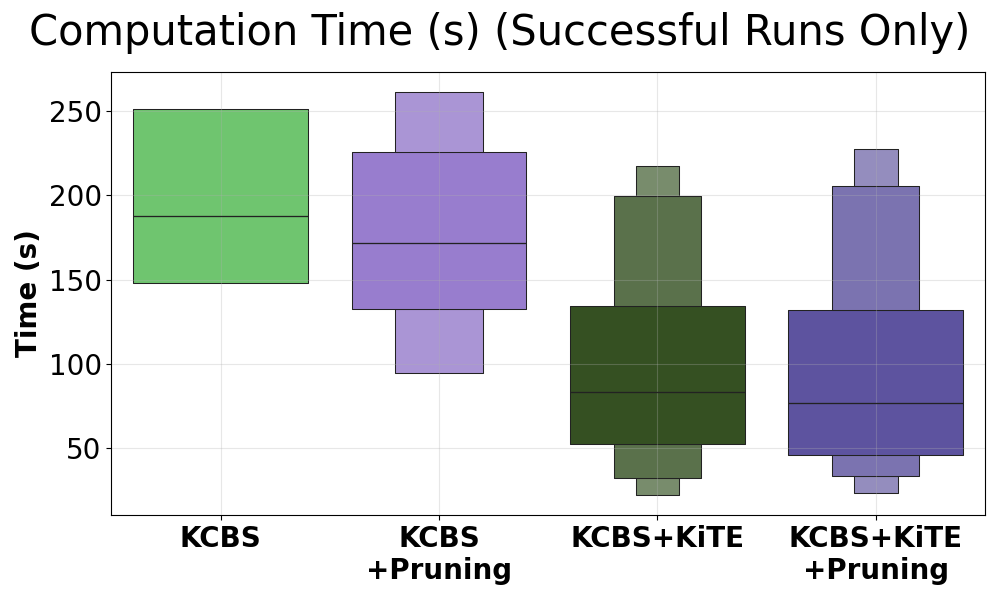}{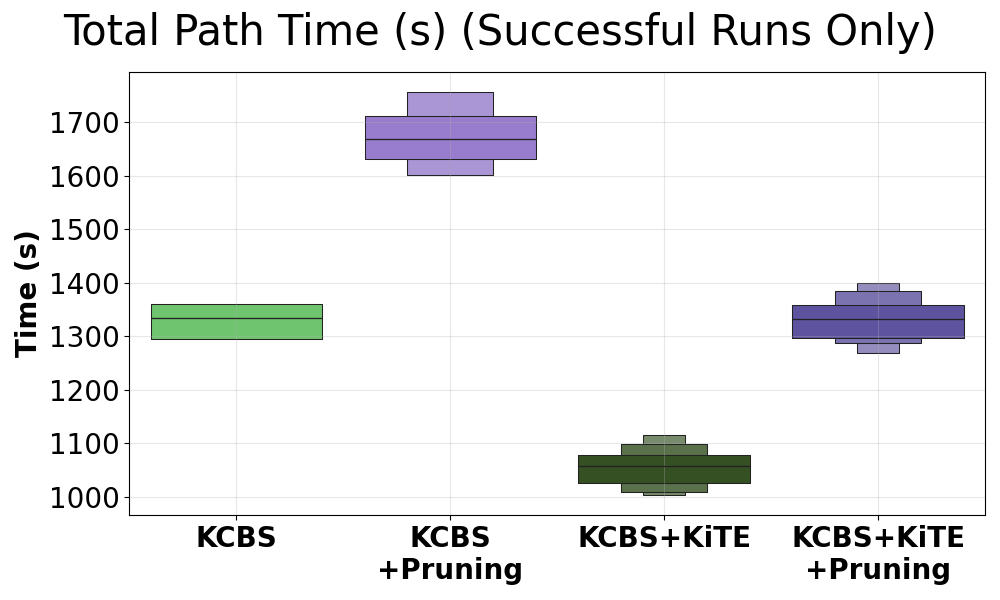}
\rowsubcap{Large Cluttered}{\(N=30\) robots}

\caption{
Detailed results for selected environments in experiments with the Second Order Car model.
Each row corresponds to one experimental setting (environment and agent count \(N\)) and contains, from left to right:
(i) success rate (SR; bar chart), (ii) computation time (CT; boxen plot), and (iii) total path time (PT; boxen plot),
comparing each KCBS planner to its pruning enabled variant.
These plots visualize variability and tail behavior underlying the aggregate results in Table~\ref{table_kcbs_pruning_results_soc}.
}
\label{fig:kcbs_pruning_rows}
\end{figure*}

To complement the aggregate statistics reported in Table~\ref{table_kcbs_pruning_results_soc},
the accompanying boxen plots in Fig.~\ref{fig:kcbs_pruning_rows} visualize the distribution of computation time and total path time
for representative SOC settings, highlighting variability and tail behavior underlying the reported means.

\begin{table*}[!ht]
\caption{\small
Planner performance for KCBS with and without low-level pruning under the Second-Order Car (SOC) model.
Results compare baseline KCBS and KCBS+Pruning planners, both with and without KiTE-Extend, across Swap and Large Cluttered environments and varying robot counts.
Metrics include success rate (SR), computation time (CT), and total path time (PT).
Bold entries indicate the best value for each metric within a given planner and environment; for success rate, all entries achieving the maximum value are bolded.
}
\begin{center}

\resizebox{1\linewidth}{!}{%
\begin{tabular}{l|c|ccc|ccc}
\toprule
\textbf{Domain} & \textbf{Variant}
& \multicolumn{3}{c|}{\textbf{KCBS}} & \multicolumn{3}{c}{\textbf{KCBS+Pruning}} \\

& 
& SR & CT (s) & PT (s) & SR & CT (s) & PT (s) \\[1mm]
\hline
\hline

{Swap} & Base & \textbf{100} & 0.145 $\pm$ 0.0098 & 113 $\pm$ 1.5 & \textbf{100} & 0.149 $\pm$ 0.011 & 114 $\pm$ 1.4 \\
{(N=3)} & KiTE & \textbf{100} & \textbf{0.0259 $\pm$ 0.0018} & \textbf{89.7 $\pm$ 0.93} & \textbf{100} & \textbf{0.0258 $\pm$ 0.0018} & \textbf{90.3 $\pm$ 0.91} \\
\hline
{Swap} & Base & \textbf{100} & 0.585 $\pm$ 0.052 & 162 $\pm$ 1.4 & \textbf{100} & 0.454 $\pm$ 0.035 & 168 $\pm$ 1.5 \\
{(N=4)} & KiTE & \textbf{100} & \textbf{0.0905 $\pm$ 0.014} & \textbf{130 $\pm$ 0.99} & \textbf{100} & \textbf{0.0662 $\pm$ 0.0043} & \textbf{136 $\pm$ 0.99} \\
\hline
{Swap} & Base & \textbf{100} & 0.404 $\pm$ 0.024 & 190 $\pm$ 1.7 & \textbf{100} & 0.408 $\pm$ 0.028 & 198 $\pm$ 1.5 \\
{(N=5)} & KiTE & \textbf{100} & \textbf{0.0790 $\pm$ 0.0052} & \textbf{154 $\pm$ 1.3} & \textbf{100} & \textbf{0.0778 $\pm$ 0.0041} & \textbf{161 $\pm$ 1.1} \\
\hline
{Swap} & Base & \textbf{100} & 2.31 $\pm$ 0.16 & 295 $\pm$ 2.1 & \textbf{100} & 2.04 $\pm$ 0.15 & 329 $\pm$ 2.0 \\
{(N=8)} & KiTE & \textbf{100} & \textbf{0.605 $\pm$ 0.059} & \textbf{236 $\pm$ 1.4} & \textbf{100} & \textbf{0.640 $\pm$ 0.066} & \textbf{268 $\pm$ 1.3} \\
\hline
{Swap} & Base & \textbf{100} & 11.9 $\pm$ 1.7 & 345 $\pm$ 2.4 & \textbf{100} & 6.14 $\pm$ 1.0 & 408 $\pm$ 1.7 \\
{(N=10)} & KiTE & \textbf{100} & \textbf{2.07 $\pm$ 0.21} & \textbf{284 $\pm$ 1.3} & \textbf{100} & \textbf{1.45 $\pm$ 0.15} & \textbf{333 $\pm$ 1.3} \\
\hline
{Swap} & Base & 30 & 128 $\pm$ 17 & 457 $\pm$ 4.1 & 88 & 82.6 $\pm$ 8.9 & 596 $\pm$ 2.3 \\
{(N=15)} & KiTE & \textbf{74} & \textbf{80.0 $\pm$ 9.0} & \textbf{395 $\pm$ 1.7} & \textbf{95} & \textbf{42.7 $\pm$ 5.9} & \textbf{493 $\pm$ 1.6} \\
\hline
{Swap} & Base & 0 & - & - & 2 & 180 $\pm$ 33 & 813 $\pm$ 3.6 \\
{(N=20)} & KiTE & 0 & - & - & \textbf{4} & \textbf{178 $\pm$ 39} & \textbf{671 $\pm$ 4.5} \\
\hline
{Swap} & Base & 0 & - & - & 0 & - & - \\
{(N=25)} & KiTE & 0 & - & - & 0 & - & - \\
\hline
{Large Cluttered} & Base & \textbf{100} & 1.43 $\pm$ 0.11 & 228 $\pm$ 3.0 & \textbf{100} & 1.10 $\pm$ 0.070 & 234 $\pm$ 3.1 \\
{(N=3)} & KiTE & \textbf{100} & \textbf{0.316 $\pm$ 0.033} & \textbf{180 $\pm$ 1.8} & \textbf{100} & \textbf{0.281 $\pm$ 0.032} & \textbf{181 $\pm$ 2.1} \\
\hline
{Large Cluttered} & Base & \textbf{100} & 1.97 $\pm$ 0.14 & 261 $\pm$ 3.5 & \textbf{100} & 1.62 $\pm$ 0.12 & 271 $\pm$ 3.3 \\
{(N=4)} & KiTE & \textbf{100} & \textbf{0.549 $\pm$ 0.062} & \textbf{206 $\pm$ 2.8} & \textbf{100} & \textbf{0.437 $\pm$ 0.043} & \textbf{212 $\pm$ 2.9} \\
\hline
{Large Cluttered} & Base & \textbf{100} & 2.11 $\pm$ 0.14 & 304 $\pm$ 3.1 & \textbf{100} & 1.85 $\pm$ 0.13 & 318 $\pm$ 3.4 \\
{(N=5)} & KiTE & \textbf{100} & \textbf{0.599 $\pm$ 0.071} & \textbf{246 $\pm$ 2.6} & \textbf{100} & \textbf{0.541 $\pm$ 0.058} & \textbf{252 $\pm$ 2.8} \\
\hline
{Large Cluttered} & Base & \textbf{100} & 4.92 $\pm$ 0.28 & 443 $\pm$ 3.3 & \textbf{100} & 4.65 $\pm$ 0.38 & 477 $\pm$ 4.4 \\
{(N=8)} & KiTE & \textbf{100} & \textbf{1.53 $\pm$ 0.14} & \textbf{355 $\pm$ 3.0} & \textbf{100} & \textbf{1.35 $\pm$ 0.11} & \textbf{373 $\pm$ 3.2} \\
\hline
{Large Cluttered} & Base & \textbf{100} & 12.3 $\pm$ 0.89 & 556 $\pm$ 4.5 & \textbf{100} & 11.5 $\pm$ 0.87 & 608 $\pm$ 5.1 \\
{(N=10)} & KiTE & \textbf{100} & \textbf{3.72 $\pm$ 0.34} & \textbf{450 $\pm$ 3.4} & \textbf{100} & \textbf{3.41 $\pm$ 0.33} & \textbf{480 $\pm$ 3.6} \\
\hline
{Large Cluttered} & Base & \textbf{100} & 30.7 $\pm$ 2.3 & 754 $\pm$ 4.3 & \textbf{100} & 23.1 $\pm$ 1.6 & 839 $\pm$ 5.8 \\
{(N=15)} & KiTE & \textbf{100} & \textbf{9.10 $\pm$ 0.73} & \textbf{615 $\pm$ 3.5} & \textbf{100} & \textbf{6.96 $\pm$ 0.68} & \textbf{663 $\pm$ 3.9} \\
\hline
{Large Cluttered} & Base & 99 & 76.9 $\pm$ 6.2 & 972 $\pm$ 5.1 & 98 & 47.3 $\pm$ 3.3 & 1077 $\pm$ 6.3 \\
{(N=20)} & KiTE & \textbf{100} & \textbf{20.2 $\pm$ 2.0} & \textbf{780 $\pm$ 3.3} & \textbf{100} & \textbf{15.8 $\pm$ 1.3} & \textbf{850 $\pm$ 4.8} \\
\hline
{Large Cluttered} & Base & 71 & 124 $\pm$ 8.0 & 1121 $\pm$ 5.1 & 92 & 96.9 $\pm$ 5.5 & 1405 $\pm$ 6.8 \\
{(N=25)} & KiTE & \textbf{98} & \textbf{33.7 $\pm$ 3.5} & \textbf{915 $\pm$ 4.0} & \textbf{99} & \textbf{45.1 $\pm$ 4.7} & \textbf{1109 $\pm$ 4.6} \\
\hline
{Large Cluttered} & Base & 14 & 201 $\pm$ 15 & 1325 $\pm$ 12 & 48 & 178 $\pm$ 9.8 & 1678 $\pm$ 9.6 \\
{(N=30)} & KiTE & \textbf{75} & \textbf{102 $\pm$ 8.0} & \textbf{1055 $\pm$ 4.4} & \textbf{89} & \textbf{96.8 $\pm$ 7.1} & \textbf{1334 $\pm$ 5.2} \\
\hline
\bottomrule
\end{tabular}
}

\vspace{0.5ex}
{\scriptsize 
KCBS: Kinodynamic CBS \textbar{}
-KiTE: planner augmented with KiTE \textbar{}
Pruning: Low-level KCBS Planner augmented to reuse portions of a previous tree
}
\label{table_kcbs_pruning_results_soc}
\end{center}
\end{table*}
\FloatBarrier

\FloatBarrier

\clearpage

\subsection{Ablation to analyze the effect of edge bundle size.}
Table~\ref{table_ablations_eb_count} varies the number of precomputed edges in the KiTE bundle for the SOC model in the Small Cluttered environment with \(N=15\) robots, using a fixed skip factor \(p=10\).
Results are averaged over a common subset of trials for each planner in which all tested bundle sizes succeeded, ensuring fair comparison across configurations.
As the bundle size increases, total path time (PT) decreases, consistent with a larger library providing higher-quality candidate segments for the progress-based ranking heuristic.
Computation time (CT) reflects a tradeoff between increased retrieval and sorting overhead and reduced downstream effort: larger bundles increase the candidate set \(|\mathcal{C}|\), but also increase the likelihood of selecting a valid edge early in the stride-\(p\) evaluation of the ranked list (Alg.~\ref{alg:kite_extend}), thereby reducing the number of expensive propagation and validity checks before acceptance.
This interaction can yield non-monotone CT behavior across bundle sizes while preserving the monotone improvement in PT.











\begin{table*}[!ht]
\caption{\small
Effect of KiTE edge-bundle size on planner performance for the Second-Order Car (SOC) model in the Small Cluttered environment with \(N=15\) robots.
Results are shown for KiTE-augmented cRRT, pRRT, and KCBS using edge bundles of varying size (25k, 50k, and 75k edges).
Reported metrics include computation time (CT) and total path time (PT), averaged over a common subset of trials for each planner in which all tested edge-bundle sizes successfully produced a solution.
The column ``Rounds'' indicates the number of such common trials used for aggregation.
}
\begin{center}

\resizebox{1\linewidth}{!}{%
\begin{tabular}{c|ccc|ccc|ccc}
\toprule
\textbf{\#Edges}
& \multicolumn{3}{c|}{\textbf{cRRT+KiTE}}
& \multicolumn{3}{c|}{\textbf{pRRT+KiTE}}
& \multicolumn{3}{c}{\textbf{KCBS+KiTE}} \\

& Rounds & CT (s) & PT (s)
& Rounds & CT (s) & PT (s)
& Rounds & CT (s) & PT (s) \\[1mm]
\hline
\hline

25k
& 4  & 204 $\pm$ 32  & 2302 $\pm$ 296
& 50 & 34.8 $\pm$ 8.7 & 509 $\pm$ 9.8
& 99 & 30.4 $\pm$ 3.0 & 257 $\pm$ 1.7 \\

\hline

50k
& 4  & 260 $\pm$ 4.6 & 1993 $\pm$ 71
& 50 & 22.7 $\pm$ 4.5 & 505 $\pm$ 8.5
& 99 & 42.8 $\pm$ 4.6 & 250 $\pm$ 1.9 \\

\hline

75k
& 4  & 264 $\pm$ 25 & 1691 $\pm$ 190
& 50 & 24.5 $\pm$ 6.1 & 488 $\pm$ 8.8
& 99 & 35.8 $\pm$ 3.5 & 244 $\pm$ 1.5 \\

\hline
\bottomrule
\end{tabular}
}

\vspace{0.5ex}
{\scriptsize
CT: Computation Time \textbar{}
PT: Total Path Time \textbar{}
Rounds: Number of trials (random seeds) for which all tested edge-bundle sizes succeeded for a given planner
}
\label{table_ablations_eb_count}
\end{center}
\end{table*}

\subsection{Ablation to analyze the effect of skip factor $p$}
Table~\ref{table_ablations_num_skip_edges} varies the skip factor \(p\) used to subsample the ranked candidate list in the KiTE-Extend routine (Alg.~\ref{alg:kite_extend}), while fixing the edge-bundle size to 50k for the SOC model in the Small Cluttered environment with \(N=15\) robots.
Across planners, total path time (PT) remains largely invariant to \(p\), indicating that the ranking heuristic tends to select similarly effective edges once a valid candidate is found.
Computation time (CT) varies non-monotonically with \(p\), reflecting a tradeoff between evaluating more candidates (smaller \(p\)) and potentially delaying selection of a valid edge by skipping viable candidates in the ranked list (larger \(p\)).
The \(p\) value that minimizes CT is planner-dependent in this setting.
Results for cRRT+KiTE are omitted because there were no trials in which all tested \(p\) values succeeded under the common-success evaluation protocol (Rounds \(=0\)).











\begin{table*}[!ht]
\caption{\small
Effect of skip factor \(p\) on KiTE-augmented planner performance for the Second-Order Car (SOC) model in the Small Cluttered environment with \(N=15\) robots, using a fixed edge-bundle size of 50k.
Results are reported for pRRT+KiTE and KCBS+KiTE in terms of computation time (CT) and total path time (PT).
Metrics are averaged over a common subset of trials for each planner in which all tested skip-factor settings successfully produced a solution.
The column ``Rounds'' indicates the number of such common trials used for aggregation.
}
\begin{center}

\resizebox{1\linewidth}{!}{%
\begin{tabular}{c|ccc|ccc}
\toprule
\textbf{Skip factor \(p\)}
& \multicolumn{3}{c|}{\textbf{pRRT+KiTE}}
& \multicolumn{3}{c}{\textbf{KCBS+KiTE}} \\

& Rounds & CT (s) & PT (s)
& Rounds & CT (s) & PT (s) \\[1mm]
\hline
\hline

5
& 54 & 41.4 $\pm$ 8.8 & 507 $\pm$ 9.4
& 99 & 27.5 $\pm$ 2.3 & 251 $\pm$ 1.8 \\

\hline

10
& 54 & 29.1 $\pm$ 7.1 & 509 $\pm$ 8.8
& 99 & 42.7 $\pm$ 4.6 & 250 $\pm$ 1.9 \\

\hline

25
& 54 & 43.4 $\pm$ 8.4 & 509 $\pm$ 9.4
& 99 & 39.3 $\pm$ 4.3 & 246 $\pm$ 1.5 \\

\hline
\bottomrule
\end{tabular}
}

\vspace{0.5ex}
{\scriptsize
CT: Computation Time \textbar{}
PT: Total Path Time \textbar{}
Rounds: Number of trials (random seeds) for which all tested skip-factor values succeeded for a given planner
}
\label{table_ablations_num_skip_edges}
\end{center}
\end{table*}

\FloatBarrier

\subsection*{Summary of Appendix Findings.}
Taken together, the results in this appendix demonstrate that the performance gains of KiTE-Extend are consistent across motion models, environments, and coordination strategies.
The complete tables and distributional plots corroborate the aggregate trends reported in the main paper, while the pruning analysis and ablations provide insight into how KiTE-Extend interacts with low-level planning reuse and how its parameters influence the tradeoff between computation time and path quality.
These additional results support the robustness of KiTE-Extend’s benefits without altering the conclusions drawn in the main text.

\end{document}